%% file: iclr2027_conference.tex
\documentclass{article} 
\PassOptionsToPackage{hyphens}{url}
\usepackage{iclr2027_conference,times}

\input{math_commands.tex}

\usepackage[utf8]{inputenc} 
\usepackage[T1]{fontenc}    
\usepackage[hypertexnames=false]{hyperref}       
\usepackage{url}            
\usepackage{xurl}
\usepackage{booktabs}       
\usepackage{amsmath}
\usepackage{amsfonts}       
\usepackage{nicefrac}       
\usepackage{microtype}      
\usepackage{xcolor}         
\usepackage{graphicx}
\usepackage{tabularray}
\usepackage{float}
\usepackage{soul}
\usepackage{tcolorbox}
\usepackage{wrapfig}
\usepackage{capt-of}        

\usepackage{pifont}
\usepackage{tikz}
\usetikzlibrary{arrows.meta,positioning}
\newcommand{\cmark}{\textcolor{green!45!black}{\ding{51}}}
\newcommand{\xmark}{\textcolor{black!55}{\ding{55}}}
\newcommand{\pmark}{\textcolor{orange!80!black}{\ensuremath{\sim}}}

\providecommand{\todo}[1]{{\color{red}\textbf{[TODO: #1]}}}
\tcbuselibrary{breakable, skins}
\let\checkmark\relax   
\usepackage{dingbat}
\usepackage{enumitem}
\newcommand{\ours}[0]{PRISM}

\title{PRISM: A Multi-Dimensional Benchmark for Evaluating LLM Peer Reviewers}

\iclrfinalcopy

\ificlrfinal
\author{%
  \makebox[\dimexpr\textwidth-2\tabcolsep\relax][c]{%
  \begin{tabular}[t]{@{}c@{}}
    Ngoc Phan Phuoc Loc\textsuperscript{1}\thanks{Co-first Authors.}\quad
    Toan Huynh La Viet\textsuperscript{1}\footnotemark[1]\quad
    Thanh Tran Khanh\textsuperscript{1}\footnotemark[1] \\
    Duy A Nguyen\textsuperscript{1,2}\quad
    Tuan Anh Nguyen Pham\textsuperscript{1}\quad
    Thanh Nguyen\textsuperscript{1} \\
    Nitesh V.\ Chawla\textsuperscript{3}\quad
    Wray Buntine\textsuperscript{1,4}\quad
    Kok-Seng Wong\textsuperscript{1} \\
    Khoa D.\ Doan\textsuperscript{1}\thanks{Co-corresponding Authors: \texttt{khoa.dd@vinuni.edu.vn} and \texttt{binh.nt2@vinuni.edu.vn}.}\quad
    Binh T.\ Nguyen\textsuperscript{1}\footnotemark[2] \\[3pt]
    {\normalfont\textsuperscript{1}VinUniversity\quad
    \textsuperscript{2}University of Illinois, Urbana-Champaign} \\
    {\normalfont\textsuperscript{3}University of Notre Dame\quad
    \textsuperscript{4}Monash University}
  \end{tabular}%
  }%
}
\else
\author{Anonymous authors}
\fi

\begin{document}

\maketitle
\lhead{Preprint}

\begin{abstract}
The rapid growth in submissions to machine learning venues has strained peer review and intensified interest in LLM-based automated reviewers. Yet how well these systems catch scientific problems, compared with human reviewers, remains poorly understood: existing evaluations either reward similarity to human reviews, which are themselves incomplete, or assign holistic LLM-judge scores that cannot be traced to specific statements. We introduce \textbf{PRISM} (\textbf{P}eer \textbf{R}eview \textbf{I}ntelligence via \textbf{S}tructured \textbf{M}ulti-dimensional assessment), a benchmark that evaluates human and LLM reviews on equal terms across four capabilities: \textbf{Depth of Analysis}, \textbf{Novelty Assessment}, \textbf{Flaw Identification \& Major Issues Prioritization}, and \textbf{Multi-dimensional Constructiveness}. PRISM scores individual review statements, verifies flaws against the manuscript and novelty claims against independently retrieved literature, and measures flaw coverage against a verified pool of flaws from all reviewers rather than against human reviews alone. Applying PRISM to five automated reviewer systems and human reviews on 600 papers from ICLR, ICML and NeurIPS, we find that LLMs can match or exceed humans on individual dimensions, with higher critical-flaw recall and more actionable feedback, but no system dominates all dimensions, and only human review scores differ between accepted and rejected papers. \textit{LLM reviewers are best understood as targeted supplements to human review, not standalone replacements.}
    \ificlrfinal Our demo and key results can be found at \href{https://prism-benchmark.github.io/}{https://prism-benchmark.github.io/}.\fi
\end{abstract}

\section{Introduction}
\label{sec:intro}

Scientific peer review is under mounting strain: NeurIPS submissions grew from 15,671 (2024) to 21,575 (2025)~\cite{neurips2024, neurips2025blog}, ICML by 44.9\% and 25.4\% in consecutive years~\cite{icml2023stats, icml2024stats, icml2025stats}, and AAAI-26 received 30,948 submissions, against 15,532 for AAAI-25~\cite{aihub2026digest}. This growth outpaces the reviewer pool and prompts load-management measures such as mandatory reviewing and author self-ranking~\cite{icml2026policy_blog}. Meanwhile, Large Language Models (LLMs) have evolved from proofreading aids into autonomous reviewer agents that draft complete critiques~\cite{chang-etal-2025-treereview, gao2024reviewer2optimizingreviewgeneration, yu-etal-2024-automated-SEA,zhu-etal-2025-deepreview,cyclereviewer}, an estimated 17--21\% of reviews at recent top-tier venues already involve LLM assistance~\cite{liang2023largelanguagemodelsprovide, Wang_2024, iclr2026policy}, and AI is entering live reviewing: ICLR 2025 piloted LLM feedback on more than 20,000 reviews~\cite{thakkar2025llmfeedback}, and every AAAI-26 main-track submission received an AI review~\cite{biswas2026aaai}. Venue policies now range from outright bans to mandatory disclosure~\cite{icml2026policy}.

This reality raises an important question: \textit{Are LLMs sufficient reviewers to evaluate scientific work---and, critically, are they better at identifying gaps in a paper than human reviewers who increasingly work under time constraints and review overload?} Answering it requires care, because human review is itself an imperfect yardstick. The NeurIPS consistency experiment~\cite{beygelzimer2023neurips} suggested that as many as 23\% of acceptance decisions may change depending purely on reviewer assignment, and human reviews cover only a subset of a paper's issues and are sometimes wrong~\cite{deng2026cocoreviewbench}. Evaluators that treat human reviews as ground truth therefore inherit their omissions and errors.

We address this by introducing a benchmark that evaluates both LLM-generated and human reviews, grounded in the official reviewer guidelines of established machine learning venues (e.g., ICLR, NeurIPS). A high-quality peer review must go beyond summarization to evaluate technical soundness, contextualize originality, diagnose critical errors, and provide actionable feedback. Accordingly, our benchmark asks four questions:
\begin{itemize}[leftmargin=2.5em, itemsep=1pt, topsep=2pt]
  \item[\textbf{RQ1}] \textbf{Depth of Analysis:} Do reviewers engage with a paper's methodological and empirical claims in depth, or default to surface-level assessment?
  \item[\textbf{RQ2}] \textbf{Novelty Assessment:} Are reviewers' novelty judgments grounded in prior literature, or do they rest on unverified or incorrect assertions?
  \item[\textbf{RQ3}] \textbf{Flaw Identification \& Major Issues Prioritization:} How comprehensively do reviewers detect critical scientific flaws, and do they rank fatal methodological concerns above minor textual anomalies?
  \item[\textbf{RQ4}] \textbf{Multi-dimensional Constructiveness:} How actionable, solution-oriented, and professionally calibrated is the reviewers' feedback?
\end{itemize}

We call this benchmark \textbf{PRISM} (\textbf{P}eer \textbf{R}eview \textbf{I}ntelligence via \textbf{S}tructured \textbf{M}ulti-dimensional assessment). PRISM evaluates human and LLM reviews on equal terms. It scores individual review statements rather than whole reviews, so every score traces to specific sentences. Flaws are verified against the manuscript, and novelty claims against independently retrieved literature. Flaw coverage is measured against a verified pool of flaws from all reviewers, not against the human reviews alone. Each dimension follows an established construct: depth as argument structure and grounding~\citep{hua-etal-2019-argument}, novelty as support relative to prior literature~\citep{novelty1,novelty2}, prioritization as a graded ranking measure~\citep{NDCG}, and constructiveness as comment-level attributes of peer-review discourse~\citep{kennard-etal-2022-disapere}.

We apply \ours{} to five automated reviewer systems---TreeReview~\citep{chang-etal-2025-treereview}, Reviewer2~\citep{gao2024reviewer2optimizingreviewgeneration}, SEA-E~\citep{yu-etal-2024-automated-SEA}, DeepReview~\citep{zhu-etal-2025-deepreview}, and CycleReviewer~\citep{cyclereviewer}---and to human reviews on 600 papers from ICLR, ICML, and NeurIPS. Our results (\S\ref{sec:results}, Figure~\ref{fig:overall_result}) show that LLM reviewers match or exceed humans on individual dimensions---DeepReview approaching human analytical depth, SEA-E in grounded novelty verification, Reviewer2 in flaw recall, and DeepReview in constructiveness---yet no single system dominates all four, and only human review scores track the eventual accept/reject decision (\S\ref{sec:accept_reject}). In summary, the key contributions of this work are:
\begin{itemize}[leftmargin=1.75em, itemsep=1pt, topsep=2pt]
    \item \textbf{PRISM, an evidence-grounded benchmark that evaluates human and LLM reviews on equal terms.} Statement-level units, verification against the manuscript and the literature, and a pooled verified-flaw reference replace similarity to human references and holistic judging; each of the four pipelines operationalizes an established construct.
    \item \textbf{A systematic human-vs-LLM analysis.} We benchmark five leading LLM reviewer systems against human reviewers on all four dimensions, validate the LLM judge against expert annotators, and reveal distinct specialization profiles and failure modes invisible to aggregate metrics.
    \item \textbf{Actionable deployment guidance.} We identify which systems best fit which roles within a human-assisted review pipeline.
\end{itemize}

\section{Related work}
\label{sec:related_work}
\paragraph{LLM-based reviewer systems.}
Structured-reasoning reviewers decompose evaluation into a tree of questions (TreeReview~\citep{chang-etal-2025-treereview}) or emulate slow, deliberate expert thinking with retrieval (DeepReview~\citep{zhu-etal-2025-deepreview}); generation-centric systems condition reviews on predicted aspects (Reviewer2~\citep{gao2024reviewer2optimizingreviewgeneration}), fine-tune on standardized review data (SEA~\citep{yu-etal-2024-automated-SEA}), or train a reviewer in an iterative preference loop (CycleReviewer~\citep{cyclereviewer}). Newer systems add multi-objective reinforcement learning (REMOR~\citep{taechoyotin2025remor}), search-enabled multi-agent reviewing (ScholarPeer~\citep{goyal2026scholarpeer}), and rubric-guided tool use (ReviewGrounder~\citep{li-etal-2026-reviewgrounder}); PRISM can benchmark all of them as systems.

\paragraph{Evaluating reviews.}
Existing evaluators differ in what they check. \emph{Overlap metrics} (ROUGE~\citep{lin-2004-rouge}, BLEU~\citep{papineni-etal-2002-bleu}, BERTScore) reward similarity to reference reviews~\citep{novikova-etal-2017-need}; pairing a review with references from a \emph{different} paper lowers BERTScore by only 1.79--2.63 points~\citep[App.~C.1, Table~7]{chang-etal-2025-treereview}. \emph{Holistic LLM judges}~\citep{liu-etal-2023-g,zheng2023judging} are biased toward position~\citep{zheng2023large}, verbosity~\citep{saito2023verbosity} and their own generations~\citep{panickssery2024llm}, locate deficient review segments poorly (about 20 F1;~\citealp{du-etal-2024-llms}), and align poorly with expert ratings of review quality~\citep{ebrahimi2025rottenreviews}. \emph{Rubric and structured evaluators} score reviews along several dimensions: ReviewEval~\citep{garg-etal-2025-revieweval}, ReviewBench~\citep{li-etal-2026-reviewgrounder}, RevUtil~\citep{sadallah-etal-2025-good}, PeeriScope~\citep{ebrahimi2026peeriscope}, and the focus-level framework of \citet{focuslevel2025}; \citet{dycke2026automatic} target reasoning faults only. \emph{Reference-based correctness} is closest to PRISM: CoCoReviewBench~\citep{deng2026cocoreviewbench} filters human references for correctness but credits only issues raised by humans, and scores a human review only against the other human reviews of the same paper. PeerPrism~\citep{sadeghian2026peerprism} addresses a different task, AI-authorship detection, not review quality. Table~\ref{tab:capability} summarizes these differences; App.~\ref{app:ext_related} gives an extended comparison.

\paragraph{Methodological precedent.}
Decomposing text into atomic units and checking each unit against evidence is an established evaluation methodology: FActScore~\citep{min-etal-2023-factscore} verifies atomic facts against a knowledge source, and SAFE~\citep{wei2024longform} against search results. PRISM applies this principle to peer review: flaws are verified against the manuscript and novelty claims against retrieved prior work.

\begin{table}[t]
\centering\footnotesize
\caption{Capabilities of review evaluators. \cmark: yes; \pmark: partially; \xmark: no.}
\label{tab:capability}
\begin{tblr}{
  width=\linewidth,
  colspec={X[2.3,l] X[c] X[c] X[c] X[c] X[c]},
  colsep=3pt, rowsep=0.5pt,
  row{1}={font=\bfseries\scriptsize},
  hline{1,Z}={0.08em}, hline{2}={0.05em}, hline{Z-1}={dashed},
}
Evaluator & Checks claims against manuscript & Checks novelty against literature & Credits valid issues no human raised & Human--AI review comparison on equal terms & Traceable to review units \\
Overlap metrics (ROUGE, BLEU, BERTScore) & \xmark & \xmark & \xmark & \xmark & \xmark \\
Generic holistic / pairwise LLM judges & \xmark & \pmark & \xmark & \xmark & \xmark \\
ReviewEval~\citep{garg-etal-2025-revieweval} & \cmark & \xmark & \pmark & \pmark & \pmark \\
ReviewBench~\citep{li-etal-2026-reviewgrounder} & \pmark & \xmark & \pmark & \xmark & \xmark \\
CoCoReviewBench~\citep{deng2026cocoreviewbench} & \xmark & \xmark & \xmark & \xmark & \cmark \\
RevUtil~\citep{sadallah-etal-2025-good} & \xmark & \xmark & \xmark & \cmark & \cmark \\
PeeriScope / RottenReviews & \xmark & \xmark & \xmark & \xmark & \xmark \\
\textbf{PRISM (ours)} & \cmark & \cmark & \cmark & \cmark & \cmark \\
\end{tblr}
\vspace{-0.3cm}
\end{table}

\section{The PRISM Framework}
\label{sec:method}

\begin{figure}[t]
\centering
\includegraphics[width=\linewidth]{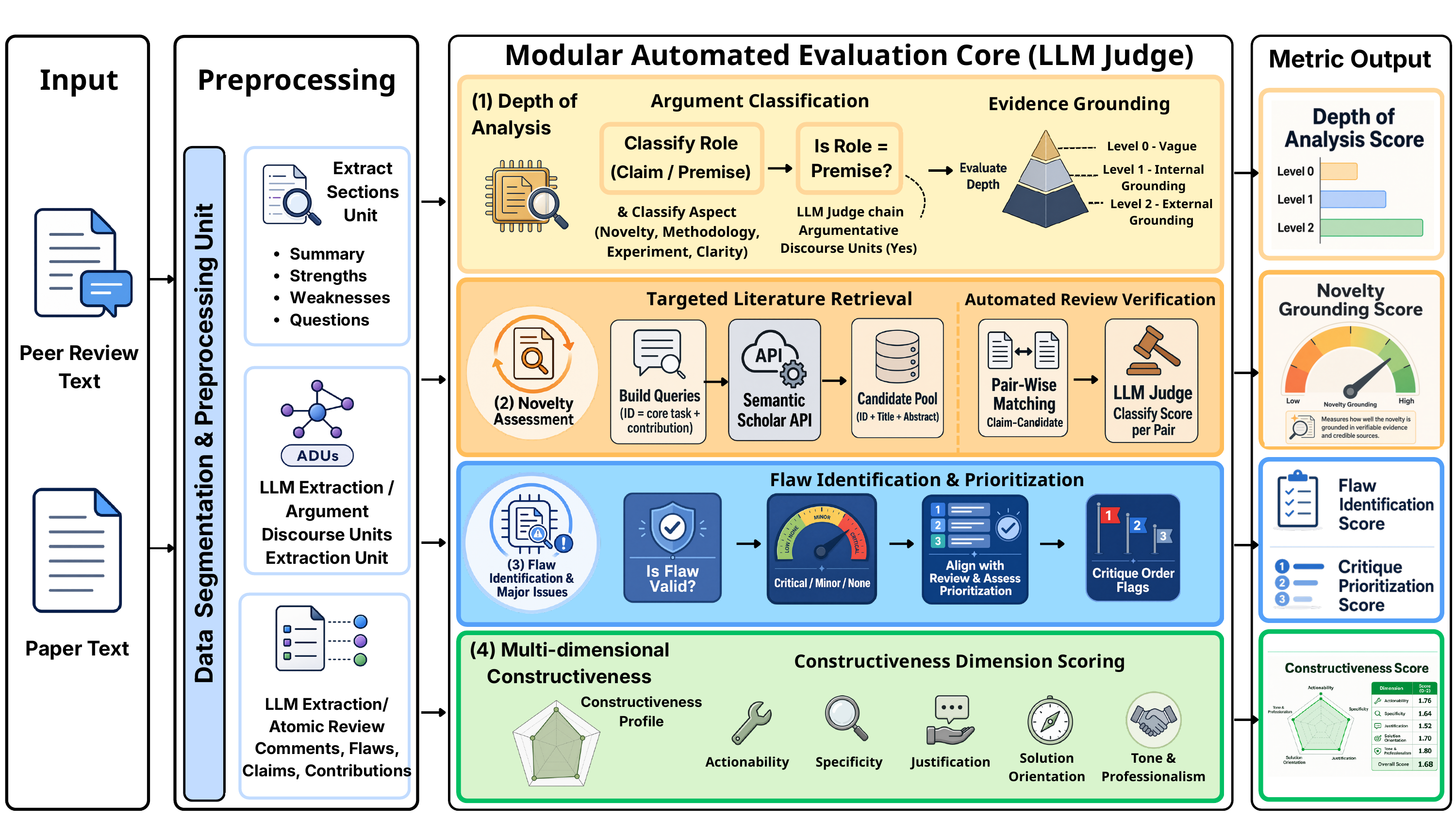}
\vspace{-0.55cm}
\caption{\textbf{Comprehensive overview of PRISM.} Reviews and manuscripts are segmented, then scored by four modular LLM-driven pipelines (\S\ref{subsec:doa}--\ref{subsec:mcs}), yielding a four-metric evaluation profile. \emph{ADU}: argumentative discourse unit, a claim or a premise supporting it. \emph{Grounding level}: 0 vague/generic, 1 refers to the manuscript, 2 refers to external literature. \emph{Pooled verified-flaw reference}: the union of flaws raised by any human or LLM review of the paper, de-duplicated and verified one by one against the manuscript. \emph{Critique Prioritization Score} (nCPS): an NDCG-style measure of whether Critical flaws precede Minor ones. \emph{ARC}: atomic review comment, the smallest independent unit of critique or suggestion.}
\label{fig:overall_flow}
\vspace{-0.3cm}
\end{figure}

PRISM rests on four design choices. (i)~\emph{Statement-level units}: each pipeline extracts discrete units from the review (argument units, novelty claims, flaw arguments, atomic comments) instead of rating the review as a whole. (ii)~\emph{Evidence-based verification}: flaws are verified against the manuscript, and novelty claims against independently retrieved literature. (iii)~\emph{A pooled verified reference}: flaw coverage is measured against the verified flaws raised by \emph{all} reviewers, human and LLM. (iv)~\emph{Established constructs}: each dimension operationalizes a construct from prior work on peer review. Table~\ref{tab:glance} summarizes the four dimensions and Figure~\ref{fig:overall_flow} the pipelines; \S\ref{subsec:doa}--\ref{subsec:mcs} give the formulas.

\begin{table}[t]
\centering\footnotesize
\caption{PRISM at a glance. All metrics lie in $[0,1]$; higher is better.}
\label{tab:glance}
\begin{tblr}{
  width=\linewidth,
  colspec={X[1.2,l] X[1.25,l] X[1.6,l] X[1.35,l] X[1.7,l]},
  colsep=3pt, rowsep=1pt,
  row{1}={font=\bfseries},
  hline{1,Z}={0.08em}, hline{2}={0.05em},
}
Dimension & Unit scored & Checked against & Metric & Meaning \\
Depth of Analysis & Argumentative discourse unit (claim / premise) & Premise grounding: 0 vague, 1 manuscript, 2 external literature & DoA = harmonic mean of premise ratio and normalized grounding & How much of the review is supporting evidence, and how well grounded it is \\
Novelty Assessment & Novelty claim & Prior literature from Semantic Scholar (published no later than the submission year) & NS $=(\bar{S}+2)/4$; SR, SSR & How well the review's novelty claims are supported by prior work \\
Flaw Identification & Flaw argument $\rightarrow$ pooled flaw & The manuscript (validity); severity labels & Critical / Minor recall & Share of verified flaws the review catches \\
Prioritization & Order of valid flaws in the review & Severity (Critical $=2$, Minor $=1$) & nCPS $=$ CPS / iCPS & Whether critical flaws are raised before minor ones \\
Constructiveness & Atomic review comment (ARC) & Rubric D1--D5 (0--2 each) & MCS $=$ mean comment score & Actionability, specificity, justification, solution, tone \\
\end{tblr}
\end{table}

\noindent The LLM judge only extracts and labels units; all scores are computed analytically from these labels.

\subsection{Depth of Analysis}
\label{subsec:doa}
A high-quality review backs its critical claims with substantive evidence~\citep{hua-etal-2019-argument}. We define \textit{Depth of Analysis} (DoA) as the degree to which a reviewer substantiates judgments with well-grounded premises.

\textbf{Pipeline.}
We break the core review sections (Summary, Strengths, Weaknesses) into Argumentative Discourse Units (ADUs)~\citep{peldzusadu}, classify each ADU by \textit{role} (\textit{Claim} or \textit{Premise}) and \textit{aspect} (Novelty, Methodology, Experiments, Clarity), and assess the \textit{grounding level} $g(p) \in \{0,1,2\}$ of each premise (Table~\ref{tab:glance}).

\textbf{Score Formulation.}
With $A$ the ADUs, $P \subseteq A$ the premises and $g_{\max}=2$, the \textit{Premise Ratio} is $R_{\mathrm{prem}} = |P|/|A|$ and the normalized grounding is $S_{\mathrm{depth}} = \frac{1}{g_{\max}|P|}\sum_{p \in P}g(p)$. DoA is their harmonic mean,
$\mathrm{DoA} = \frac{2 R_{\mathrm{prem}} S_{\mathrm{depth}}}{R_{\mathrm{prem}} + S_{\mathrm{depth}}}$,
so a review must excel in both the \emph{proportion} and the \emph{grounding} of its evidence; DoA\,$=0$ if $|P|=0$. Aspect labels do not enter the score; they show where reviewers direct their effort.

\subsection{Novelty Assessment}
\label{subsec:novelty}
Novelty is the degree to which a paper introduces non-trivial findings relative to existing knowledge~\citep{novelty1,novelty2,novelty3}, so a genuine novelty judgment must situate the paper within the prior literature. Our pipeline verifies whether a reviewer's novelty claims are supported or refuted by retrievable prior work~\citep{zhang2026opennovelty}.

\textbf{Pipeline.}
\textbf{\textit{Extraction}}: an LLM extracts the paper's core task, contribution anchors and key terms, and the verbatim novelty claims $\mathcal{C} = \{c_1,\ldots,c_n\}$ from the review. \textbf{\textit{Retrieval}}: deterministic Semantic Scholar queries built from the anchors return candidates that are restricted to papers \emph{published no later than the submission year} of the reviewed paper, de-duplicated, and diversified by Maximal Marginal Relevance into a pool $\mathcal{B} = \{b_1,\ldots,b_k\}$. Later publications, corrections or critiques of prior work are therefore not used as evidence: a review is judged against the literature its author could have known. This retrieval is independent of any sources a reviewer system may have used. \textbf{\textit{Verification}}: for each pair $(c_i, b_j)$, the judge compares the claim with the paper context (abstract + introduction) and the candidate (title + abstract), and returns an evidence-support score $s(c_i, b_j) \in \{-2,-1,0,+1,+2\}$: Unsupported, Understated, Ambiguous, Overstated, Supported.

\textbf{Score Formulation.}
To avoid optimistic inflation from a single spurious match, each claim aggregates its relevance-weighted top-3 candidates $\mathcal{T}_i$ with retrieval relevance $r_j$:
$S_{\mathrm{claim}}(c_i) = \sum_{j \in \mathcal{T}_i} s(c_i,b_j)\,r_j / \sum_{j \in \mathcal{T}_i} r_j$.
With $\bar{S}$ the mean claim score of a review $R$,
$NS(R) = (\bar{S}+2)/4$,
$SR(R) = |\{c_i : S_{\mathrm{claim}}(c_i) \ge 1\}|/n$ and
$SSR(R) = |\{c_i : S_{\mathrm{claim}}(c_i) = 2\}|/n$
give the overall score and the fractions of claims with partial and strict literature support.

\subsection{Flaw Identification \& Major Issues Prioritization}
\label{subsec:flaw}
Effective review requires both accurate diagnosis of scientific errors and a clear ordering of them. Because the number of flaws in a manuscript is unobservable, we build a \emph{pooled verified-flaw reference}: the union of all flaws raised in any human or LLM review of the paper, de-duplicated by merging arguments that describe the same problem, with each flaw verified individually against the manuscript. Following graded-relevance ranking evaluation (NDCG;~\citealp{NDCG}), we treat burying critical flaws beneath minor ones as a failure of review quality.

\textbf{Pipeline.}
\textbf{\textit{Extraction}}: an LLM extracts distinct flaw arguments from the Summary, Weaknesses and Questions of every review of the paper. \textbf{\textit{Verification and pooling}}: reading the paper, the judge (here acting as the \emph{verifier}) labels each flaw valid or invalid, merges valid flaws into the pooled reference, and assigns each pooled flaw a severity, \textit{Critical} (e.g., methodological errors, flawed proofs) or \textit{Minor} (e.g., typos, formatting). No agreement among reviewers is required: a flaw raised by a single reviewer is retained if the manuscript supports it, and the verifier is not told how many reviewers raised it. \textbf{\textit{Positional recovery}}: valid flaws are mapped back to their position in the review.

\textbf{Score Formulation.}
With $F_{\mathrm{true}}^{C}, F_{\mathrm{true}}^{M}$ the pooled Critical and Minor flaws and $F_{\mathrm{rev}}^{C}, F_{\mathrm{rev}}^{M}$ those the reviewer identified, \textbf{coverage} is the severity-stratified recall
$|F_{\mathrm{true}}^{C/M} \cap F_{\mathrm{rev}}^{C/M}| / |F_{\mathrm{true}}^{C/M}|$.
\textbf{Ordering} is measured by the normalized Critique Prioritization Score, an NDCG-style score: with $p_i$ the position of the $i$-th valid flaw in the review and $w_i$ its severity weight,
$nCPS = CPS/iCPS$, $CPS = \sum_{i=1}^{k} w_i/\log_2(p_i + 1)$,
where $iCPS$ is the score of the ideal order (all Critical flaws first). A Critical flaw counts twice as much as a Minor one ($w_i = 2$ vs.\ $1$), so nCPS rewards raising Critical flaws early and equals 1 when all Critical flaws precede all Minor flaws. For example, a review raising four valid flaws in the order Minor, Critical, Minor, Critical has $CPS = \tfrac{1}{\log_2 2} + \tfrac{2}{\log_2 3} + \tfrac{1}{\log_2 4} + \tfrac{2}{\log_2 5} = 3.623$, the ideal order gives $iCPS = 4.193$, and $nCPS \approx 0.864$ (App.~\ref{sec:app_prism}).

\textbf{Hallucination detection.} PRISM verifies what reviewers assert instead of taking it at face value. A flaw is counted as \emph{hallucinated} when the verifier, reading the manuscript, labels it invalid: fabricated, a misreading of the paper, or an unreasonable request (Fig.~\ref{fig:flaw_prompt_phase2}). Hallucinated flaws are excluded from the pooled reference, earn no recall credit, and are reported separately (Fig.~\ref{fig:comparison_flaws}). Unsupported novelty claims are handled by the novelty pipeline instead (\S\ref{subsec:novelty}): they remain in the review and lower its NS through negative support scores.

\subsection{Multi-Dimensional Constructiveness}
\label{subsec:mcs}
A review's value also lies in helping authors improve. Grounded in discourse taxonomies such as DISAPERE~\citep{kennard-etal-2022-disapere}, we decompose constructiveness into informational and social dimensions.

\textbf{Pipeline.} An LLM breaks the review into Atomic Review Comments (ARCs) and rates each ARC $c_j$ from 0 to 2 on five dimensions: \textbf{Actionability ($D_1$)}, clear, implementable guidance; \textbf{Specificity ($D_2$)}, pinpoints concrete elements such as sections or equations; \textbf{Justification ($D_3$)}, backed by reasoning or evidence; \textbf{Solution ($D_4$)}, proposes a path to improvement; \textbf{Tone ($D_5$)}, professional and encouraging, penalizing hostility that demoralizes authors without improving the science~\citep{hyland2020antithetical, rao2022civility}.

\textbf{Score Formulation.}
$CLC(c_j) = \frac{1}{10}\sum_{k=1}^{5} D_k(c_j) \in [0,1]$ and $MCS(R) = \frac{1}{n}\sum_{j=1}^{n} CLC(c_j)$, so a perfect $MCS$ requires every comment to be specific, justified, actionable and professionally toned.

\section{Experiments}
\label{sec:experiment}

\subsection{Evaluation Setting}
\label{sec:setting}

\textbf{Dataset.} PRISM is evaluated on 200 manuscripts per venue-year from \textbf{ICLR 2026}, \textbf{ICML 2025}, and \textbf{NeurIPS 2025} (600 papers; per-decision counts in Table~\ref{tab:data_selection_stats}, App.~\ref{sec:app_dataset}). We use the most recently concluded venue-years, so that manuscripts and reviews were released after the reported training cutoffs of the reviewer systems and the judge where such information is available, reducing contamination. Sampling is stratified by decision (\textit{Reject}, \textit{Poster}, \textit{Spotlight}, \textit{Oral}) and preserves each venue's score distribution.

\textbf{Reviewers.} We evaluate five automated reviewer systems spanning \emph{supervised fine-tuning} (SEA-E~\citep{yu-etal-2024-automated-SEA}, CycleReviewer~\citep{cyclereviewer}, DeepReview~\citep{zhu-etal-2025-deepreview}) and \emph{prompting} (Reviewer2~\citep{gao2024reviewer2optimizingreviewgeneration}, TreeReview~\citep{chang-etal-2025-treereview}); configurations are in App.~\ref{sec:app_baseline}. Each system receives the full text of the paper under review. Only DeepReview retrieves literature (OpenScholar; App.~\ref{sec:app_baseline}), restricted to scholarly literature available before the manuscript's submission date and excluding conference reviews, author--reviewer discussions and decisions; the other systems rely on the paper and on parametric knowledge. Our human baseline is \textbf{human reviews}: the reviews submitted to ICLR 2026, ICML 2025 and NeurIPS 2025 under each venue's LLM policy. Some may involve AI assistance (an estimated 17--21\% of reviews at recent venues; \S\ref{sec:intro}), but each was submitted under a human reviewer's responsibility. We additionally evaluate two general-purpose LLMs, Qwen3.6-27B and DeepSeek-V4.1-Flash, used as naive single-prompt reviewers (full manuscript, no tools or refinement) on a 60-paper subset (\S\ref{sec:naive}, App.~\ref{app:naive}).

\textbf{Judge.} The primary judge is Gemini~2.5 Flash Lite~\citep{gemini2025team} for all extraction and scoring (temperature 0; prompts in App.~\ref{sec:app_prism}); \S\ref{sec:robustness} adds two judges from other model families.

\subsection{Validating PRISM}
\label{sec:validation}
Because PRISM decomposes each review into units scored by individual judgments, we validate each judgment and each pipeline stage directly, then test whether the system-level conclusions survive changes of judge and prompt.

\begin{table}[t]
\begin{minipage}[t]{0.49\linewidth}\vspace{0pt}
\centering\footnotesize
\captionof{table}{Expert vs.\ Gemini 2.5 Flash Lite agreement on 30 papers ($^{\dagger}$within-one on ordinal scales).}
\label{tab:human_validation}
\begin{tblr}{
  width=\linewidth,
  colspec={X[l] c c},
  colsep=2.5pt, rowsep=0.5pt,
  row{1}={font=\bfseries},
  hline{1,Z}={0.08em}, hline{2}={0.05em},
  hline{4,5,8}={dashed},
}
Task (dimension) & $\kappa$ / $\kappa_w$ & Agreement \\
Aspect ID (DoA) & 0.699 & 77.9\% \\
Grounding (DoA) & 0.482 & 98.4\%$^{\dagger}$ \\
Novelty score & 0.345 & 74.8\%$^{\dagger}$ \\
Aspect class.\ (Flaw) & 0.514 & 63.5\% \\
Severity (Flaw) & 0.347 & 100\%$^{\dagger}$ \\
Validity (Flaw) & 0.875 (F1) & 76.8\% \\
D1--D5 (Constr.) & 0.14--0.52 & 89--99\%$^{\dagger}$ \\
\end{tblr}
\end{minipage}\hfill
\begin{minipage}[t]{0.48\linewidth}\vspace{0pt}
\centering\footnotesize
\captionof{table}{Stage-wise extraction audit: precision of extracted units and share of reviews with $\ge$1 missed important unit.}
\label{tab:audit}
\begin{tblr}{
  width=\linewidth,
  colspec={X[l] c c},
  colsep=2.5pt, rowsep=0.5pt,
  row{1}={font=\bfseries},
  hline{1,Z}={0.08em}, hline{2}={0.05em},
}
Stage & Precision [95\% CI] & $\ge$1 omission \\
Flaw arguments & 90\% [85, 95] & 25\% \\
Constr.\ ARCs & 90\% [86, 93] & 37\% \\
DoA ADUs & 85\% [75, 93] & 37\% \\
Novelty claims & 42\% [32, 55] & 35\% \\
\end{tblr}
\end{minipage}
\vspace{-0.3cm}
\end{table}

\subsubsection{Agreement with experts}
\label{sec:human_validation}
Twenty experienced researchers (doctoral researchers, senior researchers and professors who review for ICLR, ICML and NeurIPS) re-scored the pre-extracted units (ADUs, novelty claims, flaw arguments, ARCs) of 30 papers, using the same definitions and scales as the judge (protocol in App.~\ref{sec:app_human_validation}). Agreement is substantial on the more objective tasks (Aspect ID $\kappa=0.699$; Grounding $\kappa_w=0.482$, 98.4\% within one; flaw validity F1 $=0.875$; Table~\ref{tab:human_validation}). Disagreement concentrates in the inherently subjective dimensions (novelty, constructiveness, severity), where expert reviewers also disagree~\citep{cortes2021inconsistency,beygelzimer2023neurips}. A manual cause analysis of 370 stratified disagreements shows that replacing judge labels with expert labels leaves the main system comparisons unchanged; the exception is the DoA ranking, where one system's numbered review format is segmented differently (App.~\ref{app:audit}). Single-reviewer flaws are not penalized: 83.6\% of human-raised pooled flaws come from a single reviewer, and the verifier keeps 94.2\% of them vs.\ 96.4\% of multi-reviewer flaws, including 50 of the 53 single-reviewer flaws that experts marked as valid Critical (App.~\ref{app:audit}).

\subsubsection{Stage-wise audit of extraction and retrieval}
\label{sec:audit}
Researchers from the expert study also audited the extraction and retrieval stages, using the same definitions (Table~\ref{tab:audit}). Novelty-claim extraction is the weak extraction stage; its errors are mostly statements that are not novelty judgments rather than missed claims. Retrieval is the weakest stage: the pool recovers 34\% of the works cited by reviewers, evidence is judged sufficient for 17\% of claims, and the reviewed paper appears in its own pool for 56--70\% of reviews; removing it lowers NS by 0.064--0.074 for every system, with the ranking unchanged except one near-tied pair. \emph{Effect on final scores}: after removing the units marked invalid, Flaw and Constructiveness scores are unaffected (per-review shift $\le 0.017$, Kendall $\tau = 1.00$); DoA shifts by 0.029 ($\tau = 0.73$); NS levels shift by 0.06 and its ranking is not stable ($\tau = 0.33$). We therefore read NS comparisons as grounding \emph{within the retrieved literature} (\S\ref{sec:exp_novelty}).

\subsubsection{Robustness to the judge and to the prompt}
\label{sec:robustness}
\begin{table}[t]
\centering\footnotesize
\caption{Robustness of system rankings. \emph{Judges}: Kendall's $W$ over the 5 automated systems ranked by Gemini 2.5 Flash Lite, Claude Haiku 5.5 and GPT-5.6-Sol. \emph{Prompts}: Kendall's $\tau$ between the ranking of all 6 reviewer types under the original prompts and under paraphrased (P1) or definition-only (P2) prompts; the noise floor re-runs the original prompts. Setups in App.~\ref{app:judges} and~\ref{app:prompt}.}
\label{tab:robustness}
\begin{tblr}{
  colspec={l c c c},
  colsep=6pt, rowsep=0.5pt,
  row{1}={font=\bfseries},
  hline{1,Z}={0.08em}, hline{2}={0.05em},
}
Metric & $W$ across 3 judges & $\tau$ vs.\ original (P1 / P2) & $\tau$ noise floor \\
Critical recall & 0.87 & 0.867 / 0.867 & 0.600 \\
Minor recall & 0.96 & 0.733 / 1.000 & 0.867 \\
Constructiveness (MCS) & 0.82 & 0.867 / 0.600 & 0.867 \\
Novelty (NS) & 0.80 & 0.200 / 0.733 & $-0.200$ \\
Depth (DoA) & 0.78 & 0.867 / 0.067 & 0.067 \\
Prioritization (nCPS) & 0.61 & 0.200 / 0.733 & 0.067 \\
\end{tblr}
\vspace{-0.3cm}
\end{table}
\textbf{Judges.} Gemini 2.5 Flash Lite, Claude Haiku 5.5 and GPT-5.6-Sol each run the full pipeline (Table~\ref{tab:robustness}; per-system scores in App.~\ref{app:judges}). The leading system is unchanged for Critical and Minor recall (Reviewer2), Constructiveness (DeepReview) and Novelty (SEA). DoA and prioritization are more judge-sensitive: DoA because the judges weigh premise share and grounding differently, prioritization because nCPS is near ceiling (0.97--0.99) and differs mainly through flaw-validity decisions.
\textbf{Prompts.} We re-ran all pipelines with (P1) paraphrased prompts with identical content and (P2) definition-only prompts without examples (App.~\ref{app:prompt}). Flaw results agree with the original at the run-to-run noise floor (per-review Spearman $\rho$ 0.700/0.763 for Critical and 0.668/0.707 for Minor recall vs.\ floors 0.680/0.691), and extraction overlap matches the noise floor (flaws 0.751/0.680 vs.\ 0.683; ARCs 0.629/0.618 vs.\ 0.626; novelty claims 0.770/0.733, stance changed in 8\%/7\%). Across all metrics on 30 papers, only one rank swap involves a pair separated under the original prompts (Human/Reviewer2 on nCPS under P1). DoA is the most prompt-dependent axis: removing the grounding anchor examples raises DoA for every system (+0.054 to +0.170).

Together, these results show that PRISM's conclusions on flaw identification, constructiveness and novelty leadership are robust, while DoA and prioritization are more sensitive; we therefore qualify the corresponding statements in \S\ref{sec:results}.

\subsection{Results: LLM Reviewers vs.\ Human Reviews}
\label{sec:results}

\begin{table}[t]
\centering
\caption{Macro-average PRISM scores (mean $\pm$ std) across ICLR 2026, ICML 2025 and NeurIPS 2025 (600 papers; primary judge).}
\label{tab:main_results}
\resizebox{\linewidth}{!}{%
\begin{tblr}{
  column{1} = {l},
  column{2-7} = {c},
  hline{1,9} = {-}{0.08em},
  hline{3} = {-}{0.05em},
  hline{4} = {-}{dashed},
}
\SetCell[r=2]{l}\textbf{Reviewer} & \SetCell[r=2]{c}\textbf{Depth of Analysis} & \SetCell[r=2]{c}\textbf{Novelty Assessment} & \SetCell[c=2]{c}\textbf{Flaw Identification} & & \SetCell[r=2]{c}{\textbf{Prioritization}} & \SetCell[r=2]{c}{\textbf{Constructiveness}} \\
 & & & \textbf{Critical} & \textbf{Minor} & & \\
\textbf{Human} & $0.493 \pm 0.062$ & $0.802 \pm 0.195$ & $0.311 \pm 0.158$ & $0.253 \pm 0.080$ & $0.975 \pm 0.043$ & $0.556 \pm 0.072$ \\
CycleReviewer    & $\mathbf{0.483} \pm 0.135$ & $0.789 \pm 0.217$ & $0.223 \pm 0.294$ & $0.179 \pm 0.141$ & $0.969 \pm 0.114$ & $0.530 \pm 0.112$ \\
DeepReview     & $0.481 \pm 0.132$ & $0.769 \pm 0.206$ & $0.349 \pm 0.325$ & $0.231 \pm 0.143$ & $0.967 \pm 0.076$ & $\mathbf{0.636} \pm 0.087$ \\
Reviewer2      & $0.363 \pm 0.137$ & $0.800 \pm 0.214$ & $\mathbf{0.644} \pm 0.304$ & $\mathbf{0.470} \pm 0.181$ & $0.972 \pm 0.052$ & $0.586 \pm 0.096$ \\
SEA            & $0.397 \pm 0.151$ & $\mathbf{0.854} \pm 0.188$ & $0.197 \pm 0.252$ & $0.208 \pm 0.118$ & $\mathbf{0.984} \pm 0.055$ & $0.491 \pm 0.085$ \\
TreeReview     & $0.339 \pm 0.116$ & $0.813 \pm 0.209$ & $0.245 \pm 0.302$ & $0.316 \pm 0.142$ & $0.977 \pm 0.044$ & $0.495 \pm 0.130$ \\
\end{tblr}
}
\vspace{-0.3cm}
\end{table}

\begin{figure}[t]
\begin{minipage}[b]{0.40\linewidth}
\centering
\includegraphics[width=\linewidth]{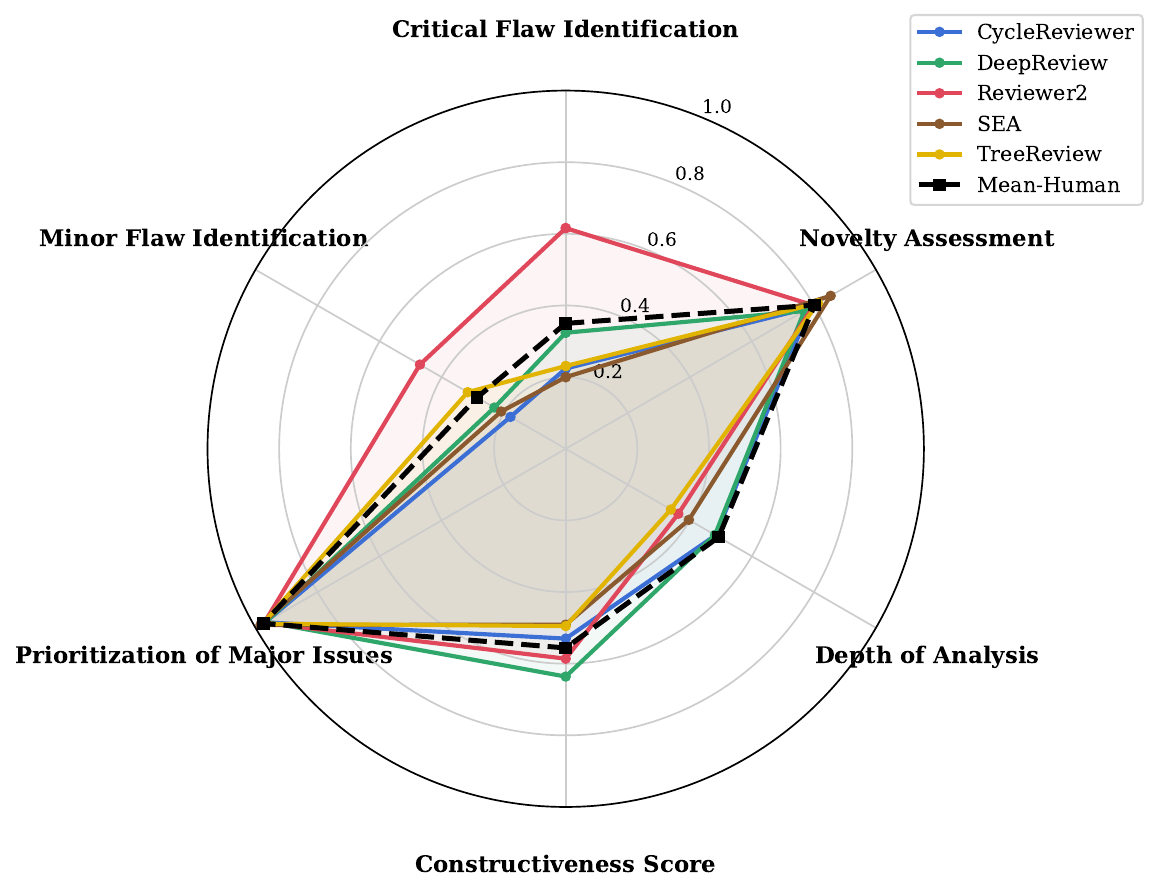}
\vspace{-0.55cm}
\captionof{figure}{Macro-averaged PRISM profiles of the five LLM reviewers and human reviews (Table~\ref{tab:main_results}).}
\label{fig:overall_result}
\end{minipage}\hfill
\begin{minipage}[b]{0.56\linewidth}
\centering
\includegraphics[width=\linewidth]{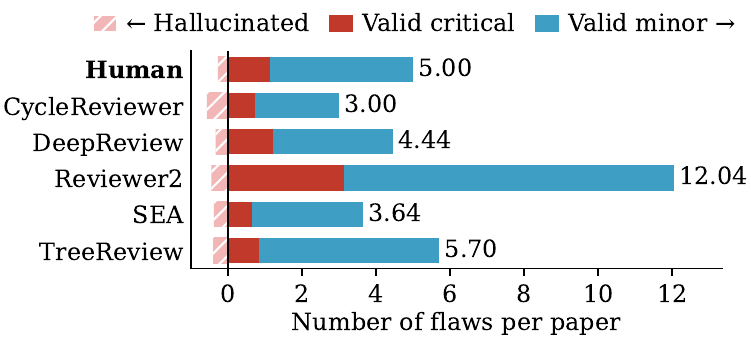}
\vspace{-0.55cm}
\captionof{figure}{Valid and hallucinated flaws per paper. Hallucinated = flaws labelled invalid by the verifier; severity is assigned only to valid flaws.}
\label{fig:comparison_flaws}
\end{minipage}
\vspace{-0.3cm}
\end{figure}

Table~\ref{tab:main_results} reports macro-averaged PRISM scores and Figure~\ref{fig:overall_result} the resulting profiles; the subsections below unpack each dimension. Extended breakdowns are in Apps.~\ref{sec:app_exp1}--\ref{app:app_exp2} and qualitative examples in App.~\ref{app:app_exp3}.

\subsubsection{Depth of Analysis}
\label{sec:exp_doa}
Human reviews rank at or near the top on DoA under all three judges (\S\ref{sec:robustness}; $0.493$ under the primary judge). Under the primary judge, \textbf{DeepReview} ($0.481$) and \textbf{CycleReviewer} ($0.483$) closely match this level, driven by a high \textit{Premise Ratio} (CycleReviewer $0.614$, DeepReview $0.588$, human $0.557$) that compensates for slightly weaker grounding; TreeReview ($0.302$) and Reviewer2 ($0.376$) fall short on premise ratio (Table~\ref{tab:doa_core}, App.~\ref{sec:detail_doa}). Under the primary judge, grounding scores are similar across systems ($0.432$--$0.481$); the other two judges separate systems more by grounding quality (App.~\ref{app:judges}). Aspect distributions (App.~\ref{sec:detail_doa}, Fig.~\ref{fig:doa_aspect}) show that most systems mirror human focus on Methodology and Experimental Design, whereas TreeReview spends $\sim 26\%$ of its effort on formatting and clarity, a shift recently observed in in-the-wild LLM reviews~\citep{llmsurfacebias}.

\textit{\textbf{Key Insight:} Human depth rests on a high Premise Ratio and a focus on methodology; the best LLMs approach it by generating abundant premises.}

\subsubsection{Novelty Assessment}
\label{sec:exp_novelty}
All systems score within $0.769$--$0.854$ on NS (Table~\ref{tab:main_results}). NS does not certify a manuscript's objective novelty: it measures how well the claims a reviewer chose to make are grounded \emph{within the retrieved literature}. Within this scope, \textbf{SEA} scores highest ($\mathbf{0.854}$, vs.\ $0.802$ for human reviews), and SEA leads under all three judges (\S\ref{sec:robustness}). Systems diverge in stance (App.~\ref{sec:detailed_novelty}, Fig.~\ref{fig:novelty_a}): SEA calls the contribution \emph{Novel} in 83\% of claims, against 62\% for human reviews, reflecting a tendency to agree with authors, whereas DeepReview is the most skeptical (41\% \emph{Novel}, 21\% \emph{Not novel}, 35\% \emph{Unclear}).

\begin{figure}[t]
\centering
\includegraphics[width=0.78\linewidth]{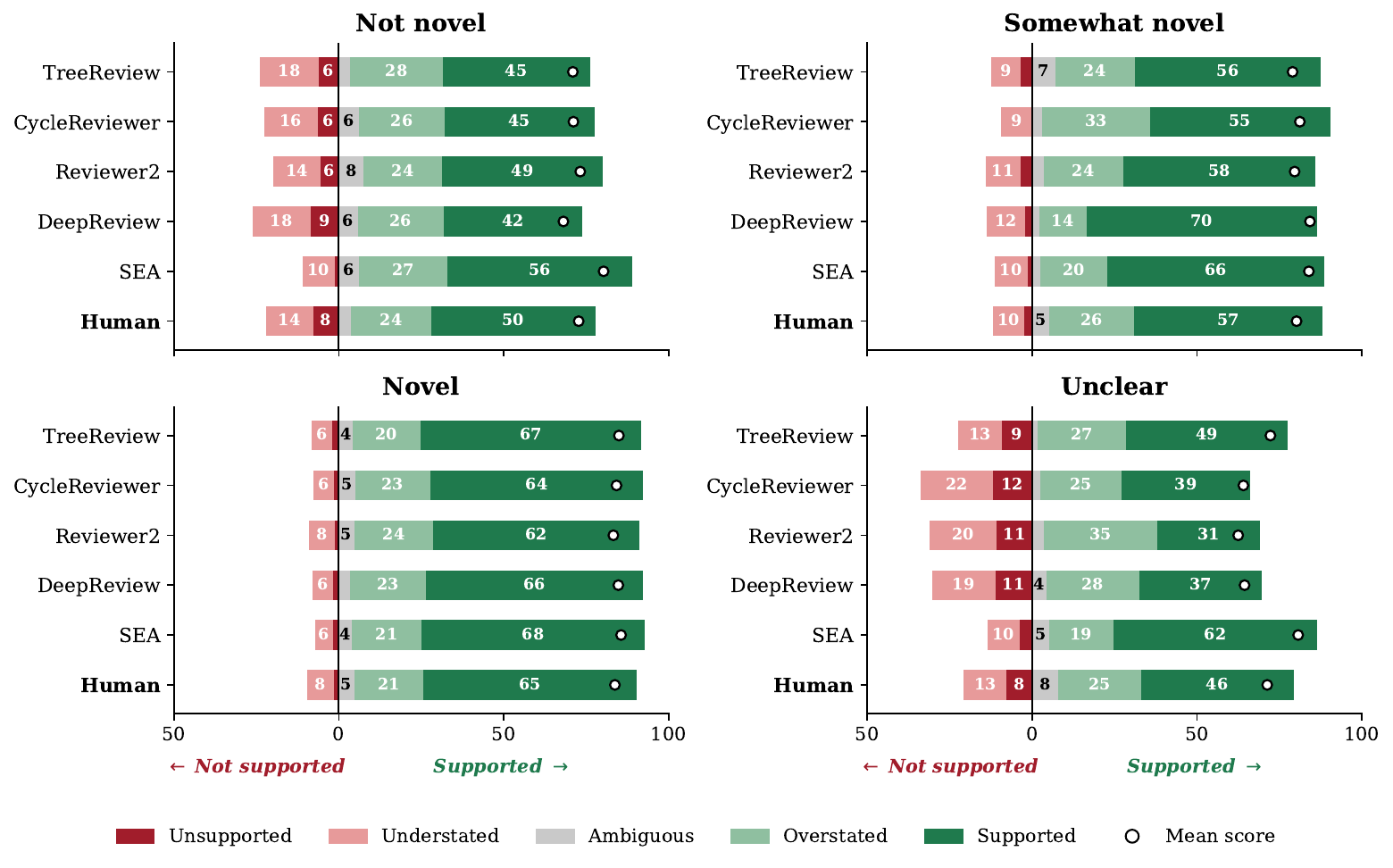}
\vspace{-0.25cm}
\caption{Evidence-support breakdown for novelty claims, stratified by the reviewer's stance (panels: Not novel, Somewhat novel, Novel, Unclear). Each claim--paper judgment takes one of five values: Unsupported ($-2$), Understated ($-1$), Ambiguous ($0$), Overstated ($+1$), Supported ($+2$). Bars show the share (\%) of each value, with unsupported values to the left of zero and supported values to the right; circles show the mean score.}
\label{fig:novelty_b}
\vspace{-0.3cm}
\end{figure}

Figure~\ref{fig:novelty_b} decomposes the scores into the five evidence-support values of \S\ref{subsec:novelty}, from Unsupported ($-2$) to Supported ($+2$). Claims with a \emph{Novel} stance attract stronger literature grounding than \emph{Not novel} or \emph{Unclear} ones, for every reviewer. This reflects the asymmetric evidentiary bar of retrieval-based verification: affirming novelty is corroborated whenever no closely matching prior work surfaces, whereas a ``not novel'' critique is fully grounded only when retrieval locates a specific anticipating paper. Because the pattern holds for humans and LLMs alike, it is a property of the verification task rather than an LLM artifact.

\textit{\textbf{Key Insight:} LLMs' well-grounded novelty claims reflect a preference for easily verifiable claims and a tendency to endorse novelty, not human-level judgment.}

\subsubsection{Flaw Identification \& Major Issues Prioritization}
\label{sec:exp_flaw}
\textbf{Reviewer2} is an exhaustive flaw scanner, with the highest recall of pooled Critical ($\mathbf{0.644}$) and Minor ($\mathbf{0.470}$) flaws, well above human reviews ($0.311$ and $0.253$); it leads under all three judges and both prompt variants (\S\ref{sec:robustness}). Structured LLM pipelines can thus surface valid flaws that individual human reviewers miss. DeepReview and human reviews follow more conservative, targeted patterns. Figure~\ref{fig:comparison_flaws} decomposes extracted flaws into valid and hallucinated counts: Reviewer2 recovers the most valid flaws at a low hallucination rate (${\sim}3.6\%$), whereas CycleReviewer's rate (${\sim}16.2\%$) signals a precision deficit; both rates are lower bounds, because the verifier is conservative in rejecting flaws, especially generic ones (App.~\ref{app:audit}, Table~\ref{tab:disagreement}). Both LLMs and humans concentrate Critical flaws on core methodology and Minor ones on presentation (App.~\ref{sec:detailed_flaw_analysis}). All reviewers, including humans, reach near-identical nCPS ($\approx 0.97$), and nCPS rankings are judge- and prompt-sensitive, so we draw no ranking conclusion from it.

\textit{\textbf{Key Insight:} Some LLMs act as high-sensitivity scanners, catching more critical flaws than individual human reviewers, but they also raise invalid flaws; critical-first ordering is common to all reviewers.}

\subsubsection{Multi-dimensional Constructiveness}
\label{sec:exp_mcs}
\textbf{DeepReview} achieves the highest constructiveness ($\mathbf{0.636}$ vs.\ $0.556$ for human reviews), a lead that holds under expert labels and all three judges (\S\ref{sec:validation}). Per-dimension scores (App.~\ref{sec:detailed_constructiveness}, Fig.~\ref{fig:constructiveness_detailed}) are per-ARC averages, so a low score means a lower \emph{density} of an attribute, not its absence. Under the primary judge, DeepReview leads on \textit{Actionability} (D1\,=\,$1.422$) and \textit{Solution} (D4\,=\,$0.781$), while human reviews identify problems but rarely propose fixes (D4\,=\,$0.427$); both humans ($1.732$) and CycleReviewer ($1.894$) score high on \textit{Specificity} (D2). Reviewer2's high \textit{Justification} (D3\,=\,$0.969$) may partly reflect verbosity, as its low \textit{Solution} rate (D4\,=\,$0.249$) leaves critiques less actionable. On \textit{Tone} (D5), DeepReview scores highest ($1.730$) under the primary judge. Some of these per-dimension differences shrink under expert labels (App.~\ref{app:audit}).

\textit{\textbf{Key Insight:} Constructiveness is a design property, not an intrinsic LLM trait: purpose-built pipelines such as DeepReview propose actionable fixes that other systems and human reviews provide less often.}

\subsubsection{Review Sensitivity to Paper Quality: Accept vs.\ Reject}
\label{sec:accept_reject}
Do the scores track a paper's merit, or are they the same regardless of manuscript quality? We compare each reviewer's per-paper scores for accepted vs.\ rejected submissions with a two-sided Mann-Whitney U test (Holm-corrected across metrics within each reviewer; App.~\ref{app:app_exp2}, Table~\ref{tab:metric_diffs}, Fig.~\ref{fig:core_accept_reject}).
\textbf{Human} review scores differ with the eventual decision, which these reviews informed: reviews of accepted papers receive higher Novelty Scores ($\Delta=+0.051^{*}$) and tighter Prioritization ($\Delta=+0.006^{***}$), while reviews of rejected papers recover significantly more pooled Critical ($\Delta=-0.049^{***}$) and Minor ($\Delta=-0.018^{**}$) flaws, with the Critical gap nearly $3\times$ the Minor one. For the five automated systems the comparison is overwhelmingly non-significant: only DeepReview's Prioritization reaches significance ($\Delta=+0.011^{***}$).

\textit{\textbf{Key Insight:} Although LLM reviewers match or exceed humans on isolated dimensions, only human review scores differ between accepted and rejected papers; LLM reviewers apply a largely uniform standard regardless of merit.}

\subsubsection{Single-prompt LLM baselines}
\label{sec:naive}
On a 60-paper subset with the same judge (App.~\ref{app:naive}), naive single-prompt reviewers are competitive on some dimensions but not across the board. DeepSeek-V4.1-Flash reaches Critical/Minor recall of $0.685$/$0.529$ and constructiveness of $0.645$ (on par with DeepReview, $0.647$), but has the lowest nCPS ($0.949$) and stays below human reviews on depth ($0.410$ vs.\ $0.481$); Qwen3.6-27B has the lowest novelty score ($0.775$). No reviewer, naive or specialized, dominates all dimensions.

\section{Conclusion}
\label{sec:conclusion}
PRISM evaluates human and LLM reviews on equal terms, scoring individual statements against the manuscript and the literature. Its results show that LLM reviewers are specialists rather than replacements.
\textbf{Deployment recommendations.} We recommend a targeted ensemble within a human-assisted pipeline: \textbf{Reviewer2} for flaw scanning (highest recall of verified flaws), \textbf{DeepReview} for drafting constructive feedback (highest actionability and solution density), and \textbf{SEA} for literature checks of novelty claims (highest literature support), with humans retaining the verdict. PRISM makes such recommendations testable: for example, an LLM-proposed flaw list validated by a human should raise Critical recall without raising the hallucination rate.
\textbf{Limitations.} PRISM provides a structured and evidence-grounded tool
for analyzing scientific-reviewing capabilities, but its
scores should be interpreted according to the nature
of each dimension. For evidence-grounded judgments,
such as flaw validity, PRISM shows strong agreement
with expert assessments, and Flaw Recall comparisons
remain consistent across judges and prompt variants.
For more subjective dimensions, such as Depth and
Constructiveness, scores depend on how arguments and
review comments are interpreted, although the main
Constructiveness comparisons remain stable. Novelty
assessment further depends on the coverage and quality
of retrieved literature, while Prioritization exhibits
near-ceiling scores that limit fine-grained comparisons.
PRISM should therefore be used to identify and analyze
specific reviewing capabilities, rather than to establish
a definitive overall ranking of reviewers. Its broader
applicability beyond ML/AI venues remains to be
examined. Further limitations and potential extensions are discussed in App.~\ref{sec:app_limitation}.
\subsection*{AI use statement}
Large language models are the object of study in this work: the five automated reviewer systems we benchmark are LLM-based, and \texttt{Gemini~2.5 Flash Lite} serves as the judge inside PRISM (Section~\ref{sec:method}, Appendix~\ref{sec:app_prism}). For robustness analyses we additionally ran PRISM with \texttt{Claude Haiku 5.5} and \texttt{GPT-5.6-Sol} as judges; Qwen3.6-27B and DeepSeek-V4.1-Flash served as single-prompt reviewers (Section~\ref{sec:validation}, Appendix~\ref{app:additional}). LLM-produced judgments were validated against expert annotators (Section~\ref{sec:human_validation}).
Generative AI tools were also used to assist with language editing, code debugging, and formatting of selected figures and tables. All AI-assisted outputs were manually reviewed and verified by the authors before inclusion. Generative AI was not used to autonomously determine the scientific conclusions of this work.
We have reviewed all AI-assisted work and take responsibility for the final content of this paper, including text, claims, and artifacts produced with the aid of generative AI.

\ificlrfinal
\subsubsection*{Acknowledgments}
This research is funded by CAIR, College of Engineering \& Computer Science, VinUniversity, Hanoi, Vietnam.
The work of Duy A. Nguyen was supported in part by a PhD fellowship from the VinUni-Illinois Smart Health Center, VinUniversity, Hanoi, Vietnam.
\fi

\bibliography{iclr2027_conference}
\bibliographystyle{iclr2027_conference}

\input{sections/appendix}

\end{document}

%% file: math_commands.tex
\usepackage{amsmath,amsfonts,bm}

\def\eqref#1{equation~\ref{#1}}

\def\1{\bm{1}}

\DeclareMathAlphabet{\mathsfit}{\encodingdefault}{\sfdefault}{m}{sl}
\SetMathAlphabet{\mathsfit}{bold}{\encodingdefault}{\sfdefault}{bx}{n}



%% file: sections/appendix.tex
\appendix

\section*{Appendix Table of Contents}
\vspace{0.4em}
{\small\setlength{\parskip}{2pt}\setlength{\parindent}{0pt}
\noindent\hyperref[sec:app_formal]{\textbf{A\quad Formal Problem Definition}}\par\vspace{3pt}
\noindent\hyperref[sec:app_baseline]{\textbf{B\quad Experimental Details}}\par\vspace{1pt}
\hspace*{1.8em}B.1\hspace{0.4em}Dataset Selection\par\vspace{1pt}
\hspace*{1.8em}B.2\hspace{0.4em}Reviewer Baselines and Implementations\par\vspace{1pt}
\hspace*{1.8em}B.3\hspace{0.4em}Review Generation Process\par\vspace{3pt}
\noindent\hyperref[sec:app_prism]{\textbf{C\quad PRISM Evaluation Framework: Pipeline Details}}\par\vspace{1pt}
\hspace*{1.8em}C.1\hspace{0.4em}The PRISM Evaluation Pipeline\par\vspace{1pt}
\hspace*{1.8em}C.2\hspace{0.4em}PRISM Judge Setup\par\vspace{1pt}
\hspace*{1.8em}C.3\hspace{0.4em}Prompt Templates by Dimension\par\vspace{1pt}
\hspace*{1.8em}C.4\hspace{0.4em}Human Validation Protocol\par\vspace{3pt}
\noindent\hyperref[sec:app_exp1]{\textbf{D\quad Metric Independence Analysis via Pearson Correlation}}\par\vspace{3pt}
\noindent\hyperref[app:app_exp2]{\textbf{E\quad Full Cross-Dataset Quantitative Results}}\par\vspace{1pt}
\hspace*{1.8em}E.1\hspace{0.4em}Statistical Significance Testing Protocol\par\vspace{1pt}
\hspace*{1.8em}{E.2\hspace{0.4em}Depth of Analysis}\par\vspace{1pt}
\hspace*{1.8em}{E.3\hspace{0.4em}Novelty Assessment}\par\vspace{1pt}
\hspace*{1.8em}{E.4\hspace{0.4em}Flaw Identification \& Prioritization}\par\vspace{1pt}
\hspace*{1.8em}{E.5\hspace{0.4em}Multi-Dimensional Constructiveness}\par\vspace{1pt}
\hspace*{1.8em}E.6\hspace{0.4em}Review Sensitivity to Paper Quality\par\vspace{1pt}
\hspace*{1.8em}E.7\hspace{0.4em}Robustness Across Judges\par\vspace{3pt}
\noindent\hyperref[app:app_exp3]{\textbf{F\quad Qualitative Analysis \& Case Studies}}\par\vspace{1pt}
\hspace*{1.8em}F.1\hspace{0.4em}Depth of Analysis\par\vspace{1pt}
\hspace*{1.8em}F.2\hspace{0.4em}Novelty Assessment\par\vspace{1pt}
\hspace*{1.8em}F.3\hspace{0.4em}Flaw Identification \& Major Issues Prioritization\par\vspace{1pt}
\hspace*{1.8em}F.4\hspace{0.4em}Multi-dimensional Constructiveness\par\vspace{3pt}
\noindent\hyperref[sec:app_limitation]{\textbf{G\quad Limitations}}\par\vspace{3pt}
\noindent\hyperref[app:additional]{\textbf{H\quad Additional Validation and Analyses}}\par\vspace{1pt}
\hspace*{1.8em}H.1\hspace{0.4em}Extended Comparison with Related Work\par\vspace{1pt}
\hspace*{1.8em}H.2\hspace{0.4em}Prompt Sensitivity\par\vspace{1pt}
\hspace*{1.8em}H.3\hspace{0.4em}Stage-wise Audit and Disagreement Analysis\par\vspace{1pt}
\hspace*{1.8em}H.4\hspace{0.4em}Single-prompt LLM Baselines\par
}
\newpage


\section{Formal problem definition}
\label{sec:app_formal}

The fundamental challenge in benchmarking automated peer reviewers lies in the highly subjective, domain-specific, and unstructured nature of scientific critiques. While existing literature often treats LLMs as either pure text generators or generic evaluators, assessing a scientific peer review requires measuring cognitive depth rather than mere linguistic fluency.

To systematically evaluate this, we formalize the peer review benchmarking process. Let $P$ denote a submitted scientific manuscript. In our setting, an LLM-based reviewer baseline $\mathcal{M}$ processes $P$ to generate an automated review, denoted as $R_{LLM} = \mathcal{M}(P)$. Simultaneously, we possess the human reviews $R_{human}$ submitted for the same manuscript $P$. PRISM does not treat $R_{human}$ as ground truth: human and LLM reviews are scored by the same function.

The central problem addressed in this work is to construct a multi-dimensional evaluation function, denoted as $\mathcal{E}$. Rather than relying on superficial n-gram matching metrics (like ROUGE) or unconstrained prompting, our framework requires $\mathcal{E}$ to score any review $R$ of $P$, human or LLM, against evidence outside the review: the manuscript, independently retrieved literature, and the pooled verified-flaw reference $\mathcal{F}(P)$ built from \emph{all} reviews of $P$:
$$ \mathcal{S}(R) = \mathcal{E}(R, P, \mathcal{F}(P)), \qquad R \in \{R_{LLM}, R_{human}\}, $$
where $\mathcal{S}$ represents a set of quantitative scores spanning diverse cognitive aspects. Comparing $\mathcal{S}(R_{LLM})$ with $\mathcal{S}(R_{human})$ then measures the capability gap between automated and human reviews on equal terms, penalizing superficial summarization, hallucinated flaws, ungrounded novelty claims and un-actionable feedback.

\section{Experimental details}
\label{sec:app_baseline}
\subsection{Dataset selection.}
\label{sec:app_dataset}

\begin{minipage}[t]{0.52\textwidth}\vspace{0pt}
\centering
\captionof{table}{Distribution of selected papers across conferences and decision categories.}
\label{tab:data_selection_stats}
\resizebox{\linewidth}{!}{%
\begin{tblr}{
  column{1} = {l},
  column{2} = {c},
  column{3} = {c},
  column{4} = {c},
  column{5} = {c},
  column{6} = {c},
  hline{1,6} = {-}{0.08em},
  hline{2} = {-}{0.05em},
  hline{5} = {-}{dashed},
}
\textbf{Venue (Year)} & \textbf{Oral} & \textbf{Spotlight} & \textbf{Poster} & \textbf{Reject} & \textbf{Total} \\
ICLR 2026             & 26            & -                  & 80              & 94              & 200            \\
ICML 2025             & 50            & 39                 & 59              & 52              & 200            \\
NeurIPS 2025          & 50            & 50                 & 50              & 50              & 200            \\
\textbf{Total}        & \textbf{126}  & \textbf{89}        & \textbf{189}    & \textbf{196}    & \textbf{600}   \\
\end{tblr}
}   
\end{minipage}
\hfill
\begin{minipage}[t]{0.45\textwidth}\vspace{0pt}
    \includegraphics[width=\linewidth]{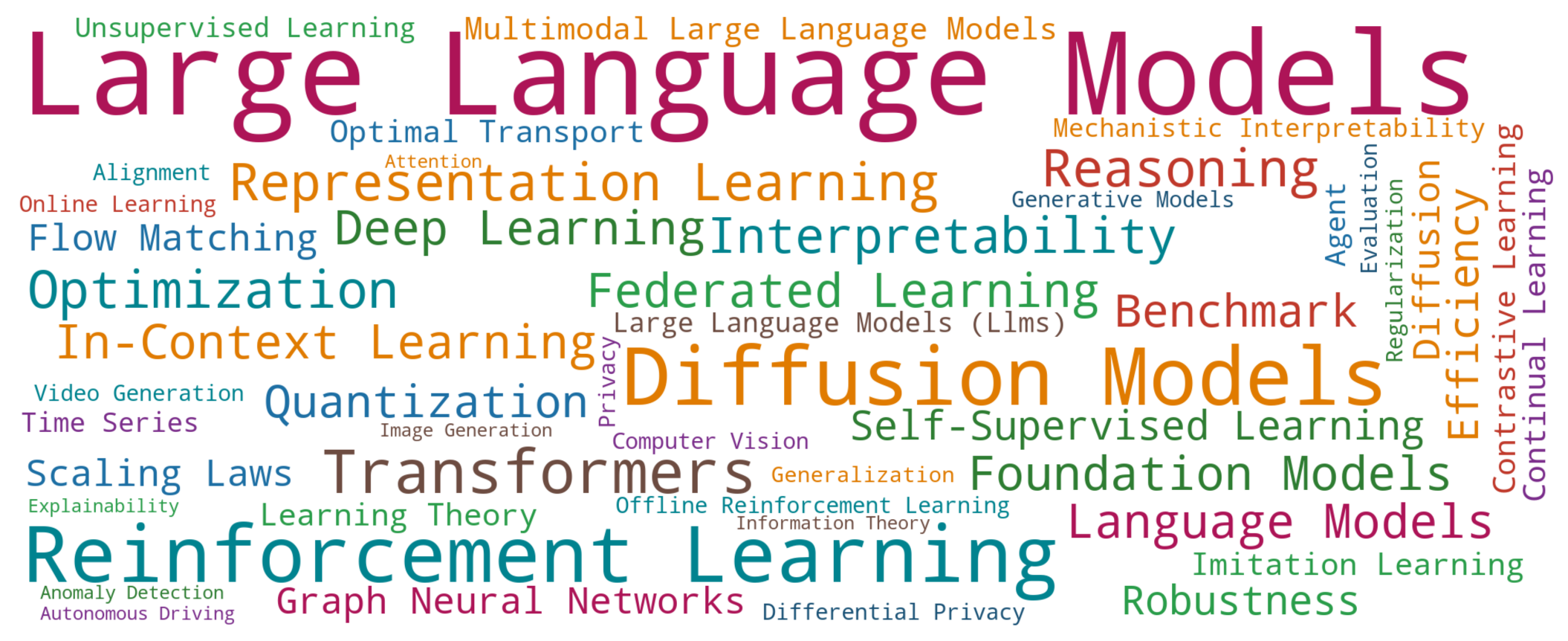}
    \par\vspace{-6px}
    \captionof{figure}{Top 50 popular keywords within our benchmark.}
    \label{fig:keyword_cloud}
\end{minipage}

We evaluate PRISM on a stratified benchmark drawn from three conference splits: \textbf{ICLR 2026}, \textbf{ICML 2025}, and \textbf{NeurIPS 2025} (Table~\ref{tab:data_selection_stats}). These venue-years were selected as the most recently concluded at the time of benchmark construction, so that both the manuscripts and their peer reviews postdate the training cutoff of every evaluated baseline (see Appendix~\ref{sec:app_limitation} for further discussion). For each of the three venue-years, we construct a representative subset of exactly 200 manuscripts, stratified across their final decision categories (\textit{Reject}, \textit{Poster}, \textit{Spotlight}, \textit{Oral}) and cover various topics (Figure \ref{fig:keyword_cloud}). During the sampling process, we strictly preserve the original score distribution of a full conference pool. As a result, the number of papers within each decision tier organically reflects the natural acceptance dynamics and quality distribution of each specific venue. This approach ensures comprehensive outcome coverage and high-fidelity review-quality diversity, while keeping the benchmark tractable for end-to-end multi-system evaluation.

\subsection{Reviewer baselines and implementations.} 
\subsubsection{Taxonomy of Baseline LLM Reviewers}
We compare human reviews against five automated reviewer systems: \textbf{SEA-E}, \textbf{DeepReview}, \textbf{Reviewer2}, \textbf{CycleReviewer}, and \textbf{TreeReview}. These systems span two broad paradigms.

\paragraph{Supervised fine-tuning methods.}
SEA-E~\cite{yu-etal-2024-automated-SEA} is a structured evaluation model trained to output review components such as summaries, strengths, weaknesses, and questions. CycleReviewer~\cite{cyclereviewer} is optimized through an iterative preference-based training framework in which a reviewer model is progressively refined from win/lose comparisons. DeepReview~\cite{zhu-etal-2025-deepreview} uses a multi-stage reasoning pipeline that explicitly models multi-perspective analysis, and reliability checking before producing a final review.

\paragraph{Prompting-based methods.}
Reviewer2~\cite{gao2024reviewer2optimizingreviewgeneration} is based on a two-stage rubric-driven process that first generates aspect-specific questions and then answers them to synthesize the final review. TreeReview~\cite{chang-etal-2025-treereview} follows a hierarchical reasoning strategy, decomposing the review into a tree of sub-questions and aggregating leaf-level evidence into a complete critique. In our experiments, prompting-based baselines are executed under a standardized backbone configuration to isolate the effect of the prompting framework rather than confounding it with backbone choice.\todo{AUTHOR: confirm backbone names; B.2.2 lists Qwen-3.5-14B for Reviewer2 and Qwen3-14B for TreeReview}

\subsubsection{Baseline Implementation and Configuration}
\paragraph{SEA-E}
SEA-E operates as a structured evaluation model, utilizing the model \textit{ECNU-SEA/SEA-E} to generate comprehensive review components such as summaries, strengths, weaknesses, and numerical ratings. To accommodate full-length manuscripts, the engine is configured with a 70,000-token context window. The pipeline processes a batch size of 4 papers simultaneously, generating the final critique with the maximum output length capped at 8,000 tokens. To ensure a balance between analytical diversity and factual coherence, the generation hyperparameters are strictly configured with a temperature of 0.7 and a top-p of 0.9 to maintain high-quality and academically rigorous outputs.

\paragraph{CycleReviewer}
CycleReviewer utilizes a model \textit{WestlakeNLP/CycleReviewer-ML-Llama-3.1-8B} optimized through an iterative preference-based reasoning framework. All inference workloads are executed on NVIDIA RTX A5000 GPUs. The model employs a 24,000-token context window to accommodate and process complete manuscript texts. For each manuscript, the system executes 2 to 3 iterative refinement passes to progressively enhance the review quality. The 8B engine operates on a single GPU configuration. It generates critiques with a maximum generation length of 3,000 tokens. Generation hyperparameters are configured with a temperature of 0.7, top-p of 0.9, top-k of 50, and a repetition penalty of 1.2.

\paragraph{DeepReview}
DeepReview utilizes the \textit{WestlakeNLP/DeepReviewer-14B} core reasoning engine alongside a retrieval-augmented subsystem powered by OpenScholar. All inference workloads are executed on NVIDIA RTX A5000 GPUs. Aligning with the original architecture, the retrieval module employs \textit{Llama-3.1-OpenScholar-8B} (configured with a 70,000-token context limit) for evidence synthesis and \textit{Qwen-2.5-3B-Instruct} (configured with a 10,000-token context limit) for query processing. For each manuscript, the system transforms generated questions into search keywords to retrieve approximately 30 candidate papers, utilizing a dedicated reranking model to select the top 10 most relevant sources for grounding. Retrieval is restricted to scholarly literature available before the manuscript's submission date and excludes existing conference reviews, author--reviewer discussions and acceptance decisions, so the module cannot access the evaluation outcomes of the papers under review. DeepReview is the only evaluated system with a retrieval component; SEA-E, CycleReviewer, Reviewer2 and TreeReview rely only on the paper and on parametric knowledge. The core 14B engine operates across two GPUs via tensor parallelism. It processes a batch size of 8 papers with a maximum generation length of 7,000 tokens. Generation hyperparameters are configured with a temperature of 0.8, top-p of 0.9, top-k of 50, and a repetition penalty of 1.2.

\paragraph{Reviewer2}
The Reviewer2 framework originally operates on a two-stage prompting methodology utilizing custom checkpoints (\textit{GitBag/Reviewer2\_Mp} and \textit{GitBag/Reviewer2\_Mr}). However, due to the suboptimal generation quality observed from these native models during our preliminary evaluations, we replace the underlying generation engine with the open-weights \textit{Qwen-3.5-14B} model. Crucially, we strictly retain Reviewer2's official prompt templates to preserve the methodological integrity of their two-stage pipeline. All inference workloads are executed on NVIDIA RTX A5000 GPUs. To process extensive manuscripts, the engine is configured with an 80,000-token context window. The pipeline processes a batch size of 4 papers simultaneously. During Phase 1, the model generates specific review questions, and in Phase 2, it synthesizes the final critique based on these generated prompts, with the maximum generation length capped at 7,000 tokens. Across both stages, the generation hyperparameters are strictly configured with a temperature of 0.8, top-p of 0.9, top-k of 50, and a repetition penalty of 1.2 to ensure diverse yet academically rigorous outputs.

\paragraph{TreeReview}
TreeReview models the peer review process as a hierarchical and bidirectional question-answering framework. While the original implementation utilizes GPT-4o, we standardized the backbone to \textit{Qwen3-14B} to ensure a fair comparison across all prompting-based baselines. All inference workloads are executed on NVIDIA RTX A5000 GPUs. To accommodate the full paper text alongside the dynamically expanding tree of sub-questions, the engine is configured with an 80,000-token context window. The pipeline processes a batch size of 4 papers simultaneously. Following its core logic, the system recursively decomposes high-level review objectives into fine-grained sub-questions and aggregates answers from leaf to root to synthesize the final critique, with the maximum output length capped at 7,000 tokens. Across all reasoning stages, the generation hyperparameters are strictly set to a temperature of 0.8, top-p of 0.9, top-k of 50, and a repetition penalty of 1.2.

\subsubsection{Review generation process.}
Before applying our evaluation framework, we must first generate the corresponding AI reviews. For each of the 600 papers in our dataset, we provide the complete textual content—including all sections and tables represented in text form—to all five LLM reviewer baselines. Figures and other visual elements are excluded, as the LLM reviewers considered in this study do not yet reliably support multimodal (vision–language) understanding. Each model then independently generates a complete peer review based on its respective methodology. This process yields a comprehensive corpus of 3,000 automated reviews, which serves as the primary testbed for all subsequent PRISM evaluations.

\section{PRISM Evaluation Framework: Pipeline Details and Experimental Setup}
\label{sec:app_prism}
To ensure full reproducibility and provide transparency into our methodology, this section details the hyperparameters used for all evaluated baselines and the exact prompt templates deployed for multi-dimensional assessment.

\subsection{The PRISM Evaluation Pipeline}
\begin{figure}[H]
    \centering
    \includegraphics[width=\linewidth]{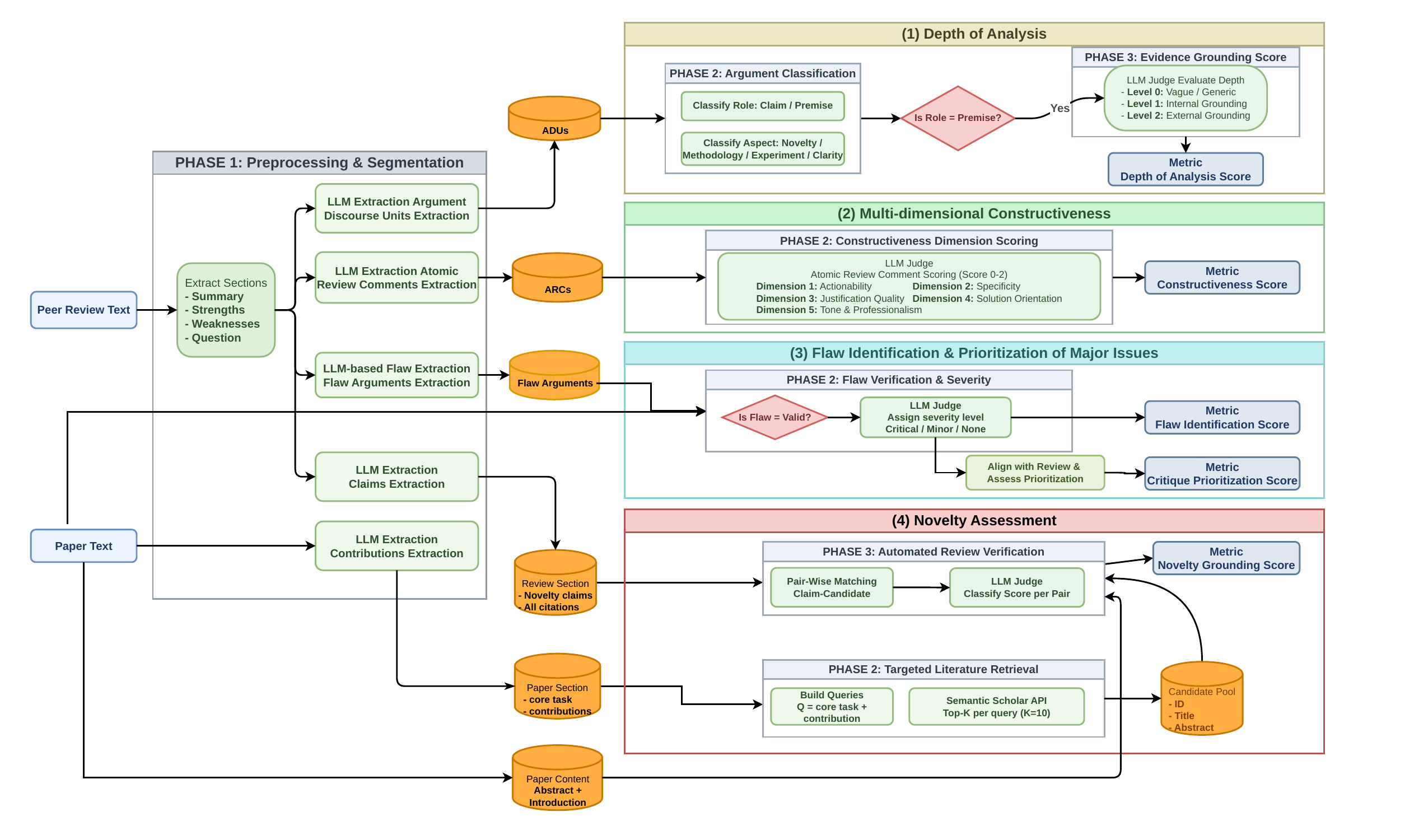}
    \caption{Detailed Flow of PRISM}
    \label{fig:detailed_diagram}
\end{figure}

\subsection{PRISM Judge Setup.}
To compute the diverse evaluation metrics defined in our framework PRISM, we adopt the LLM-as-a-Judge paradigm. We deploy \textbf{Gemini 2.5 Flash Lite} (\texttt{gemini-2.5-flash-lite})~\citep{gemini2025team} as the core evaluation engine for all metric extraction and scoring tasks. To ensure strict reproducibility and minimize generation variance, we explicitly configure the model parameters by setting the temperature to 0.0 and top-$p$ to 0.95, without utilizing top-$k$ sampling. During the evaluation phase, the model is strictly prompted with our standardized rubrics to systematically extract arguments, verify flaw validity against the manuscript, and compute component scores for both human and automated reviewers equally.

\subsection{Prompt Templates by Dimension}
\paragraph{Dimension 1: Depth of Analysis.}

The evaluation of Depth of Analysis is executed through a sequential, three-phase prompting pipeline designed to isolate and score scientific arguments. The first phase segments the raw review text into discrete Argumentative Discourse Units (ADUs), as detailed in Figure \ref{fig:doa_prompt_phase1}. Subsequently, the second phase classifies each extracted ADU into its corresponding argument role (Claim or Premise) and maps it to one of four predefined aspect topics, illustrated in Figure \ref{fig:doa_prompt_phase2}. Finally, the third phase evaluates the empirical depth of the review by computing a categorical Grounding Score specifically for the identified Premises, which is presented in Figure \ref{fig:doa_prompt_phase3}.

\begin{figure}[H]
\centering
\begin{tcolorbox}[colback=gray!5!white, colframe=black!75, title=\textbf{Phase 1: Argumentative Discourse Unit (ADU) Segmentation}]
\textbf{ROLE AND OBJECTIVE} \\
You are an expert NLP researcher specializing in Scholarly Argumentation Mining. Your task is to segment the following scientific peer review text into distinct Argumentative Discourse Units (ADUs).

\vspace{5pt}
\textbf{GUIDELINES}
\begin{enumerate}
    \item After each independent logical unit, insert the exact marker \texttt{<sep>}.
    \item Keep the original text in the exact same order WITHOUT adding, removing, or altering any words.
    \item CRITICAL: DO split complex/compound sentences when a conclusion/claim is joined with its supporting reason/evidence. Split at:
    \begin{itemize}
        \item Logical and causal conjunctions ("because", "as", "since", "due to", "but", "however", etc).
        \item Relative pronouns (e.g., "which", "that", "who", etc).
        \item Participial phrases indicating result/proof ("demonstrating", "showing", "proving", "making", "resulting in", etc).
    \end{itemize}
    \item IGNORE structural headings entirely (e.g., "**Summary:**", "**Strengths:**"). DO NOT append \texttt{<sep>} to them.
\end{enumerate}

\vspace{5pt}
\textbf{INPUT REVIEW TEXT:} \\
\{raw\_review\_text\}
\end{tcolorbox}
\caption{PRISM Depth of Analysis Prompt (1/3): Argumentative Discourse Unit (ADU) Segmentation.}
\label{fig:doa_prompt_phase1}
\end{figure}

\begin{figure}[H]
\centering
\begin{tcolorbox}[colback=gray!5!white, colframe=black!75, title=\textbf{Phase 2: Argument Role and Aspect Topic Classification}]
\textbf{ROLE AND OBJECTIVE} \\
Classify the Argument Role and Aspect Topic for the provided list of segmented ADUs. Use the full review as macro-context. For EACH argument in the list, classify its Argument Role and Aspect Topic.

\vspace{5pt}
\textbf{1. ARGUMENT ROLE CLASSIFICATION (Choose exactly ONE)}
\begin{itemize}
    \item \textbf{Claim (Conclusion / Point):} The statement that is being argued \textit{for}. It is the controversial statement or the central point that needs support.
    \item \textbf{Premise (Reason / Support):} The statement that is used to \textit{support} the Claim. It provides the reasons, evidence, justifications, or grounds to accept the Claim.
\end{itemize}

\textbf{Example:}
\begin{itemize}
    \item \textit{Macro-Context:} "The proposed method is not novel. Similar architectures were already introduced by Smith et al. (2023)."
    \item \textit{ADU:} "The proposed method is not novel." $\rightarrow$ \textbf{Claim} (Needs evidence to be proven).
    \item \textit{ADU:} "Similar architectures were already introduced by Smith et al. (2023)." $\rightarrow$ \textbf{Premise} (Direct evidence to support the Claim).
\end{itemize}

\vspace{5pt}
\textbf{2. ASPECT TOPIC CLASSIFICATION (Choose exactly ONE)}
\begin{itemize}
    \item \textbf{Novelty \& Related Work:} Discusses originality, plagiarism, or literature review.
    \item \textbf{Methodology \& Theoretical Soundness:} Discusses math, algorithms, architecture, or dataset guidelines.
    \item \textbf{Experimental Design \& Evaluation:} Discusses empirical setups, baselines, ablation studies, and metrics.
    \item \textbf{Clarity, Presentation \& Reproducibility:} Discusses writing quality, typos, or formatting.
\end{itemize}

\vspace{5pt}
\textbf{OUTPUT FORMAT} \\
Respond ONLY with a valid JSON object. \\
\textbf{MACRO-CONTEXT:} \{macro\_context\} \\
\textbf{LIST OF ARGUMENTS:} \{argument\_list\}
\end{tcolorbox}
\caption{PRISM Depth of Analysis Prompt (2/3): Argument Role and Aspect Topic Classification.}
\label{fig:doa_prompt_phase2}
\end{figure}

\begin{figure}[H]
\centering
\begin{tcolorbox}[colback=gray!5!white, colframe=black!75, title=\textbf{Phase 3: Premise Grounding Score Evaluation}]
\textbf{ROLE AND OBJECTIVE} \\
You are an expert NLP researcher. I will provide a full peer review (for context) and a list of PREMISES (evidence). Evaluate the "Grounding Score" (Depth of Evidence) for EACH premise.

\vspace{5pt}
\textbf{GROUNDING SCORE DEFINITIONS}
\begin{itemize}
    \item \textbf{Score 0 (Generic / Vague):} The premise is vague and lacks specific anchors (e.g., "The datasets", "Past research", "The equations").
    \item \textbf{Score 1 (Internal Grounding):} The premise explicitly anchors to elements INSIDE the manuscript (e.g., "Equation 4", "Table 2", "The proposed module").
    \item \textbf{Score 2 (External / Comparative):} The premise anchors to EXTERNAL knowledge outside the manuscript (e.g., "(Smith et al., 2023)", "RoBERTa", "The GLUE benchmark").
\end{itemize}

\vspace{5pt}
\textbf{OUTPUT FORMAT} \\
Respond ONLY with a valid JSON object. \\
\textbf{MACRO-CONTEXT:} \{macro\_context\} \\
\textbf{LIST OF PREMISES:} \{premise\_list\}
\end{tcolorbox}
\caption{PRISM Depth of Analysis Prompt (3/3): Premise Grounding Score Evaluation.}
\label{fig:doa_prompt_phase3}
\end{figure}

\paragraph{Running Example: Depth of Analysis.}

To illustrate the full Depth of Analysis pipeline, consider the following excerpt from a raw review: \textit{``3 seeds is too few to get any statistical confidence, especially without doing independent hyperparameter sweeps for each baseline. While in the past this has been standard, as a field we continually have shown that the statistical power of our experiments are laughably poor. The performance of the proposed goal-conditioned RL algorithm on the most challenging tasks was less than 50\%. QRL assumes deterministic dynamics of the environment, while TD InfoNCE learns without such assumption.''}

Processing this text through our three-phase framework yields the following structured output:
\begin{itemize}
    \item \textbf{Claim:} \textit{``3 seeds is too few to get any statistical confidence...''} \\
    $\rightarrow$ \textbf{Aspect:} Experimental Design \& Evaluation.
    
    \item \textbf{Premise 1:} \textit{``While in the past this has been standard, as a field we continually have shown that the statistical power of our experiments are laughably poor.''} \\
    $\rightarrow$ \textbf{Aspect:} Experimental Design \& Evaluation. \\
    $\rightarrow$ \textbf{Grounding Score: 0} (Generic/Vague --- purely a community-level assertion with no specific anchor).
    
    \item \textbf{Premise 2:} \textit{``The performance of the proposed goal-conditioned RL algorithm on the most challenging tasks was less than 50\%.''} \\
    $\rightarrow$ \textbf{Aspect:} Experimental Design \& Evaluation. \\
    $\rightarrow$ \textbf{Grounding Score: 1} (Internal --- directly references a quantitative result reported inside the manuscript).
    
    \item \textbf{Premise 3:} \textit{``QRL assumes deterministic dynamics of the environment, while TD InfoNCE learns without such assumption.''} \\
    $\rightarrow$ \textbf{Aspect:} Methodology \& Theoretical Soundness. \\
    $\rightarrow$ \textbf{Grounding Score: 2} (External/Comparative --- explicitly references QRL, a published external baseline with a known stated assumption).
\end{itemize}

Finally, the overall DoA score is defined as the harmonic mean of the Premise Ratio and the Normalized Grounding Score. This formulation ensures that a high DoA score requires both a sufficient volume of supporting arguments and a high degree of external/internal grounding:
$$DoA = 2 \times \frac{R_{premise} \times S_{grounding}}{R_{premise} + S_{grounding}}$$

\textit{Calculation for the Running Example:} 
In the excerpt above, the pipeline extracted a total of $4$ ADUs ($1$ Claim and $3$ Premises). The premises received grounding scores of $0$, $1$, and $2$. Applying our aggregation formulas yields:
$$R_{premise} = \frac{3}{4} = 0.75$$
$$S_{grounding} = \frac{0 + 1 + 2}{3 \times 2} = 0.5$$
$$DoA = 2 \times \frac{0.75 \times 0.5}{0.75 + 0.5} = 0.6$$

\paragraph{Dimension 2: Novelty Assessment.}

The evaluation of Novelty Assessment is executed through a three-phase pipeline that grounds reviewer novelty claims against an external body of retrieved prior work, enabling verifiable, evidence-based scoring rather than purely introspective LLM judgment. The pipeline is summarized in Figure~\ref{fig:detailed_diagram}.

\textbf{Phase 1: Structured Target Extraction.} The first phase, illustrated in Figure~\ref{fig:novelty_prompt_phase1}, processes both the submitted paper and the peer review in a single LLM call. From the paper, the model extracts a \texttt{core\_task} (the concrete problem addressed, $\leq$20 words), up to three structured \texttt{contributions} (each with a verbatim author claim, a normalized paraphrase, and a source location hint), a set of \texttt{key\_terms}, and a list of \texttt{must\_have\_entities} (named models, datasets, metrics). From the review, the model identifies \texttt{novelty\_claims} — verbatim review sentences asserting that the paper is or is not novel — and annotates each with its argumentative stance (\texttt{not\_novel} / \texttt{somewhat\_novel} / \texttt{novel} / \texttt{unclear}), confidence language, and any cited prior-work strings. A deterministic regex fallback independently augments the citation list from the raw review text to prevent hallucinated references.

\textbf{Phase 2: Related Work Retrieval (non-LLM).} The second phase is entirely deterministic and requires no LLM call. The \texttt{core\_task} and contribution names extracted in Phase 1 are composed into structured search queries, which are issued to the Semantic Scholar Academic Graph API. Raw candidates are then deduplicated via approximate title-abstract similarity ($\geq$0.96 threshold), filtered for non-technical documents (editorials, errata, etc.), and temporally constrained to papers published no later than the submission year. Finally, a Maximal Marginal Relevance (MMR) algorithm~\cite{carbonell1998use} selects a diverse top-$K$ candidate pool ($K=30$ by default) that balances retrieval relevance against redundancy.

\textbf{Phase 3: LLM Judge Verification.} The third phase, depicted in Figure~\ref{fig:novelty_prompt_phase3}, instantiates an LLM Judge for each (novelty claim, related-work candidate) pair. Given the review sentence, the abstract and introduction of the paper under review, and the title and abstract of the related work, the judge assigns a five-level verdict on a $[-2, +2]$ integer scale.

\noindent A verdict of \textbf{SUPPORTED} ($+2$) indicates that the retrieved evidence confirms the reviewer's novelty assessment. \textbf{OVERSTATED} ($+1$) flags cases where partial similarity exists but the reviewer's claim of ``same as / not novel'' is too strong. \textbf{AMBIGUOUS} ($0$) denotes unverifiable claims. \textbf{UNDERSTATED} ($-1$) penalizes reviewers who miss closely related prior work that is present in the candidate pool. \textbf{UNSUPPORTED} ($-2$) indicates that retrieved evidence contradicts the claim or that no supporting evidence was found.

Per-sentence scores are aggregated across the retrieved candidate pool using a configurable policy. In the experiments reported in this paper, the default is a relevance-weighted top-3 rule, not \texttt{max}: each claim is scored against its three highest-relevance candidates so that one spuriously favorable match does not dominate the result. We retain \texttt{max} only as an ablation alternative, where it yields a more optimistic estimate of support. The final \textbf{Novelty Verification Score (NS)} for a reviewer is the mean aggregated score over all their novelty claims.

\begin{figure}[H]
\centering
\begin{tcolorbox}[colback=gray!5!white, colframe=black!75, title=\textbf{Phase 1: Structured Target Extraction from Paper and Review}]
\textbf{ROLE AND OBJECTIVE} \\
TASK: Extract structured targets for verifiable novelty checking. \\
You will receive TWO sources below: PAPER TEXT and REVIEW TEXT. \\
Return STRICT JSON only (no markdown, no code fences, no extra keys). \\
The output MUST contain BOTH top-level keys: \texttt{"paper"} and \texttt{"review"}. \\
For novelty claims, the \texttt{"text"} field MUST be verbatim from the REVIEW TEXT (1--2 sentences max). \\
If the review contains no novelty claims, return an empty \texttt{novelty\_claims} list but still include \texttt{review}.

\vspace{5pt}
\textbf{OUTPUT JSON SCHEMA (must match exactly):} \\
{\small\ttfamily
\noindent\{\\
\hspace*{1em}"paper": \{\\
\hspace*{2em}"core\_task": "string (<=20 words)",\\
\hspace*{2em}"contributions": [\\
\hspace*{3em}\{\\
\hspace*{4em}"name": "short name for contribution (<=15 words)",\\
\hspace*{4em}"author\_claim\_text": "verbatim quote from paper (<=40 words)",\\
\hspace*{4em}"description": "normalized paraphrase (<=60 words)",\\
\hspace*{4em}"source\_hint": "location tag e.g. Abstract, Introduction §1, Conclusion"\\
\hspace*{3em}\}\\
\hspace*{2em}],\\
\hspace*{2em}"key\_terms": ["5-12 short phrases"],\\
\hspace*{2em}"must\_have\_entities": ["model/dataset/metric names if any"]\\
\hspace*{1em}\},\\
\hspace*{1em}"review": \{\\
\hspace*{2em}"novelty\_claims": [\\
\hspace*{3em}\{\\
\hspace*{4em}"claim\_id": "C1",\\
\hspace*{4em}"text": "verbatim review claim (1-2 sentences max)",\\
\hspace*{4em}"stance": "not\_novel | somewhat\_novel | novel | unclear",\\
\hspace*{4em}"confidence\_lang": "high | medium | low",\\
\hspace*{4em}"mentions\_prior\_work": true,\\
\hspace*{4em}"prior\_work\_strings": ["author-year strings or titles as written"],\\
\hspace*{4em}"evidence\_expected": "method\_similarity | task\_similarity | results\_similarity | theory\_overlap | dataset\_overlap"\\
\hspace*{3em}\}\\
\hspace*{2em}],\\
\hspace*{2em}"all\_citations\_raw": ["everything that looks like a citation/title/arxiv id/url"]\\
\hspace*{1em}\}\\
\}\par}

\vspace{5pt}
\textbf{INPUT:} \{paper\_text\} \quad \{review\_text\}
\end{tcolorbox}
\caption{PRISM Novelty Assessment Prompt (1/2): Extracting structured paper contributions and review novelty claims as verifiable targets.}
\label{fig:novelty_prompt_phase1}
\end{figure}

\begin{figure}[H]
\centering
\begin{tcolorbox}[colback=gray!5!white, colframe=black!75, title=\textbf{Auxiliary Prompt: Core Task Extraction}]
\textbf{ROLE AND OBJECTIVE} \\
TASK: Extract the core task from a research paper. \\
You will receive the full text of a paper below. \\
Return STRICT JSON only (no markdown, no code fences, no extra keys).

\vspace{5pt}
\textbf{OUTPUT JSON SCHEMA (must match exactly):} \\
{\small\ttfamily
\noindent\{\\
\hspace*{1em}"core\_task": "string (<=20 words)"\\
\}\par}

\vspace{5pt}
\textbf{INPUT:} \{paper\_text\}
\end{tcolorbox}
\caption{Auxiliary novelty prompt for extracting the paper's core task.}
\label{fig:novelty_core_task_prompt}
\end{figure}

\begin{figure}[H]
\centering
\begin{tcolorbox}[colback=gray!5!white, colframe=black!75, title=\textbf{Auxiliary Prompt: Contribution Extraction}]
\textbf{ROLE AND OBJECTIVE} \\
TASK: Extract the main contributions claimed by the authors. \\
You will receive the full text of a paper below. \\
Return STRICT JSON only (no markdown, no code fences, no extra keys).

\vspace{5pt}
\textbf{OUTPUT JSON SCHEMA (must match exactly):} \\
{\small\ttfamily
\noindent\{\\
\hspace*{1em}"contributions": [\\
\hspace*{2em}\{\\
\hspace*{3em}"name": "complete method type phrase (<=10 words, e.g. 'A gradient-based adversarial attack for ViTs')",\\
\hspace*{3em}"author\_claim\_text": "verbatim quote from paper (<=40 words)",\\
\hspace*{3em}"description": "normalized paraphrase (<=60 words)",\\
\hspace*{3em}"source\_hint": "location tag e.g. Abstract, Introduction §1, Conclusion"\\
\hspace*{2em}\}\\
\hspace*{1em}]\\
\}\par}

\vspace{5pt}
\textbf{INPUT:} \{paper\_text\}
\end{tcolorbox}
\caption{Auxiliary novelty prompt for extracting author-claimed contributions.}
\label{fig:novelty_contributions_prompt}
\end{figure}

\begin{figure}[H]
\centering
\begin{tcolorbox}[colback=gray!5!white, colframe=black!75, title=\textbf{Phase 3: LLM Judge — Novelty Claim Verification}]
\textbf{ROLE AND OBJECTIVE} \\
INSTRUCTION: \\
You are an impartial Judge that verifies whether the review sentence is a claim about the paper, and how it relates to the related work evidence. \\
Use ONLY the provided text. If the claim is too vague or evidence is missing, return \texttt{"insufficient"}.

\vspace{5pt}
\textbf{CLASSIFICATION}
\begin{itemize}
    \item \textbf{\texttt{claim}}: 1 if the sentence is a reviewer claim about the paper being reviewed; else 0.
    \item \textbf{\texttt{proof}}: 1 if the sentence provides evidence or support for a claim about the paper; else 0.
\end{itemize}

\vspace{5pt}
\textbf{AXIS 1 --- EVIDENCE SUPPORT (\texttt{stance\_alignment})}
\begin{itemize}
    \item \texttt{"aligned"}: reviewer claim aligns with and is supported by the related work evidence.
    \item \texttt{"partial"}: some relation exists but evidence is not conclusive.
    \item \texttt{"insufficient"}: claim is too vague, evidence is missing, or unverifiable.
    \item \texttt{"contradicted"}: evidence contradicts the reviewer claim or no supporting evidence was found.
\end{itemize}

\vspace{5pt}
\textbf{AXIS 2 --- CALIBRATION}
\begin{itemize}
    \item \texttt{"accurate"}: reviewer's strength of claim matches the actual evidence.
    \item \texttt{"overstated"}: reviewer claims too strongly given the evidence.
    \item \texttt{"understated"}: reviewer should have been stronger given the evidence.
    \item \texttt{"N/A"}: not applicable when evidence is insufficient to judge calibration.
\end{itemize}

\vspace{5pt}
\textbf{VERDICT SCALE ($[-2, +2]$)}
\begin{itemize}
    \item \textbf{$+2$ SUPPORTED}: Reviewer's novelty assessment aligns with the retrieved evidence.
    \item \textbf{$+1$ OVERSTATED}: Some relation exists, but the reviewer's ``same as / not novel'' claim is too strong.
    \item \textbf{$\phantom{+}0$ AMBIGUOUS}: Claim is too vague or unverifiable with the provided evidence.
    \item \textbf{$-1$ UNDERSTATED}: Reviewer misses very close prior work present in the candidate pool.
    \item \textbf{$-2$ UNSUPPORTED}: Evidence contradicts the claim, or no supporting evidence was found.
\end{itemize}

\vspace{5pt}
\textbf{OUTPUT FORMAT} \\
Return STRICT JSON only (no markdown, no code fences, no extra keys). \\
{\small\ttfamily
\noindent\{\\
\hspace*{1em}"review\_sentence\_id": "S\_001",\\
\hspace*{1em}"related\_paper\_id": "P123",\\
\hspace*{1em}"classification": \{"claim": 1, "proof": 0\},\\
\hspace*{1em}"stance\_alignment": "aligned",\\
\hspace*{1em}"calibration": "accurate",\\
\hspace*{1em}"score": 2,\\
\hspace*{1em}"label": "SUPPORTED",\\
\hspace*{1em}"explanation": "Short explanation"\\
\}\par}

\vspace{5pt}
\textbf{INPUTS:} \{review\_sentence\} \quad \{paper\_abstract\_intro\} \quad \{related\_work\_title\_abstract\}
\end{tcolorbox}
\caption{PRISM Novelty Assessment Prompt (2/2): LLM Judge verifying each novelty claim against a retrieved related-work candidate on a five-level evidence scale.}
\label{fig:novelty_prompt_phase3}
\end{figure}

\paragraph{Running Example: Novelty Assessment.}

To illustrate the full Novelty Assessment pipeline, consider a human review of a NeurIPS~2025 oral paper (\texttt{qYkhCah8OZ}), \textit{Boosting Knowledge Utilization in Multimodal Large Language Models via Adaptive Logits Fusion and Attention Reallocation}. The benchmark concatenates all human reviews for the same submission; Phase~1 extracts three novelty claims from distinct reviewers:
\begin{itemize}
    \item \textbf{C1} (\textit{not\_novel}, from Reviewer~1's Weaknesses): \textit{``The proposed method appears incremental, as the techniques involving attention weighting and logits fusion are already well-known and mainly borrowed from previous works.''}
    
    \item \textbf{C2} (\textit{novel}, from Reviewer~2's Strengths): \textit{``The proposed two modules, attention reallocation and adaptive logits fusion, offer a novel and effective perspective to enhance MLLM performance in knowledge-intensive tasks.''}
    
    \item \textbf{C3} (\textit{somewhat\_novel}, from Reviewer~4's Originality assessment): \textit{``The paper proposes a novel approach to reallocate attentions and fuse knowledge, although similar ideas for attention reallocations have been introduced by earlier work.''}
\end{itemize}

Phase~2 issues structured queries to the Semantic Scholar API, retrieving 20 candidate related works. We report the three most relevant per the top-3 relevance-weighted aggregation policy: (RW1) MambaTrans: Multimodal Fusion Image Translation via LLM Priors; (RW2) Can Multimodal LLMs be Guided to Improve Industrial Anomaly Detection?; and (RW3) CAT+: Investigating and Enhancing Audio-Visual Understanding in LLMs.

Phase~3 evaluates each (claim, related-work) pair via the LLM Judge:
\begin{itemize}
    \item \textbf{C1} (\textit{not\_novel}): RW1 is off-topic (image fusion, not attention reallocation) and ``does not contain information about attention weighting or logits fusion techniques'' - Unsupported ($-2$). RW2 corroborates that attention weighting is a known technique, as its related work section discusses MLLM enhancement via attention mechanisms - Supported ($+2$). RW3 similarly confirms these are established techniques in multimodal retrieval augmented generation - Supported ($+2$).
    
    \item \textbf{C2} (\textit{novel}): RW1 again ``does not contain any information about the paper being reviewed'' - Unsupported ($-2$). RW2 ``discusses a novel multi-expert framework'' for MLLM tasks, validating the novelty claim - Supported ($+2$). RW3 ``explicitly states [a] proposed [module] addressing audio-visual understanding enhancement in LLMs, aligning with the novelty claim'' - Supported ($+2$).
    
    \item \textbf{C3} (\textit{somewhat\_novel}): All three related works corroborate the nuanced stance. RW1 discusses ``multimodal fusion with attention, supporting the claim that similar ideas exist'' - Supported ($+2$). RW2 ``demonstrates that attention-based approaches are common, supporting the `somewhat novel' stance'' - Supported ($+2$). RW3 ``proposes attention-based enhancement for LLMs, confirming that similar attention reallocation ideas have been introduced by earlier work'' - Supported ($+2$).
\end{itemize}

The Novelty Verification Score is the mean aggregated score over all $K$ novelty claims, where each claim's aggregated score $s_k$ is the relevance-weighted average of its top-3 pair verdicts $v_{k,j}$. Throughout the main results, this top-3 relevance-weighted policy is the adopted default; a max-pooled variant is treated only as a sensitivity check because it typically produces more optimistic scores. Because the raw verdicts lie on a $[-2, +2]$ scale, the raw review-level mean $\bar{s}(R)$ is linearly normalized into the final $NS(R) \in [0,1]$ via:
$$\bar{s}(R) = \frac{1}{K} \sum_{k=1}^{K} s_k, \quad s_k = \sum_{j=1}^{3} w_j \cdot v_{k,j}, \quad NS(R) = \frac{\bar{s}(R) + 2}{4}$$

\textit{Calculation for the Running Example:} With $K = 3$ claims and equal relevance weights ($w_j = 1/3$):
$$s_{C_1} = \frac{(-2) + 2 + 2}{3} = 0.667, \quad s_{C_2} = \frac{(-2) + 2 + 2}{3} = 0.667, \quad s_{C_3} = \frac{2 + 2 + 2}{3} = 2.0$$
$$\bar{s}(R) = \frac{0.667 + 0.667 + 2.0}{3} = \frac{3.333}{3} = 1.111, \quad NS(R) = \frac{1.111 + 2}{4} = \mathbf{0.778}$$

This example illustrates three key behaviors of the pipeline. First, the \textit{somewhat\_novel} claim (C3) achieves the highest per-claim score because its nuanced stance (``novel\ldots{} although similar ideas exist'') is confirmed by all retrieved evidence, demonstrating that calibrated claims are rewarded. Second, the \textit{not\_novel} (C1) and \textit{novel} (C2) claims receive identical aggregated scores ($0.667$) despite opposing stances, because the same off-topic related work (MambaTrans, which is about image fusion rather than MLLM knowledge utilization) penalizes both equally, highlighting that retrieval quality, not just claim content, drives the verdict. Third, the overall normalized score $NS(R)=0.778$ reflects a review whose novelty assessments are partially well-grounded but sensitive to the composition of the retrieval pool.

\paragraph{Dimension 3: Flaw Identification \& Prioritization.}

The evaluation of Flaw Identification and Prioritization is implemented as a two-phase pipeline that cross-validates reviewer arguments against the manuscript and quantifies both detection coverage and critical issue ordering. The first phase, illustrated in Figure~\ref{fig:flaw_prompt_phase1}, consolidates raw review texts from multiple reviewers (Human and LLM) by atomizing and grouping arguments into a structured inventory of \emph{Micro-flaws}, each categorized within a hierarchical taxonomy of seven macro-topics (e.g., \textit{Novelty \& Contribution}, \textit{Methodology \& Theoretical Soundness}, \textit{Experimental Design \& Evaluation}). The second phase, presented in Figure~\ref{fig:flaw_prompt_phase2}, acts as an independent LLM meta-reviewer that verifies each Micro-flaw against the paper text, assigning a binary validity label (\texttt{is\_valid}) and a severity rating (\texttt{Critical} or \texttt{Minor}) according to a predefined ontology grounded in whether fixing the issue requires new experiments or is purely editorial.

From these two phases, three complementary metrics are derived. \textbf{Critical Recall} and \textbf{Minor Recall} measure the proportion of pooled verified Critical and Minor flaws, respectively, that a given reviewer successfully identified, enabling fine-grained diagnosis of detection coverage across severity strata. The primary ranking metric, the \textbf{normalized Critique Prioritization Score (nCPS)}, adapts the standard NDCG formulation~\cite{NDCG} to capture whether a reviewer front-loads their most severe critiques within each structural section of their review.

\begin{figure}[H]
\centering
\begin{tcolorbox}[colback=gray!5!white, colframe=black!75, title=\textbf{Phase 1: Micro-flaw Atomization and Grouping}]
\textbf{ROLE AND OBJECTIVE} \\
You are an expert meta-reviewer for top-tier computer science conferences. Analyze raw review texts from multiple reviewers and consolidate their arguments into a structured list of unique \emph{Micro-flaws}.

\vspace{5pt}
\textbf{GROUPING RULES (STRICTLY ENFORCED)}
\begin{enumerate}
    \item \textbf{Conceptual Consistency (Must Split):} Arguments grouped into the same Micro-flaw MUST address the same fundamental problem. Do not merge distinct scientific issues merely because they share a broad topic.
    \item \textbf{Allowed Aggregation (Can Group):} Arguments MAY be grouped if they point to the exact same specific error in the paper (e.g., multiple reviewers citing the same missing baseline).
    \item \textbf{No Forced Fit:} If an argument does not fit any existing Micro-flaw precisely, create a new one.
    \item \textbf{No Upper Bound:} There is no limit on the number of Micro-flaws. Multiple Micro-flaws may share the same Macro-topic.
\end{enumerate}

\vspace{5pt}
\textbf{TAXONOMY (7 Macro-topics):} Novelty \& Contribution; Clarity \& Presentation; Applicability, Scalability \& Limitations; Experimental Design \& Evaluation; Related Work \& Citations; Methodology \& Theoretical Soundness; Reproducibility \& Open Science.

\vspace{5pt}
\textbf{OUTPUT FORMAT} \\
Respond ONLY with a valid JSON object.

\vspace{5pt}
\textbf{INPUT REVIEWS:} \{input\_text\}
\end{tcolorbox}
\caption{PRISM Flaw Identification Prompt (1/2): Atomizing and grouping reviewer arguments into a canonical Micro-flaw inventory.}
\label{fig:flaw_prompt_phase1}
\end{figure}

\begin{figure}[H]
\centering
\begin{tcolorbox}[colback=gray!5!white, colframe=black!75, title=\textbf{Phase 2: Meta-reviewer Validity and Severity Judgement}]
\textbf{ROLE AND OBJECTIVE} \\
You are a strict and objective Meta-Reviewer. Given the full paper text and a JSON list of Micro-flaws raised by reviewers, independently verify each flaw against the manuscript.

\vspace{5pt}
\textbf{FOR EACH MICRO-FLAW, ANSWER:}
\begin{enumerate}
    \item \textbf{\texttt{is\_valid} (True/False):} Does this flaw genuinely exist in the paper? Return \texttt{False} if it is a hallucination, misunderstanding, or unreasonable request.
    \item \textbf{\texttt{severity} ("Critical" / "Minor"):} If valid, assign severity per the ontology below.
\end{enumerate}

\vspace{5pt}
\textbf{SEVERITY ONTOLOGY}
\begin{itemize}
    \item \textbf{Critical} ($w=2$): Flaws requiring new experiments, proof revisions, or core claim changes — covering Methodology, Experimental Design, Novelty, and severe Reproducibility/Applicability issues.
    \item \textbf{Minor} ($w=1$): Flaws fixable via textual/editorial revision — covering Clarity \& Presentation, missing citations, or documentation gaps.
    \item \textbf{Borderline rule:} Prefer Critical if fixing the issue can plausibly alter main conclusions; prefer Minor if the fix is purely editorial.
\end{itemize}

\vspace{5pt}
\textbf{OUTPUT FORMAT} \\
Respond ONLY with a valid JSON object.

\vspace{5pt}
\textbf{INPUT:} \{paper\_text\} \quad \{micro\_flaws\_json\}
\end{tcolorbox}
\caption{PRISM Flaw Identification Prompt (2/2): LLM meta-reviewer verifying flaw validity and assigning severity labels for the pooled verified-flaw reference. The verifier receives the de-duplicated flaw list without the identity or number of the reviewers who raised each flaw.}
\label{fig:flaw_prompt_phase2}
\end{figure}

\paragraph{Running Example: Flaw Identification \& Major Issues Prioritization.}

To illustrate both the Flaw Identification and Prioritization pipelines, consider a
reviewer evaluating a graph neural network paper. The \textbf{pooled verified-flaw reference} (``GT'' below),
defined as the union of all valid flaws identified across all reviewers for this paper,
consists of:

\begin{itemize}
    \item \textbf{Critical flaws (GT):}
    \begin{itemize}
        \item FC1: \textit{``Missing comparison to GraphSAGE baseline in Table 2.''}
        \item FC2: \textit{``Convergence proof contains a gap in Lemma 3: the Lipschitz
        assumption is invoked but never verified.''}
        \item FC3: \textit{``No ablation study on the effect of message-passing depth.''}
    \end{itemize}
    \item \textbf{Minor flaws (GT):}
    \begin{itemize}
        \item FM1: \textit{``Inconsistent notation: $\mathbf{A}$ and $\tilde{\mathbf{A}}$
        used interchangeably in Eq.~4 and Eq.~5.''}
        \item FM2: \textit{``Figure 2 axes are unlabeled.''}
        \item FM3: \textit{``Related work omits Liu et al.~(2022).''}
    \end{itemize}
\end{itemize}

Reviewer $X$ produces the following flaw list, in the order they appear in the review:

\begin{enumerate}
    \item \textbf{[Minor]} FM1 --- inconsistent notation $\checkmark$
    \item \textbf{[Critical]} FC1 --- missing GraphSAGE comparison $\checkmark$
    \item \textbf{[Minor]} FM2 --- Figure 2 axes unlabeled $\checkmark$
    \item \textbf{[Critical]} FC2 --- convergence proof gap $\checkmark$
\end{enumerate}

\noindent\textit{Calculation for Flaw Identification:}

Reviewer $X$ matched 2 out of 3 critical flaws (missed FC3) and 2 out of 3 minor flaws
(missed FM3):
$$\text{Critical Recall} = \frac{|\{\text{FC1, FC2}\}|}{|\{\text{FC1, FC2, FC3}\}|}
= \frac{2}{3} \approx 0.667$$
$$\text{Minor Recall} = \frac{|\{\text{FM1, FM2}\}|}{|\{\text{FM1, FM2, FM3}\}|}
= \frac{2}{3} \approx 0.667$$

\noindent\textit{Calculation for Major Issue Prioritization (nCPS):}

The $nCPS$ is computed over all $k$ GT-matched valid flaws identified by
Reviewer~$X$, ordered by their position of appearance in the review.
Each flaw receives a position discount $\log_2(p_i+1)$ and a severity
weight $w_i \in \{2,1\}$ for Critical/Minor respectively.
The GT-matched valid flaws are: FM1 (Minor, position~1), FC1 (Critical,
position~2), FM2 (Minor, position~3), FC2 (Critical, position~4).

\begin{align*}
CPS &= \frac{w_{\text{FM1}}}{\log_2(1+1)}
     + \frac{w_{\text{FC1}}}{\log_2(2+1)}
     + \frac{w_{\text{FM2}}}{\log_2(3+1)}
     + \frac{w_{\text{FC2}}}{\log_2(4+1)} \\
    &= \frac{1}{\log_2 2}
     + \frac{2}{\log_2 3}
     + \frac{1}{\log_2 4}
     + \frac{2}{\log_2 5} \\
    &= 1.000 + 1.262 + 0.500 + 0.861 \;=\; 3.623
\end{align*}

The ideal score $iCPS$ places all Critical flaws first, then Minor:

\begin{align*}
iCPS &= \frac{2}{\log_2 2}
      + \frac{2}{\log_2 3}
      + \frac{1}{\log_2 4}
      + \frac{1}{\log_2 5} \\
     &= 2.000 + 1.262 + 0.500 + 0.431 \;=\; 4.193
\end{align*}

$$nCPS = \frac{CPS}{iCPS} = \frac{3.623}{4.193} \approx \mathbf{0.864}$$

\paragraph{Dimension 4: Multi-dimensional Constructiveness.}

To minimize cognitive overload and ensure high-fidelity scoring, the evaluation of constructiveness is decoupled into a two-phase pipeline. Phase 1, detailed in Figure \ref{fig:constructiveness_phase1}, decomposes the holistic peer review into discrete, non-overlapping Atomic Review Comments (ARCs) while retaining their original context via anchor quotes. Phase 2, illustrated in Figure \ref{fig:constructiveness_phase2}, then evaluates each isolated ARC against the five granular dimensions of constructiveness (D1-D5) using a stringent $[0, 2]$ rubric.

\begin{figure}[H]
\centering
\begin{tcolorbox}[colback=gray!5!white, colframe=black!75, title=\textbf{Phase 1: Atomic Review Comment (ARC) Extraction}]
\textbf{ROLE AND OBJECTIVE} \\
You are an expert peer-review analyst. Your task is to decompose a comprehensive peer review into distinct Atomic Review Comments (ARCs). 

\vspace{5pt}
\textbf{EXTRACTION RULES}
\begin{enumerate}
    \item Extract ALL distinct points from Summary, Strengths, Weaknesses, Questions, and Suggestions.
    \item \textbf{One point per ARC:} If a single sentence contains two critiques, split it into two separate ARCs.
    \item \textbf{Anchor Quote:} Provide a verbatim 5-25 word substring copied EXACTLY from the review to anchor the comment.
    \item \textbf{Comment Type:} Classify each ARC as exactly one of: \texttt{weakness}, \texttt{strength}, \texttt{question}, \texttt{suggestion}, or \texttt{observation}.
\end{enumerate}

\vspace{5pt}
\textbf{OUTPUT FORMAT} \\
Respond ONLY with a valid JSON object.

\vspace{5pt}
\textbf{INPUT REVIEW TEXT:} \\
\{raw\_review\_text\}
\end{tcolorbox}
\caption{PRISM Constructiveness Prompt (1/2): Decomposing the review into distinct Atomic Review Comments (ARCs).}
\label{fig:constructiveness_phase1}
\end{figure}

\begin{figure}[H]
\centering
\begin{tcolorbox}[colback=gray!5!white, colframe=black!75, title=\textbf{Phase 2: Multi-dimensional Constructiveness Scoring}]
\textbf{ROLE AND OBJECTIVE} \\
You are an expert peer-review analyst. Given the full peer review as macro-context and a list of extracted Atomic Review Comments (ARCs), score EACH comment on 5 dimensions.

\vspace{5pt}
\textbf{SCORING RUBRIC (Score 0, 1, or 2 for each)}
\begin{itemize}
    \item \textbf{D1\_actionability} — Can the author act on this?
        \begin{itemize}
            \item \textbf{0}: Opinion with no guidance (e.g., "poorly written")
            \item \textbf{1}: General direction (e.g., "needs more baselines")
            \item \textbf{2}: Specific, implementable (e.g., "add [MethodX] on CIFAR-10")
        \end{itemize}
    \item \textbf{D2\_specificity} — References concrete paper elements?
        \begin{itemize}
            \item \textbf{0}: Vague (e.g., "has issues")
            \item \textbf{1}: Semi-specific (e.g., "methodology section unclear")
            \item \textbf{2}: Pinpoints exact element (e.g., "Eq 7 in Sec 4.2 missing term")
        \end{itemize}
    \item \textbf{D3\_justification} — Evidence-backed?
        \begin{itemize}
            \item \textbf{0}: Bare assertion (e.g., "not novel")
            \item \textbf{1}: Partial reasoning (e.g., "similar to prior work on X")
            \item \textbf{2}: Full evidence (e.g., "same loss as [Author2020] Eq 3")
        \end{itemize}
    \item \textbf{D4\_solution} — Suggests improvements?
        \begin{itemize}
            \item \textbf{0}: Problem-only (e.g., "baselines weak")
            \item \textbf{1}: Implicit fix (e.g., "lacks recent SOTA")
            \item \textbf{2}: Explicit fix (e.g., "add [M2023] achieving X\%")
        \end{itemize}
    \item \textbf{D5\_tone} — Respectful?
        \begin{itemize}
            \item \textbf{0}: Hostile / dismissive
            \item \textbf{1}: Neutral, factual
            \item \textbf{2}: Professional-constructive, encouraging
        \end{itemize}
\end{itemize}

\vspace{5pt}
\textbf{OUTPUT FORMAT} \\
Respond ONLY with a valid JSON object containing the \texttt{arc\_id} and the numerical scores for \texttt{D1} through \texttt{D5}.

\vspace{5pt}
\textbf{MACRO-CONTEXT:} \{raw\_review\_text\} \\
\textbf{LIST OF ARCS:} \{arc\_json\_list\}
\end{tcolorbox}
\caption{PRISM Constructiveness Prompt (2/2): Evaluating isolated ARCs against the five-dimensional constructiveness rubric.}
\label{fig:constructiveness_phase2}
\end{figure}

\paragraph{Running Example: Multi-dimensional Constructiveness.}

To illustrate the MCS pipeline, consider the following excerpt from a human review 
of a theoretical machine learning paper:
\textit{``The paper lacks a clear comparison of its theoretical results (Table 1, Section 5) 
with prior related work. No experimental results. The toy example should correspond to the 
motivation example. Provide concrete toy examples illustrating setup and theorems, including 
specific distributions and query complexity bounds. A detailed comparison to existing results 
in the bandits literature is needed.''}

Processing this text through Gemini yields the following Atomic Review Comments (ARCs):
\begin{itemize}
    \item \textbf{ARC 1 (Weakness):} \textit{``Lacks clear comparison of theoretical results 
    (Table 1, Sec. 5) with prior related work.''} \\
    $\rightarrow$ \textbf{Scores:} D1=1 (identifies gap but no specific fix), D2=2 (names Table 1, Sec. 5), 
    D3=1 (partial reasoning), D4=0 (no fix), D5=1 (neutral tone).
    
    \item \textbf{ARC 2 (Weakness):} \textit{``No experimental results to validate theoretical findings.''} \\
    $\rightarrow$ \textbf{Scores:} D1=2 (clear: add experiments), D2=2 (specific), 
    D3=0 (no justification why), D4=1 (implicit: conduct experiments), D5=1 (factual).
    
    \item \textbf{ARC 3 (Question):} \textit{``Provide concrete toy examples with distributions 
    and query complexity bounds.''} \\
    $\rightarrow$ \textbf{Scores:} D1=2 (actionable), D2=2 (specific), D3=0 (no WHY explained), 
    D4=1 (suggested fix), D5=2 (professional-constructive framing).
    
    \item \textbf{ARC 4 (Weakness):} \textit{``Detailed comparison to existing bandit results needed.''} \\
    $\rightarrow$ \textbf{Scores:} D1=2 (actionable), D2=2 (references bandit literature), 
    D3=0 (no justification), D4=0 (problem-only), D5=1 (neutral).
\end{itemize}

Aggregating dimension-wise over these 4 ARCs:
$$\overline{D_1} = \frac{1 + 2 + 2 + 2}{4} = 1.75, \quad 
\overline{D_2} = \frac{2 + 2 + 2 + 2}{4} = 2.00$$
$$\overline{D_3} = \frac{1 + 0 + 0 + 0}{4} = 0.25, \quad 
\overline{D_4} = \frac{0 + 1 + 1 + 0}{4} = 0.50$$
$$\overline{D_5} = \frac{1 + 1 + 2 + 1}{4} = 1.25$$

The Multi-dimensional Constructiveness Score is the normalized average:
$$\text{MCS} = \frac{1}{5 \times 2} \left( \overline{D_1} + \overline{D_2} + \overline{D_3} 
+ \overline{D_4} + \overline{D_5} \right)$$

\textit{Calculation for the Running Example:}
$$\text{MCS} = \frac{1.75 + 2.00 + 0.25 + 0.50 + 1.25}{10} = \frac{5.75}{10} = \mathbf{0.575}$$

\subsection{Human Validation Protocol}
\label{sec:app_human_validation}

To rigorously validate our evaluation engine, we conducted a targeted human evaluation on a subset of 30 papers. We recruited 20 senior researchers, including doctoral researchers,
senior researchers, and professors, with prior reviewing experience at venues such as ICLR, ICML, NeurIPS. We bypassed the initial extraction phases; granular components (Argumentative Discourse Units, Novelty Claims, Flaw Arguments, Atomic Review Comments) were pre-extracted. 

Our validation scale is comparable to, and in terms of annotator coverage extends, those adopted in prior work on automated peer review. For example, TreeReview~\cite{chang-etal-2025-treereview} conducted human evaluation on 20 papers with five experts, while ReviewEval~\cite{garg-etal-2025-revieweval} evaluated 30 papers with five experts. Our study therefore operates within a similar paper-level evaluation scale while involving a broader pool of human annotators.

To focus the study specifically on the reliability of the evaluation stage, we bypassed the initial extraction phases and provided annotators with pre-extracted granular components, including Argumentative Discourse Units (ADUs), novelty claims, flaw arguments, and Atomic Review Comments (ARCs). 
Requiring annotators to independently repeat the extraction process would substantially increase the annotation burden, as each paper yields numerous fine-grained units across multiple evaluation dimensions, each of which must subsequently be assessed under dimension-specific criteria. More importantly, using pre-extracted components controls for extraction variability: human and LLM evaluators assess identical inputs, so their agreement reflects differences in evaluation rather than differences in what each evaluator extracts from the review. Annotators were then asked to score these pre-segmented inputs using the same criteria provided to our LLM evaluator:

\begin{itemize}
    \item \textbf{Depth of Analysis:} Annotators evaluate pre-segmented ADUs through three tasks: (1) Role Classification (Claim vs. Premise), (2) Aspect Classification (assigning one aspect topic), and (3) Grounding Score (0/1/2, for premises only).
    \item \textbf{Novelty Assessment:} Given pre-extracted novelty claims and pre-retrieved related papers, annotators assign a literature-grounded score (-2 to +2) for each claim.
    \item \textbf{Flaw Identification:} Annotators work pairwise (Human Reviewer vs. LLM Reviewer), creating a local flaw pool, assigning validity labels, and classifying severity (Minor/Critical) alongside aspect topics.
    \item \textbf{Multi-dimensional Constructiveness:} Using pre-extracted ARCs, annotators score each unit across five dimensions (Actionability, Specificity, Justification, Solution, Tone) on a 0/1/2 scale.
\end{itemize}
Table~\ref{tab:human_validation} in Section~\ref{sec:human_validation} reports the resulting Cohen's $\kappa$/$\kappa_w$ and exact/$\pm1$ agreement rates between annotators and \texttt{Gemini 2.5 Flash Lite} for each task above.

\section{Metric Independence Analysis via Pearson Correlation}
\label{sec:app_exp1}
\subsection{Motivation and Objective}

A key requirement for a multi-dimensional evaluation benchmark is that its 
constituent metrics should capture \textit{distinct, non-overlapping} aspects 
of review quality. If two metrics were highly correlated, they would convey 
redundant information and effectively reduce the dimensionality of the evaluation, 
undermining the claim that different facets of peer review are independently 
assessed. To verify this property, we conduct a pairwise Pearson correlation 
analysis across the five evaluation dimensions of our benchmark: 
Depth of Analysis (\textbf{DoA}$_{\text{HM}}$), Novelty Assessment (\textbf{NS}), 
Flaw Identification (\textbf{Critical Recall}, \textbf{Minor Recall}), 
Issue Prioritization (\textbf{nCPS}) and Multi-dimensional Constructiveness 
(\textbf{MCS}).

\subsection{Statistical Method}

\paragraph{Pearson Correlation Coefficient.}
For two metric vectors $\mathbf{x} = (x_1, \ldots, x_n)$ and 
$\mathbf{y} = (y_1, \ldots, y_n)$ measured over $n$ paper-review samples, 
the Pearson correlation coefficient is defined as:

\begin{equation}
    r_{xy} = \frac{\displaystyle\sum_{i=1}^{n}(x_i - \bar{x})(y_i - \bar{y})}
                  {\sqrt{\displaystyle\sum_{i=1}^{n}(x_i - \bar{x})^2}
                  \;\cdot\;
                   \sqrt{\displaystyle\sum_{i=1}^{n}(y_i - \bar{y})^2}}
    \label{eq:pearson}
\end{equation}

\noindent where $r_{xy} \in [-1, 1]$. A value of $r = 0$ indicates no linear 
association; $|r| = 1$ indicates perfect linear dependence.

\paragraph{Significance Test.}
To assess whether an observed $r_{xy}$ differs significantly from zero, we apply 
the two-tailed \textit{t}-test under the null hypothesis $H_0\colon \rho = 0$ 
(no linear correlation in the population). The test statistic is:

\begin{equation}
    t = \frac{r_{xy}\,\sqrt{n - 2}}{\sqrt{1 - r_{xy}^2}}
    \label{eq:pearson_ttest}
\end{equation}

\noindent which follows a Student's $t$-distribution with $n - 2$ degrees of 
freedom under $H_0$. We report two-tailed $p$-values with significance thresholds 
$p < 0.001$ (***), $p < 0.01$ (**), $p < 0.05$ (*), and label 
non-significant results as \textit{ns}.

\paragraph{Effect Size Interpretation.}
Statistical significance alone is insufficient because large samples can render 
even trivially small correlations significant. We therefore assess the 
\textit{practical magnitude} of each $|r_{xy}|$ using the conventional 
thresholds of~\citep{cohen1988statistical}: $|r| < 0.10$ (negligible), $0.10 \leq |r| < 0.30$ (small), $0.30 \leq |r| < 0.50$ (moderate) and $|r| \geq 0.50$ (large).
Only correlations that are both statistically significant \textit{and} of 
moderate-to-large magnitude are considered substantively meaningful.

\subsection{Results and Discussion}

\begin{figure}[htbp]
    \centering
    \includegraphics[width=0.82\linewidth]{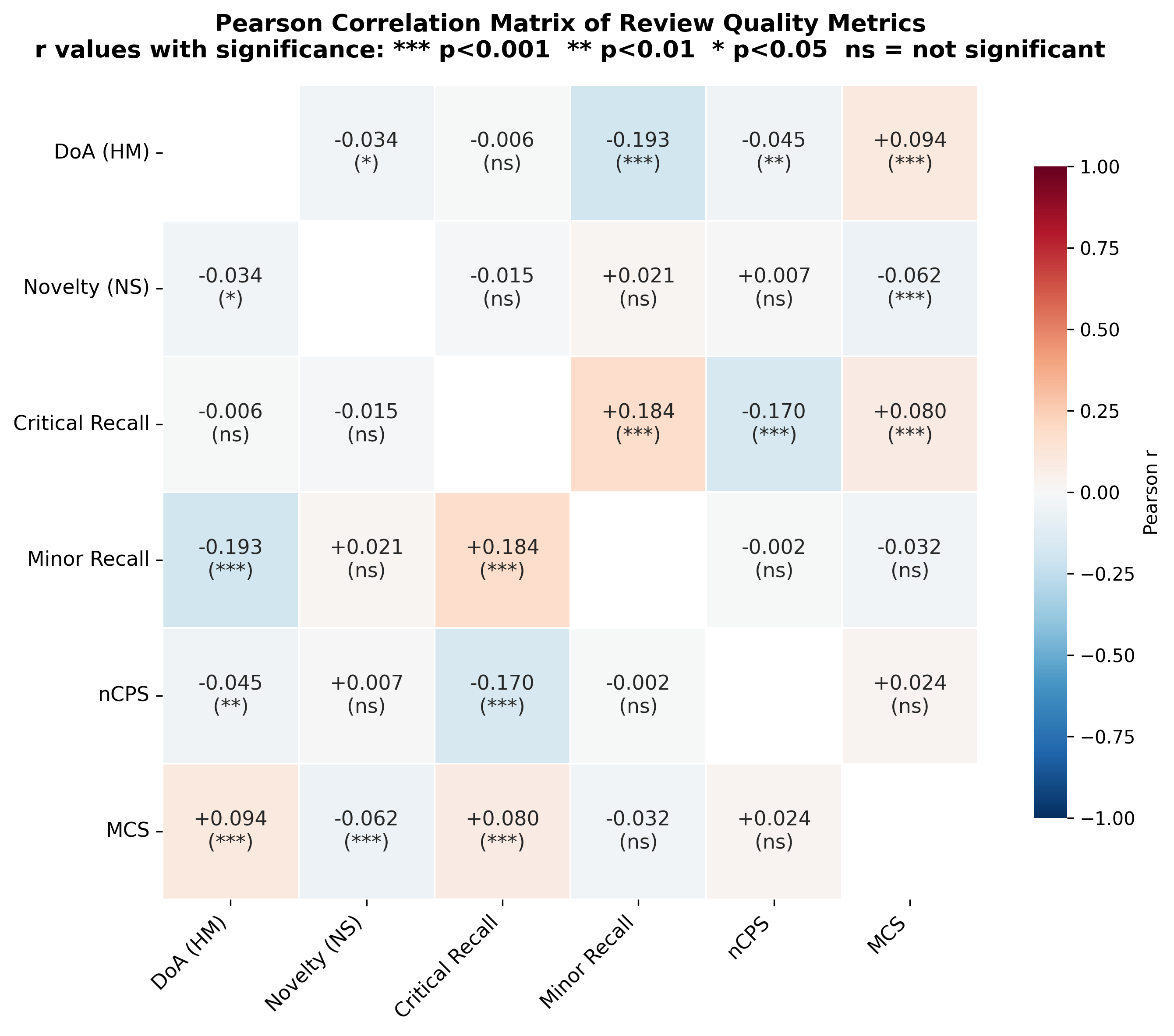}
    \caption{Pearson correlation matrix of the six review quality metrics. 
    Each cell reports the Pearson $r$ coefficient with significance annotation: 
    $^{***}p<0.001$, $^{**}p<0.01$, $^{*}p<0.05$; unmarked cells are 
    not significant ($p \geq 0.05$). Blue shading denotes negative correlation; 
    red shading denotes positive correlation.}
    \label{fig:corr_heatmap}
\end{figure}

Figure~\ref{fig:corr_heatmap} presents the full pairwise Pearson correlation 
matrix across all six metrics. The results consistently show very weak 
inter-metric associations, with a maximum absolute coefficient of 
$|r|_{\max} = 0.193$.

\noindent\textbf{Cross-dimension independence.}
The most critical finding concerns correlations \textit{across} evaluation 
dimensions. \textbf{Novelty (NS)} shows no significant association with 
any flaw-related metric (Critical Recall: $r = -0.015$, $p = 0.40$; 
Minor Recall: $r = +0.021$, $p = 0.23$; nCPS: $r = +0.007$, $p = 0.70$), 
confirming that evaluating the novelty of reviewer claims is entirely 
decoupled from flaw detection ability. 
\textbf{DoA}$_{\text{HM}}$ likewise exhibits no meaningful linear relationship 
with Critical Recall ($r = -0.006$, $p = 0.72$), demonstrating that 
structural argumentation depth is independent of a reviewer's capacity 
to identify methodological flaws.
The correlation between DoA$_{\text{HM}}$ and MCS is marginally significant 
($r = +0.094$, $p < 0.001$), yet the effect size remains negligible 
($r^2 < 0.01$), confirming that argumentative depth and constructiveness constitute distinct dimensions.

\noindent\textbf{Overall assessment.}
Seven of fifteen metric pairs show no statistically significant correlation ($p \geq 0.05$). All significant pairs have $|r| < 0.20$, placing them in the \textit{negligible-to-small} range with shared variance below $4\%$ ($r^2 < 0.04$). These results collectively confirm that the six metrics are \textbf{empirically near-orthogonal}: each captures a distinct dimension of peer review quality, thereby justifying their joint use as a comprehensive multi-dimensional evaluation benchmark.

\section{Full Cross-Dataset Quantitative Results}
\label{app:app_exp2}

\subsection{Statistical Significance Testing Protocol}
\label{sec:statistical_protocol}

To rigorously assess the performance differences between the LLM baselines and the human reviews, we conduct non-parametric statistical testing across all metrics. Given the non-normal distribution of the evaluation scores, we employ the \textbf{Wilcoxon signed-rank test} to compute the uncorrected $p$-values for paired comparisons.

Furthermore, to stringently control the Family-Wise Error Rate (FWER) while evaluating the cross-venue consistency of the models, we construct our hypothesis families based on model-metric pairs. Specifically, the \textbf{Holm-Bonferroni step-down correction} is applied independently for each LLM baseline within each specific evaluation dimension across the 3 conferences ($N=3$ comparisons per family, corresponding to the 3 venues used throughout this paper: ICLR 2026, ICML 2025, and NeurIPS 2025). This approach ensures that any statistically significant result reflects a model's robust and consistent capability across different peer-review distributions, rather than an isolated success at a single venue.

Throughout the subsequent tables, we report the effect size (rank-biserial correlation $r$) and denote the Holm-corrected statistical significance using the following standard notation:
\begin{itemize}
    \item \textbf{ns} (Not Significant): $p_{holm} \ge 0.05$
    \item $\mathbf{*}$ : $p_{holm} < 0.05$
    \item $\mathbf{**}$ : $p_{holm} < 0.01$
    \item $\mathbf{***}$ : $p_{holm} < 0.001$
\end{itemize}
Results marked as \textbf{ns} suggest that the model's performance is statistically indistinguishable from human reviews, whereas the starred results indicate a robust difference that survives rigorous multiple-comparison correction.

\subsection{Depth of Analysis}
\label{sec:detail_doa}

\noindent\begin{minipage}{0.42\textwidth}
  \centering
  \captionof{table}{Core Depth of Analysis metrics (macro-average, primary judge).}
  \label{tab:doa_core}
  \small
  \resizebox{\linewidth}{!}{
  \begin{tblr}{
    column{1} = {l},
    column{2-3} = {c},
    hline{1,8} = {-}{0.08em},
    hline{2} = {-}{0.05em},
    hline{3} = {-}{dashed},
  }
  \textbf{System} & \textbf{Prem.\ Ratio} & \textbf{Grounding} \\
  \textbf{Human} & $0.557 \pm 0.091$ & $0.481 \pm 0.062$ \\
  CycleReviewer & $\mathbf{0.614} \pm 0.185$ & $0.437 \pm 0.137$ \\
  DeepReview & $0.588 \pm 0.177$ & $0.446 \pm 0.134$ \\
  Reviewer2 & $0.376 \pm 0.185$ & $0.432 \pm 0.151$ \\
  SEA & $0.393 \pm 0.210$ & $\mathbf{0.464} \pm 0.119$ \\
  TreeReview & $0.302 \pm 0.136$ & $0.445 \pm 0.115$ \\
  \end{tblr}
  }
\end{minipage}
\hfill
\begin{minipage}{0.55\textwidth}
    \includegraphics[width=\linewidth]{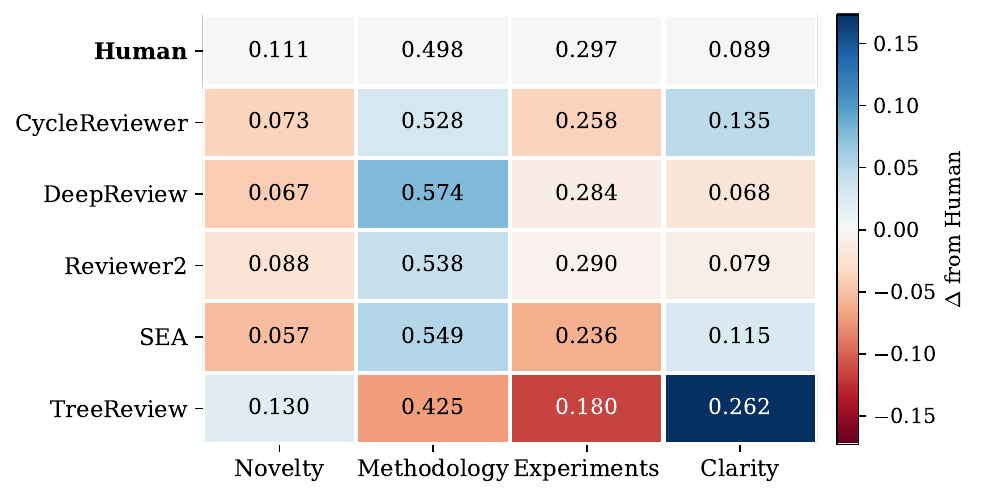}
    \par\vspace{-6px}
    \captionof{figure}{Aspect distributions of DoA arguments across four topics, colored by deviation from human reviews ($\Delta$). Advanced pipelines (DeepReview, CycleReviewer, Reviewer2, SEA) keep \textit{Clarity} close to human levels ($\sim 7$--$14\%$), whereas TreeReview spends $\sim 26\%$ of its effort on it.}
    \label{fig:doa_aspect}
\end{minipage}
\vspace{0.5em}

\begin{table}[H]
\centering
\caption{Detailed Cross-Venue Performance for \textbf{Depth of Analysis}. Statistical significance is computed independently for each baseline across the 3 conferences to evaluate consistency.}
\label{tab:doa_full}
\vspace{0.15cm}
\resizebox{\linewidth}{!}{%
\begin{tblr}{
  column{1} = {l},
  column{2-4} = {c},
  hline{1,9} = {-}{0.08em},
  hline{3} = {-}{0.05em},
  hline{4} = {-}{dashed},
}
\SetCell[r=2]{l}\textbf{Baselines} & \SetCell[c=3]{c}\textbf{Depth of Analysis Scores (Mean $\pm$ Std)} \\
 & \textbf{ICLR 2026} & \textbf{ICML 2025} & \textbf{NeurIPS 2025} \\
\textbf{Human} & $0.497 \pm 0.061$ & $0.483 \pm 0.063$ & $0.500 \pm 0.059$ \\
CycleReviewer & $\mathbf{0.479 \pm 0.121}$\textsuperscript{ns} & $0.476 \pm 0.145$\textsuperscript{ns} & $\mathbf{0.494 \pm 0.138}$\textsuperscript{ns} \\
DeepReview & $0.471 \pm 0.142$\textsuperscript{ns} & $\mathbf{0.482 \pm 0.126}$\textsuperscript{ns} & $0.490 \pm 0.128$\textsuperscript{ns} \\
Reviewer2 & $0.346 \pm 0.146^{***}$ & $0.363 \pm 0.135^{***}$ & $0.379 \pm 0.128^{***}$ \\
SEA & $0.382 \pm 0.161^{***}$ & $0.403 \pm 0.139^{***}$ & $0.407 \pm 0.152^{***}$ \\
TreeReview & $0.335 \pm 0.115^{***}$ & $0.335 \pm 0.123^{***}$ & $0.346 \pm 0.111^{***}$ \\
\end{tblr}
}
\end{table}

The granular DoA results presented in Table \ref{tab:doa_full} highlight a stark contrast in the evidentiary capabilities of the evaluated baselines. TreeReview, SEA, and Reviewer2 yield consistently lower DoA scores across all three venues, with statistical significance underscoring their systematic deficiency in substantiating critiques compared to human reviewers. Conversely, CycleReviewer and DeepReview successfully bridge this gap. The consistent lack of statistical significance (\textsuperscript{ns}) when compared to human reviews indicates that these models achieve a comparable level of analytical depth. As analyzed previously, this statistical parity is primarily driven by their robust internal grounding mechanisms and high premise ratios, which effectively compensate for the inherent limitations of standard LLM generation.

\begin{figure}[H]
\centering
    \includegraphics[width=\linewidth]{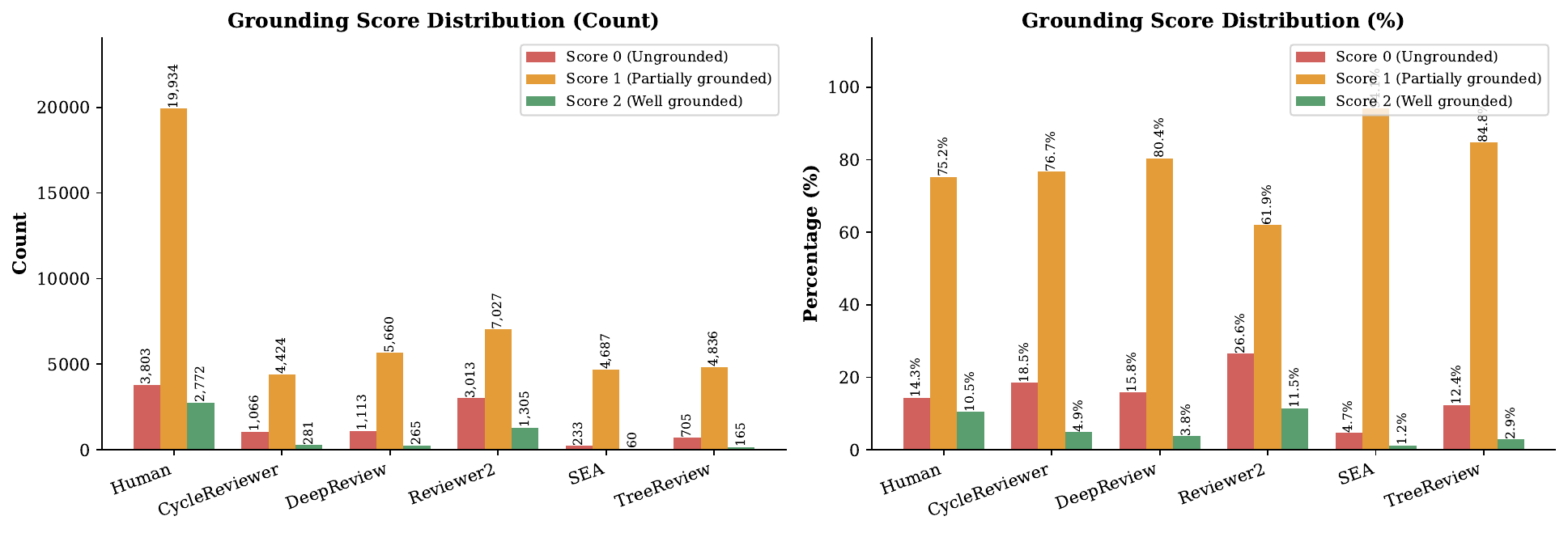}
    \caption{Absolute count (left) and percentage distribution (right) of extracted premises categorized by grounding scores for human reviews and five evaluated LLMs. Score 0 represents ungrounded or vague premises, Score 1 indicates premises anchored to internal manuscript elements, and Score 2 denotes premises anchored to external literature or benchmarks.}
    \label{fig:grounding_dist}
\end{figure}

As illustrated in Table \ref{tab:doa_full} and Figure \ref{fig:grounding_dist}, the disparity in DoA scores stems from a complex interplay between the Premise Ratio (the consistency of providing justification) and the Grounding Score (the evidentiary quality of that justification).

\paragraph{The Illusion of Depth in Reviewer2.}
Reviewer2 presents an intriguing paradox at the grounding level. Despite producing an overwhelming absolute volume of vague, unanchored statements (Score 0), it surprisingly generates more externally grounded premises (Score 2) than other LLM baselines. However, as the per-aspect analysis in Figure~\ref{fig:doa_score_across_aspect} confirms, this marginal grounding advantage is entirely negated by its severely low Premise Ratio, which dilutes the overall analytical depth across all aspects.


\begin{figure}[H]
    \centering
    \includegraphics[width=\linewidth]{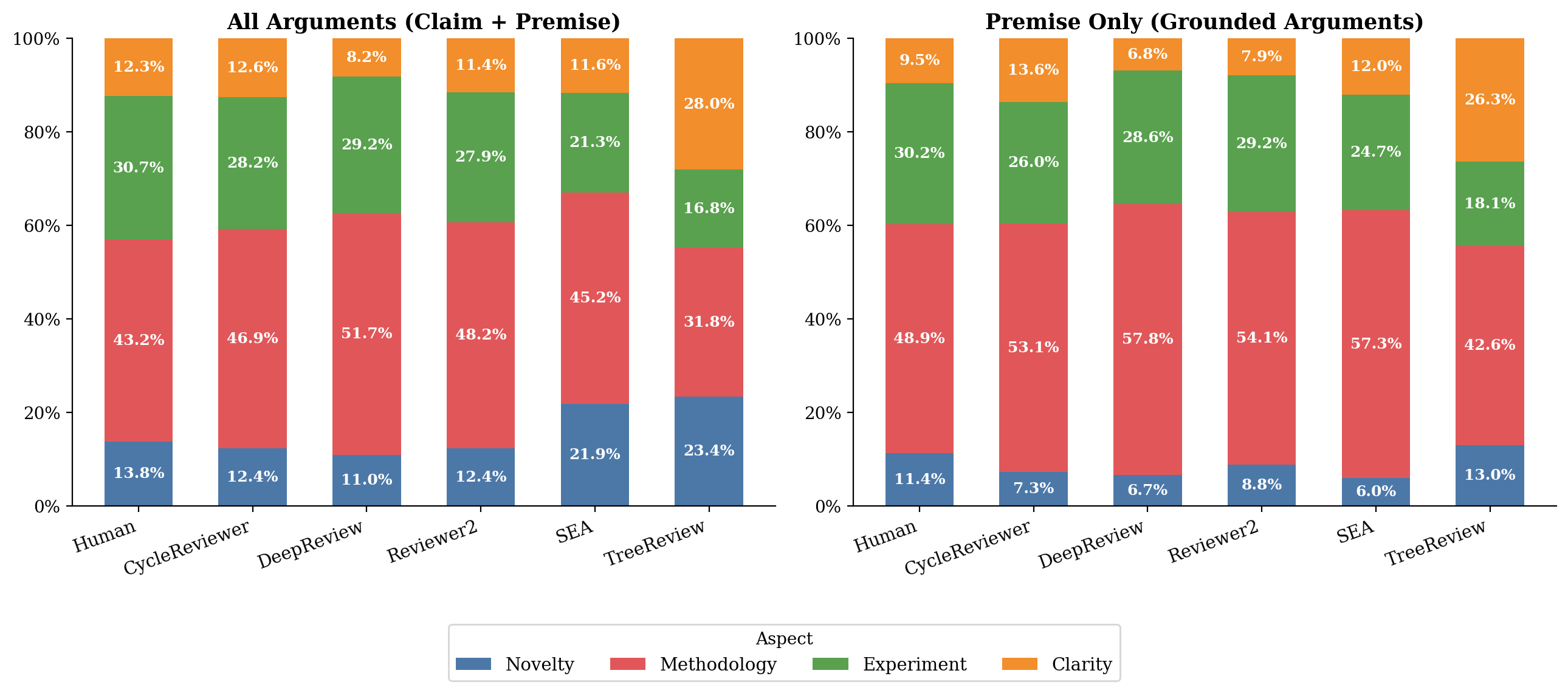}
    \caption{Distribution of review focus across 4 key aspects. The left panel shows the distribution across all extracted arguments, while the right panel highlights the distribution strictly for premises (grounded arguments).}
    \label{fig:aspect_distribution}
\end{figure}

\begin{table}[H]
\centering
\caption{Macro-average aspect distribution (premise-level) and alignment with Human reviewers
         measured by Jensen-Shannon Divergence (JSD). JSD $\in [0,1]$; lower values indicate
         closer alignment to the Human aspect distribution.
         $H$ (bits) denotes Shannon entropy of the aspect distribution (max $= \log_2 4 \approx 2$ bits).}
\label{tab:aspect_jsd}
\setlength{\tabcolsep}{6pt}
\begin{tabular}{lcccc|cc}
\toprule
\textbf{Reviewer} & \textbf{Novelty} & \textbf{Methodology} & \textbf{Experiment} & \textbf{Clarity}
                  & \textbf{JSD $\downarrow$} & \textbf{H (bits) $\uparrow$} \\
\midrule
Human        & 0.114 & 0.489 & 0.302 & 0.095 & ---            & 1.552 \\
\midrule
CycleReviewer  & 0.073 & 0.531 & 0.260 & 0.136 & 0.089          & 1.271 \\
DeepReview   & 0.067 & \textbf{0.578} & 0.286 & \textbf{0.068} & 0.095 & 1.088 \\
Reviewer2    & 0.088 & 0.541 & 0.292 & 0.079 & \textbf{0.055} & 1.305 \\
SEA          & 0.060 & 0.573 & 0.247 & 0.120 & 0.110          & 1.210 \\
TreeReview   & \textbf{0.130} & 0.426 & 0.181 & \textbf{0.263} & 0.205 & 1.497 \\
\bottomrule
\end{tabular}
\end{table}

As illustrated in Figure~\ref{fig:aspect_distribution} and Table~\ref{tab:aspect_jsd},
analyzing the substantive distribution of review focus reveals critical divergences in how analytical effort is allocated across aspects. Fundamentally, both human experts and LLMs dedicate the largest proportion of their critiques to a paper's core technical components: Methodology ($\sim$43\%) and Experimental Design ($\sim$31\%), confirming a broad consensus on the most critical review dimensions. To quantify the degree of alignment between each LLM and Human reviewers, we compute the Jensen-Shannon Distance (JSD, the square root of the Jensen-Shannon Divergence) between each system's macro-averaged premise-level aspect distribution and the Human distribution. JSD $\in [0,1]$, where $0$ indicates identical distributions.

\paragraph{Alignment with Human Priorities.} 

Most advanced baselines successfully mirror human intuitive focus, dedicating the largest proportion of their grounded premises to core technical components: Methodology ($\sim 43-58\%$) and Experimental Design ($\sim 18-29\%$). Notably, \textbf{Reviewer2} achieves the closest alignment to Human reviewers with the lowest JSD of $0.055$. Its premise distribution (Methodology $54.1\%$, Experiment $29.2\%$, Clarity $7.9\%$) closely traces the human pattern ($48.9\%$, $30.2\%$, $9.5\%$), demonstrating a well-calibrated allocation of critical effort. \textbf{DeepReview} achieves the lowest Clarity proportion ($6.8\%$) and the highest Methodology concentration ($57.8\%$). While this results in the lowest Shannon entropy ($H = 1.088$ bits)---indicating a narrower, highly specialized focus---it confirms the model's capacity to firmly anchor its feedback in the most critical dimensions of the submission.


\paragraph{The Surface-Level Trap.} 
Rather than being an inherent LLM limitation, the ``surface-level trap'' manifests when automated frameworks lack explicit, domain-specific evaluation constraints. Remarkably, even highly structured pipelines can fall into this trap. This is starkly pronounced in \textbf{TreeReview}, which, despite its complex reasoning topology, allocates an excessive $26.3\%$ of its premise-level effort to Clarity, nearly $2.8\times$ the proportion of Human reviewers ($9.5\%$). Consequently, TreeReview records the highest JSD against Humans ($0.205$) and the lowest Methodology coverage ($42.6\%$), confirming that without strict dimensional guidance, its analytical distribution naturally diverges toward superficial nitpicking.

\begin{figure}[H]
    \centering
    \includegraphics[width=\linewidth]{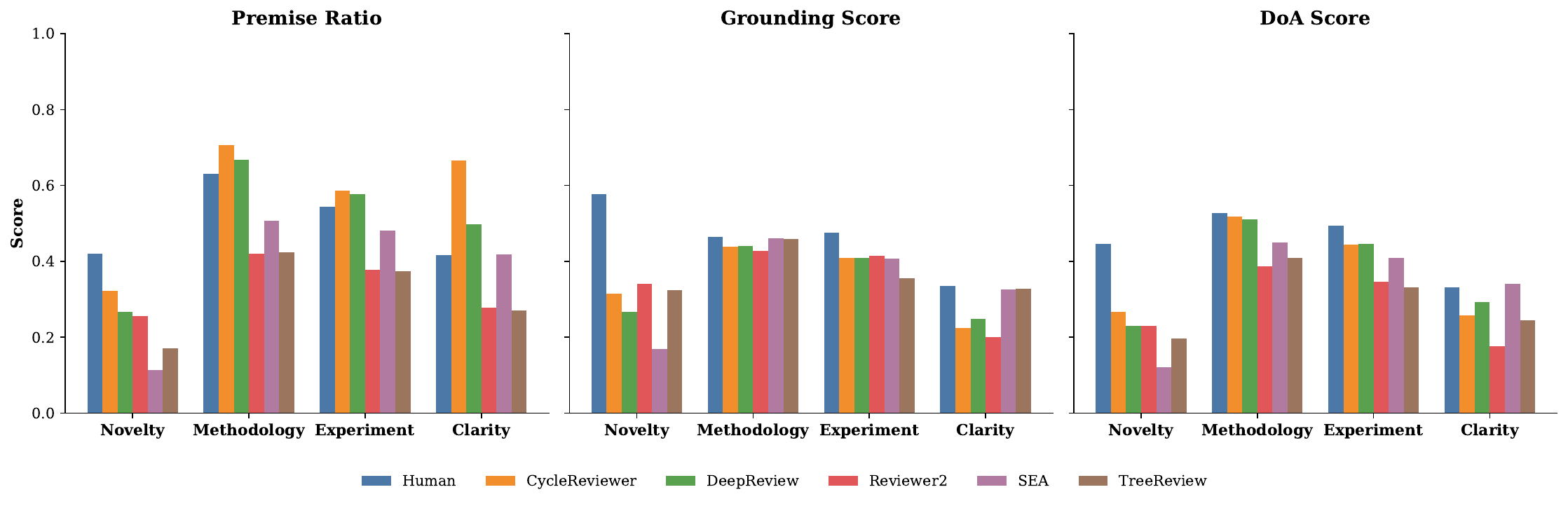}
    \caption{Per-aspect Depth of Analysis scores for Human and five LLM baselines, averaged across the three conferences. Each group corresponds to one of the four review aspects (Novelty \& Related Work, Methodology, Experimental Design, Clarity)}
    \label{fig:doa_score_across_aspect}
\end{figure}

Figure~\ref{fig:doa_score_across_aspect} decomposes the overall DoA score into its four constituent aspects---Novelty \& Related Work, Methodology, Experimental Design, and Clarity---revealing the underlying drivers of the scores reported in Figure~\ref{fig:grounding_dist} and Table~\ref{tab:aspect_jsd}. For each aspect, the harmonic mean of Premise Ratio and Grounding Score is reported per reviewer, allowing a fine-grained comparison of \emph{where} and \emph{how deeply} each system substantiates its arguments.

\paragraph{Shared Prioritization of Core Technical Aspects.}
A consistent pattern emerges across both human and LLM reviewers: DoA scores are substantially higher for Methodology and Experimental Design than for Novelty and Clarity. 
For Human reviewers, the Methodology aspect achieves the highest DoA Score ($0.527 \pm 0.067$), followed closely by Experiment ($0.494 \pm 0.090$), while Novelty ($0.447 \pm 0.207$) and Clarity ($0.332 \pm 0.189$) trail significantly behind.
This pattern holds uniformly across all five baselines, confirming that both humans and LLMs inherently recognize the need to anchor their most substantive arguments in the methodological and experimental core of a paper.
The effect is directly traceable to elevated Premise Ratios and Grounding Scores in these two aspects: for instance, Human reviewers achieve a Premise Ratio of $0.630$ on Methodology versus only $0.417$ on Clarity, indicating that technical claims are far more consistently backed by evidence than presentation-level observations.

\paragraph{Evidence Density as the Key Differentiator Across Baselines.}
The second insight concerns the systematic gap between baselines. Across all four aspects, CycleReviewer and DeepReview consistently achieve DoA scores closest to---and in several cases statistically indistinguishable from---Human reviewers.
This cross-aspect parity is not coincidental: both systems maintain the highest Premise Ratios among LLM baselines across every aspect (e.g., CycleReviewer reaches $0.707$ and DeepReview $0.668$ on Methodology, both exceeding the Human value of $0.630$), demonstrating that their superior overall DoA is driven by a systematic tendency to substantiate claims with grounded evidence regardless of the aspect under discussion.
In contrast, Reviewer2, SEA, and TreeReview show markedly lower Premise Ratios particularly on Novelty ($0.255$, $0.114$, $0.171$ respectively), producing the steepest per-aspect DoA drops and confirming that their aggregate weakness is not confined to any single dimension but reflects a globally deficient evidentiary discipline.

\paragraph{Summary of Analytical Depth.} 
In conclusion, achieving a human-level Depth of Analysis requires more than merely generating a high volume of text. Models that fall into the trap of unsupported verbosity (Reviewer2) or surface-level nitpicking (TreeReview) are severely penalized. To bridge the analytical gap, automated reviewers must systematically substantiate their claims and strictly prioritize core technical dimensions over formatting issues.

\subsection{Novelty Assessment}
\label{sec:detailed_novelty}

\begin{figure}[H]
    \centering
    \includegraphics[width=0.42\linewidth]{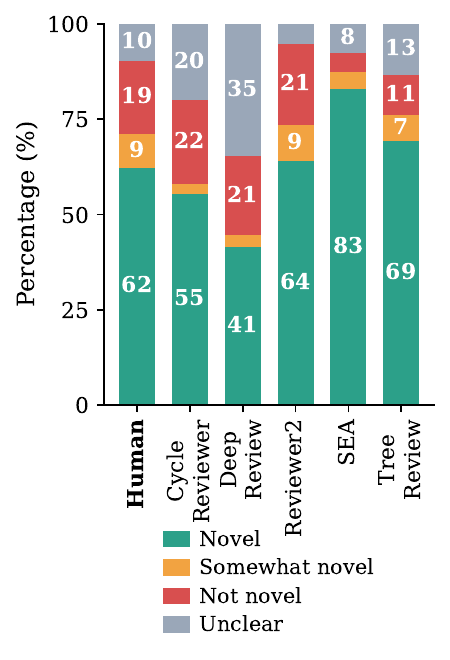}
    \caption{Novelty stance distribution per system. SEA calls the contribution \emph{Novel} in 83\% of claims, against 62\% for human reviews; DeepReview is the most skeptical (41\% \emph{Novel}, 21\% \emph{Not novel}, 35\% \emph{Unclear}).}
    \label{fig:novelty_a}
\end{figure}

\begin{table}[H]
\centering
\caption{Detailed Cross-Venue Performance for \textbf{Novelty Assessment}. Statistical significance is computed independently for each baseline across the 3 conferences to evaluate consistency.}
\label{tab:novelty_full}
\vspace{0.15cm}
\resizebox{\linewidth}{!}{%
\begin{tblr}{
  column{1} = {l},
  column{2-4} = {c},
  hline{1,9} = {-}{0.08em},
  hline{3} = {-}{0.05em},
  hline{4} = {-}{dashed},
}
\SetCell[r=2]{l}\textbf{Baselines} & \SetCell[c=3]{c}\textbf{Novelty Assessment Scores (Mean $\pm$ Std)} \\
 & \textbf{ICLR 2026} & \textbf{ICML 2025} & \textbf{NeurIPS 2025} \\
\textbf{Human} & $0.764 \pm 0.195$ & $0.833 \pm 0.189$ & $0.814 \pm 0.183$ \\
CycleReviewer & $0.793 \pm 0.219$\textsuperscript{ns} & $0.781 \pm 0.210^{*}$ & $0.793 \pm 0.228$\textsuperscript{ns} \\
DeepReview & $0.737 \pm 0.220$\textsuperscript{ns} & $0.785 \pm 0.188^{**}$ & $0.784 \pm 0.203^{*}$ \\
Reviewer2 & $0.792 \pm 0.215$\textsuperscript{ns} & $0.792 \pm 0.216$\textsuperscript{ns} & $0.815 \pm 0.195$\textsuperscript{ns} \\
SEA & $\mathbf{0.831 \pm 0.184}^{***}$ & $\mathbf{0.870 \pm 0.187}^{*}$ & $\mathbf{0.860 \pm 0.202}^{**}$ \\
TreeReview & $0.813 \pm 0.201^{**}$ & $0.808 \pm 0.216$\textsuperscript{ns} & $0.818 \pm 0.197$\textsuperscript{ns} \\
\end{tblr}
}
\end{table}


Table \ref{tab:novelty_full} reports the cross-venue scalar \textit{Novelty Assessment} scores. These values should be interpreted carefully: they do not certify a manuscript's objective novelty, but rather measure whether the novelty claims expressed in a review can be grounded in retrievable prior work under the PRISM pipeline. Across almost all venues, mean scores fall between $0.730$ and $0.870$, indicating that both human reviewers and automated baselines frequently produce novelty claims that the retrieval-and-verification procedure can resolve with substantial evidence. The large number of non-significant differences (\textsuperscript{ns}) for models such as Reviewer2, and TreeReview therefore suggests similar \emph{evidence-grounding performance on this scalar metric}, not full claim-by-claim agreement with human reviewers.

\paragraph{The Outperformance of SEA.} 
The most notable result on this scalar metric is the performance of the \textbf{SEA} baseline. While SEA is not uniformly strongest on the other review dimensions, it achieves the highest novelty-assessment score in all three venues and significantly exceeds human reviews in ICLR 2026 ($p_{Holm} < 0.001$), ICML 2025 ($p_{Holm} < 0.05$), and NeurIPS 2025 ($p_{Holm} < 0.01$). Within the interpretation above, this suggests that SEA's structured generation style tends to produce novelty claims that are especially easy for the PRISM retrieval-and-verification pipeline to ground in prior work. As the agreement analysis below will show, this should not be conflated with universally stronger alignment to human novelty judgments.




\begin{figure}[H]
    \centering
    \includegraphics[width=\linewidth]{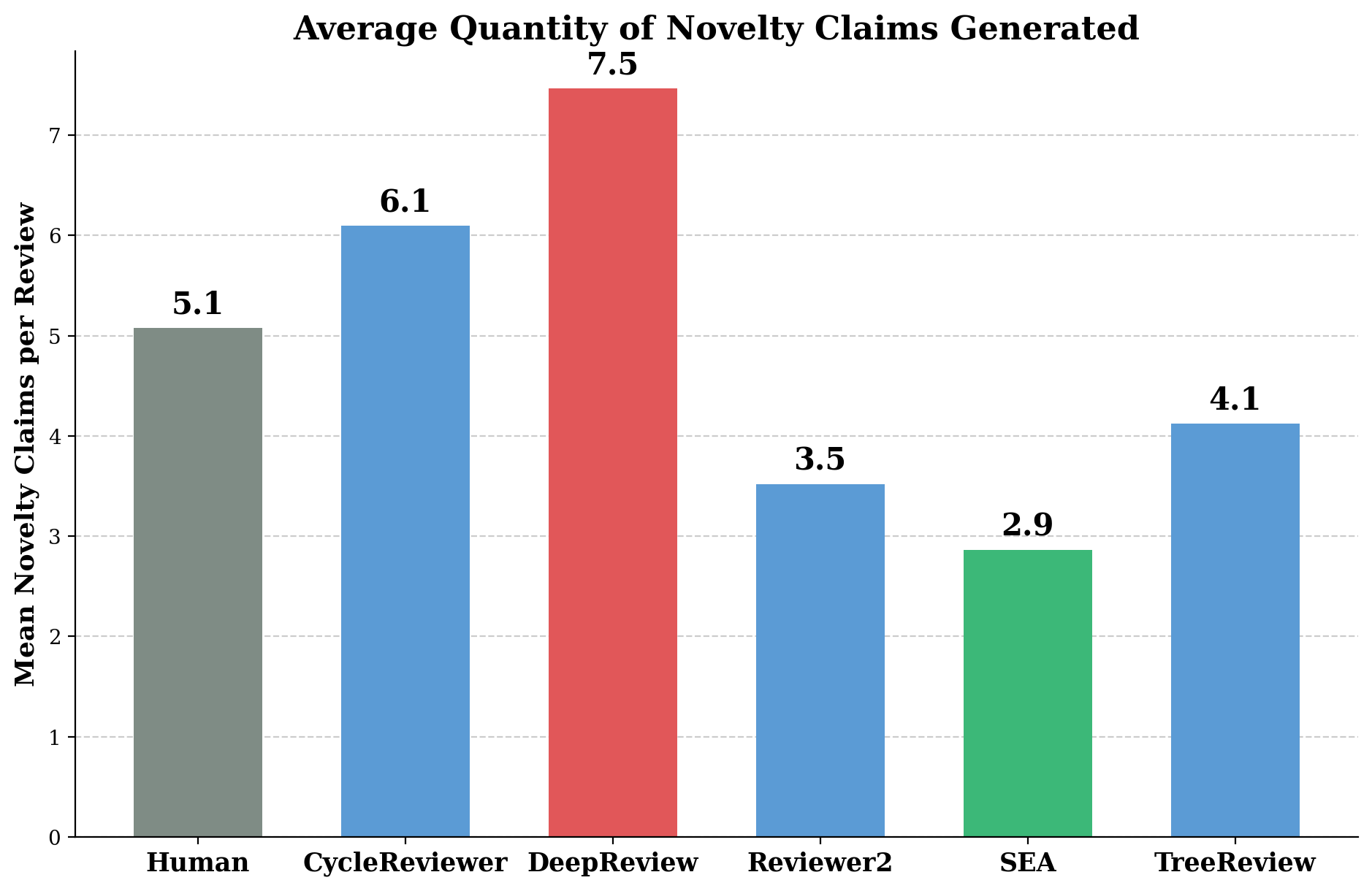}
    \caption{Average number of novelty claims generated per review across the human and LLM Reviewers.}
    \label{fig:mean_claims_per_review}
\end{figure}

\paragraph{Claim Volume and Generation Verbosity.} Figure~\ref{fig:mean_claims_per_review} reveals another important difference that scalar novelty scores alone do not capture: review sources vary substantially in how many novelty claims they choose to make. Because the human input concatenates multiple independent reviews for the same paper, raw human claim counts must be normalized by the number of individual reviews before they are interpreted as per-review verbosity. In the unnormalized benchmark output, the concatenated human-review bundle averages $5.1$ extracted novelty-related claims per paper; the benchmark code therefore reports the human claim-volume statistic as claims per individual human review whenever review-count metadata is available. \textbf{DeepReview} is markedly more verbose at $7.5$ claims per generated review, reflecting a finer-grained decomposition of contributions and comparisons. By contrast, \textbf{SEA} ($2.9$ claims) and \textbf{Reviewer2} ($3.5$ claims) are much more conservative, concentrating their novelty discussion into fewer statements. This matters because downstream agreement depends not only on how claims are scored once extracted, but also on claim granularity and boundary choices at the extraction stage.

\paragraph{Summary.} Taken together, these results show that LLM reviewers can produce novelty claims that are often well grounded under the PRISM retrieval-and-verification pipeline. However, this scalar score should not be read as evidence that LLMs can certify the objective novelty of a manuscript, nor that they fully replicate human novelty judgments. The next subsection addresses the harder question: once all review sources are normalized through the same pipeline, do they actually arrive at the same claim-level novelty conclusions?

\subsection{Flaw Identification \& Prioritization of Major Issue}
\label{sec:detailed_flaw_analysis}

\begin{table}[H]
\centering
\caption{Detailed Cross-Venue Performance for \textbf{Flaw Identification \& Prioritization}. Statistical significance is computed independently for each baseline across the 3 conferences to evaluate consistency.}
\label{tab:flaw_full}
\vspace{0.15cm}
\resizebox{\linewidth}{!}{%
\begin{tblr}{
  column{1} = {l},
  column{2-4} = {c},
  hline{1,24} = {-}{0.08em},
  hline{3} = {-}{0.05em},
  hline{10,17} = {-}{dashed},
}
\SetCell[r=2]{l}\textbf{Metric / Baselines} & \SetCell[c=3]{c}\textbf{Scores (Mean $\pm$ Std)} \\
 & \textbf{ICLR 2026} & \textbf{ICML 2025} & \textbf{NeurIPS 2025} \\
\SetCell[c=4]{l}\textit{Critical Flaw Identification} \\
\quad \textbf{Human} & $0.327 \pm 0.149$ & $0.299 \pm 0.167$ & $0.306 \pm 0.159$ \\
\quad CycleReviewer & $0.186 \pm 0.263^{***}$ & $0.254 \pm 0.319^{***}$ & $0.232 \pm 0.302^{***}$ \\
\quad DeepReview & $0.309 \pm 0.291\textsuperscript{ns}$ & $0.360 \pm 0.351\textsuperscript{ns}$ & $0.383 \pm 0.335\textsuperscript{ns}$ \\
\quad Reviewer2 & $\mathbf{0.593 \pm 0.317}^{***}$ & $\mathbf{0.683 \pm 0.304}^{***}$ & $\mathbf{0.657 \pm 0.290}^{***}$ \\
\quad SEA & $0.174 \pm 0.237^{***}$ & $0.219 \pm 0.259^{***}$ & $0.201 \pm 0.262^{***}$ \\
\quad TreeReview & $0.216 \pm 0.281^{***}$ & $0.254 \pm 0.333^{**}$ & $0.265 \pm 0.298^{**}$ \\
\SetCell[c=4]{l}\textit{Minor Flaw Identification} \\
\quad \textbf{Human} & $0.250 \pm 0.080$ & $0.262 \pm 0.085$ & $0.248 \pm 0.073$ \\
\quad CycleReviewer & $0.186 \pm 0.151^{***}$ & $0.195 \pm 0.151^{***}$ & $0.154 \pm 0.120^{***}$ \\
\quad DeepReview & $0.256 \pm 0.147\textsuperscript{ns}$ & $0.233 \pm 0.143^{***}$ & $0.203 \pm 0.135^{***}$ \\
\quad Reviewer2 & $\mathbf{0.467 \pm 0.185}^{***}$ & $\mathbf{0.480 \pm 0.182}^{***}$ & $\mathbf{0.463 \pm 0.179}^{***}$ \\
\quad SEA & $0.224 \pm 0.124^{***}$ & $0.207 \pm 0.114^{***}$ & $0.193 \pm 0.116^{***}$ \\
\quad TreeReview & $0.339 \pm 0.147^{***}$ & $0.303 \pm 0.135^{***}$ & $0.306 \pm 0.143^{***}$ \\
\SetCell[c=4]{l}\textit{Prioritization of Major Issue (nCPS)} \\
\quad \textbf{Human} & $0.969 \pm 0.047$ & $0.984 \pm 0.047$ & $0.973 \pm 0.034$ \\
\quad CycleReviewer & $0.969 \pm 0.110^{*}$ & $0.970 \pm 0.108\textsuperscript{ns}$ & $0.968 \pm 0.127^{*}$ \\
\quad DeepReview & $0.968 \pm 0.053\textsuperscript{ns}$ & $0.978 \pm 0.048\textsuperscript{ns}$ & $0.956 \pm 0.110\textsuperscript{ns}$ \\
\quad Reviewer2 & $0.970 \pm 0.078^{*}$ & $0.975 \pm 0.033^{***}$ & $0.970 \pm 0.033\textsuperscript{ns}$ \\
\quad SEA & $\mathbf{0.986 \pm 0.037}^{***}$ & $\mathbf{0.984 \pm 0.038}\textsuperscript{ns}$ & $\mathbf{0.981 \pm 0.080}^{**}$ \\
\quad TreeReview & $0.977 \pm 0.044\textsuperscript{ns}$ & $0.980 \pm 0.042\textsuperscript{ns}$ & $0.975 \pm 0.046\textsuperscript{ns}$ \\
\end{tblr}
}
\end{table}

The detailed results in Table \ref{tab:flaw_full} unequivocally confirm the exhaustive diagnostic capability of Reviewer2. Across every single evaluated venue, Reviewer2 consistently achieves the highest Recall for both \textit{Critical} (ranging from $0.593$ to $0.683$) and \textit{Minor} flaws. More importantly, the statistical tests ($p_{Holm} < 0.001$, denoted as $^{***}$) validate that this over-performance relative to human reviews is structurally ingrained in the model's generation style, not merely a statistical artifact of a specific dataset. This elevated Recall is directly correlated with its overall generation volume: as previously illustrated in the valid flaw counts (Figure \ref{fig:comparison_flaws} - main text), Reviewer2 acts as a high-volume "flaw scanner," extracting an unprecedented absolute number of valid issues per review. By casting a wider diagnostic net, it naturally captures a higher proportion of both fatal methodologies and minor anomalies compared to the more conservative human baseline.
Conversely, other LLM baselines generally exhibit a diagnostic deficit compared to human experts. Models like CycleReviewer and SEA consistently underperform human reviews in extracting both major and minor flaws across all three venues. An interesting anomaly, however, is observed in \textbf{TreeReview}. While it struggles to detect fatal methodological errors (consistently scoring lower than humans in Critical Recall), it consistently outperforms humans in \textit{Minor Flaw Identification} (ranging from $0.303$ at ICML 2025 to $0.339$ at ICLR 2026, all with strong statistical significance). This further corroborates our earlier finding that TreeReview suffers from a "surface-level trap," heavily over-indexing its analytical effort on presentation and formatting anomalies rather than scientific rigor. DeepReview, maintaining a highly conservative profile, yields scores that are statistically indistinguishable (\textsuperscript{ns}) from human experts in every venue on Critical Recall, demonstrating a strong alignment with human reviewing patterns.

Concluding the table analysis, the granular breakdown of the Critique Prioritization Score ($nCPS$) reinforces a macro-level observation: across all three conferences, every reviewer achieves near-ceiling scores ($nCPS > 0.95$). Regardless of their diagnostic capability, the automated systems' structural organization adheres to the academic norm of surfacing major flaws first, and they are statistically indistinguishable from human reviews in most venues. Because nCPS is near ceiling and its rankings are judge- and prompt-sensitive (App.~\ref{app:judges}, \ref{app:prompt}), we draw no ranking conclusion from it.

\begin{figure}[H]
    \centering
    \includegraphics[width=\linewidth]{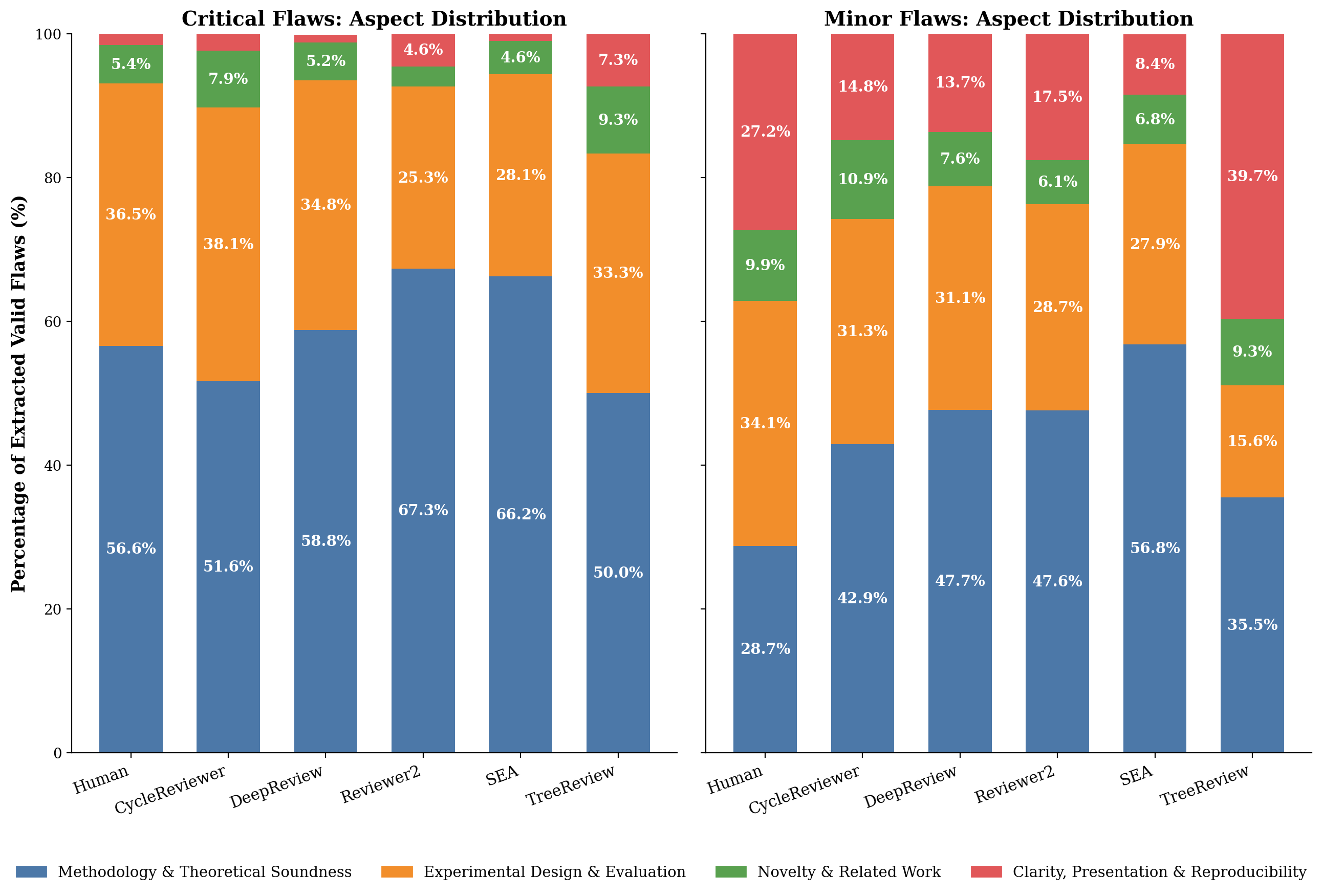}
    \caption{Comparative Aspect Topic Distribution of extracted valid flaws, stratified by severity (\textit{Critical} vs. \textit{Minor}).}
    \label{fig:severity_aspect}
\end{figure}

To gain a deeper understanding of reviewer behavior, we stratify the aspect topic distribution by flaw severity (Critical vs. Minor), as visualized in Figure \ref{fig:severity_aspect}. This decomposition reveals profound insights into the contextual adaptability—or lack thereof—within LLM baselines compared to human experts.

\paragraph{Alignment on Critical Vulnerabilities.} 
When evaluating \textit{Critical} flaws, human experts demonstrate a laser-focused approach, allocating an overwhelming $93.1\%$ of their critiques to core technical components ($56.6\%$ on Methodology and $36.5\%$ on Experimental Design). Formatting issues (Clarity) account for merely $1.5\%$ of critical flaws. \textbf{DeepReview} exhibits a remarkably identical signature, dedicating $93.5\%$ of its critical flaw detection to Methodology and Experiments, with only $1.1\%$ wasted on Clarity. This confirms that at the highest severity level, well-calibrated models like DeepReview successfully emulate the human capacity to isolate fatal scientific errors. Conversely, \textbf{TreeReview} broadens the scope of critical evaluation. Rather than limiting itself strictly to methodology, it proactively identifies critical presentation errors at a rate of $7.3\%$, ensuring that even formal aesthetics are held to the highest standards of severity.

\paragraph{The Adaptive Focus on Minor Flaws.}
The distribution of \textit{Minor} flaws reveals a compelling alignment between human experts and LLM baselines. Unlike critical flaws, which are overwhelmingly technical, the cognitive focus for minor flaws naturally shifts toward surface-level anomalies. Both human reviewers and LLMs significantly reduce their extreme scrutiny on Methodology, while substantially increasing their attention to \textit{Clarity, Presentation \& Reproducibility}. For instance, the human focus on Clarity surges to $27.2\%$ for minor flaws (up from a mere $1.5\%$ in critical flaws). Remarkably, the automated baselines successfully mirror this contextual shift. \textbf{TreeReview} captures a massive $39.7\%$ of clarity-related minor issues, while even verbose models like \textbf{Reviewer2} nearly quadruple their attention to presentation ($17.5\%$, up from $4.6\%$) compared to their critical flaw distribution.

\paragraph{Emulating Human Flaw Categorization.} 
This dynamic shift demonstrates that LLMs are not rigidly applying a single evaluation template. Instead, they contextually adapt the nature of their critiques based on severity. While they still capture minor methodological nitpicks (as seen in DeepReview and SEA), they correctly associate a large portion of minor flaws with presentation errors, directly aligning with how human experts intuitively classify minor review points.

\paragraph{Summary.}
These findings highlight a sophisticated capability in modern automated peer review: LLMs can successfully mimic human intuition in identifying and categorizing different types of flaws. By instinctively mapping core technical breakdowns to \textit{Critical} severity and correctly shifting their lens toward presentation issues for \textit{Minor} anomalies, LLMs prove they possess a nuanced, human-like understanding of manuscript evaluation.

\subsection{Multi-Dimensional Constructiveness}
\label{sec:detailed_constructiveness}

\begin{figure}[H]
    \centering
    \includegraphics[width=0.55\linewidth]{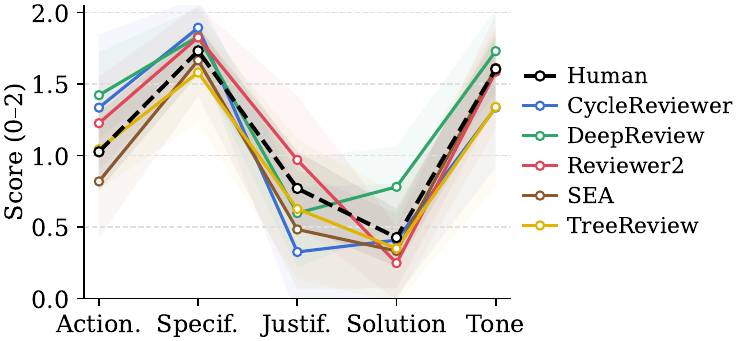}
    \caption{Constructiveness sub-dimensions (D1--D5, per-ARC averages on a 0--2 scale) under the primary judge.}
    \label{fig:constructiveness_detailed}
\end{figure}

\begin{table}[H]
\centering
\caption{Detailed Cross-Venue Performance for \textbf{Constructiveness Score (MCS)}. Statistical significance is computed independently for each baseline across the 3 conferences to evaluate consistency.}
\label{tab:mcs_full}
\vspace{0.15cm}
\resizebox{\linewidth}{!}{%
\begin{tblr}{
  column{1} = {l},
  column{2-4} = {c},
  hline{1,9} = {-}{0.08em},
  hline{3} = {-}{0.05em},
  hline{4} = {-}{dashed},
}
\SetCell[r=2]{l}\textbf{Baselines} & \SetCell[c=3]{c}\textbf{Constructiveness Scores (Mean $\pm$ Std)} \\
 & \textbf{ICLR 2026} & \textbf{ICML 2025} & \textbf{NeurIPS 2025} \\
\textbf{Human} & $0.601 \pm 0.063$ & $0.521 \pm 0.070$ & $0.546 \pm 0.057$ \\
CycleReviewer & $0.540 \pm 0.120^{***}$ & $0.523 \pm 0.095\textsuperscript{ns}$ & $0.527 \pm 0.119^{*}$ \\
DeepReview & $\mathbf{0.636 \pm 0.091}^{***}$ & $\mathbf{0.638 \pm 0.085}^{***}$ & $\mathbf{0.635 \pm 0.087}^{***}$ \\
Reviewer2 & $0.587 \pm 0.100\textsuperscript{ns}$ & $0.584 \pm 0.101^{***}$ & $0.586 \pm 0.087^{***}$ \\
SEA & $0.502 \pm 0.098^{***}$ & $0.487 \pm 0.084^{***}$ & $0.485 \pm 0.070^{***}$ \\
TreeReview & $0.490 \pm 0.144^{***}$ & $0.498 \pm 0.125^{**}$ & $0.496 \pm 0.120^{***}$ \\
\end{tblr}
}
\end{table}

Table \ref{tab:mcs_full} presents the granular breakdown of the MCS and its sub-dimensions across the three evaluated conference venues. A longitudinal analysis reveals that while the constructiveness of both human and automated reviewers fluctuates slightly across different venues, DeepReview consistently maintains a substantial lead. Specifically, DeepReview achieves the highest MCS across all three conferences (ranging from $0.635$ at NeurIPS 2025 to $0.638$ at ICML 2025). Importantly, Holm's post-hoc tests confirm that DeepReview's superiority over human reviews ($0.601$ at ICLR 2026 to $0.546$ at NeurIPS 2025) is statistically significant ($p_{Holm} < 0.05$) in every venue evaluated. Furthermore, Reviewer2 demonstrates competitive performance, particularly at ICML 2025 and NeurIPS 2025, where its MCS scores ($0.584$, and $0.586$) slightly match human reviews. In stark contrast, models such as SEA and TreeReview consistently underperform compared to human reviews across all three venues, highlighting a persistent limitation in their capacity to formulate genuinely constructive critiques.



\begin{table}[H]
\centering
\caption{Detailed Constructiveness Sub-dimensions (D1-D5) across the 3 Conferences evaluated on a raw scale of $[0, 2]$.}
\label{tab:constructiveness_detailed}
\vspace{0.15cm}
\resizebox{\linewidth}{!}{%
\begin{tblr}{
  column{1} = {l},
  column{2-6} = {c},
  hline{1,8} = {-}{0.08em},
  hline{2} = {-}{0.05em},
  hline{3} = {-}{dashed},
}
\textbf{System} & {\textbf{D1:} \\ \textbf{Actionability}} & {\textbf{D2:} \\ \textbf{Specificity}} & {\textbf{D3:} \\ \textbf{Justification}} & {\textbf{D4:} \\ \textbf{Solution}} & {\textbf{D5:} \\ \textbf{Tone}} \\
\textbf{Human}      & $1.026 \pm 0.270$ & $1.732 \pm 0.163$ & $0.770 \pm 0.259$ & $0.427 \pm 0.202$ & $1.605 \pm 0.230$ \\
CycleReviewer         & $1.335 \pm 0.511$ & $\mathbf{1.894} \pm 0.211$ & $0.326 \pm 0.452$ & $0.410 \pm 0.395$ & $1.335 \pm 0.413$ \\
DeepReview          & $\mathbf{1.422} \pm 0.299$ & $1.832 \pm 0.202$ & $0.598 \pm 0.381$ & $\mathbf{0.781} \pm 0.284$ & $\mathbf{1.730} \pm 0.290$ \\
Reviewer2           & $1.226 \pm 0.324$ & $1.826 \pm 0.222$ & $\mathbf{0.969} \pm 0.447$ & $0.249 \pm 0.259$ & $1.586 \pm 0.401$ \\
SEA                 & $0.819 \pm 0.395$ & $1.666 \pm 0.248$ & $0.484 \pm 0.413$ & $0.333 \pm 0.259$ & $1.613 \pm 0.266$ \\
TreeReview          & $1.045 \pm 0.283$ & $1.581 \pm 0.354$ & $0.627 \pm 0.487$ & $0.351 \pm 0.382$ & $1.340 \pm 0.556$ \\
\end{tblr}
}
\end{table}

\begin{figure}[H]
    \centering
    \includegraphics[width=\linewidth]{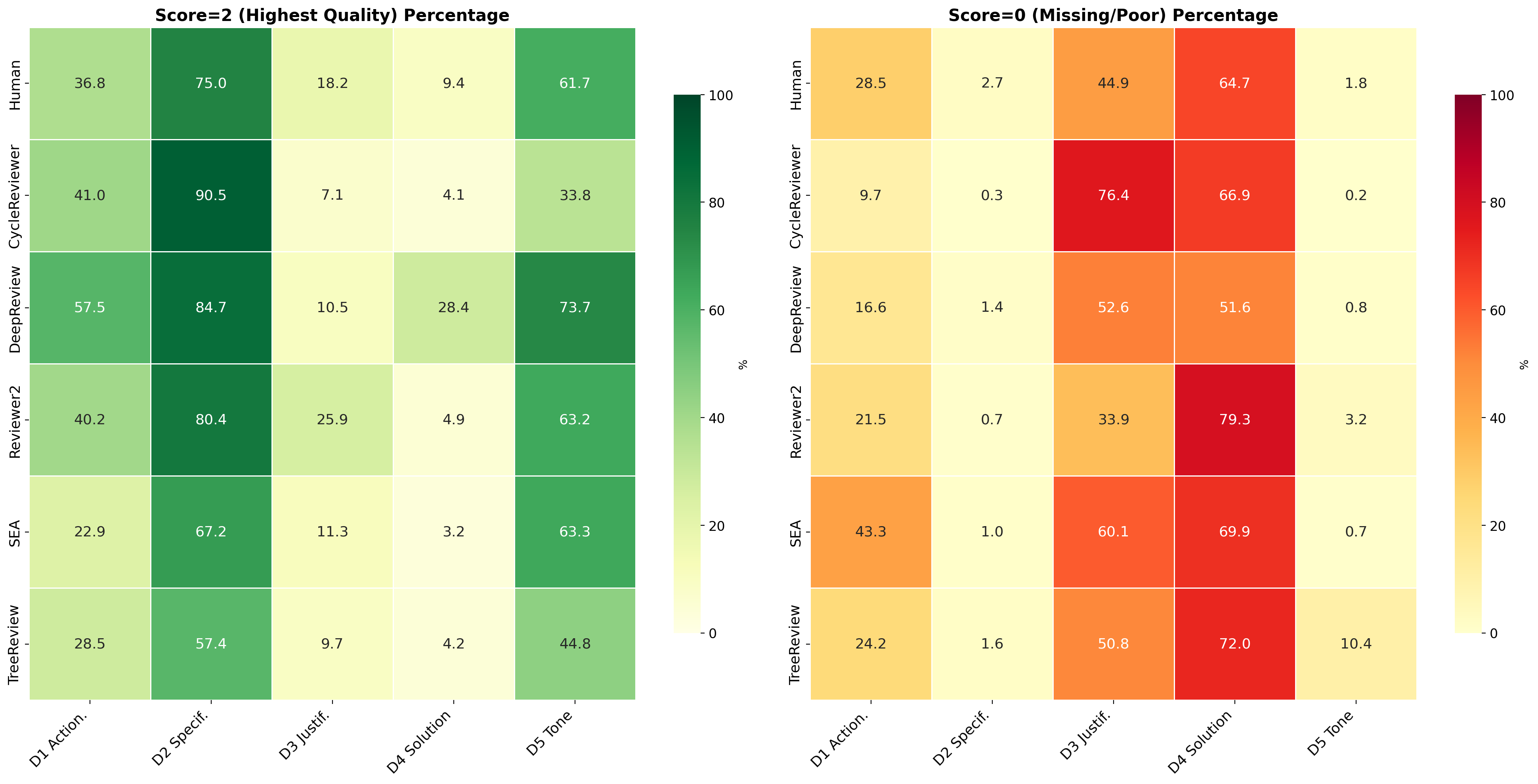}
    \caption{Heatmap detailing the performance across the five core sub-metrics of constructiveness (Actionability, Specificity, Justification, Solution, and Tone).}
    \label{fig:mcs_radar}
\end{figure}

\begin{figure}[H]
    \centering
    \includegraphics[width=\linewidth]{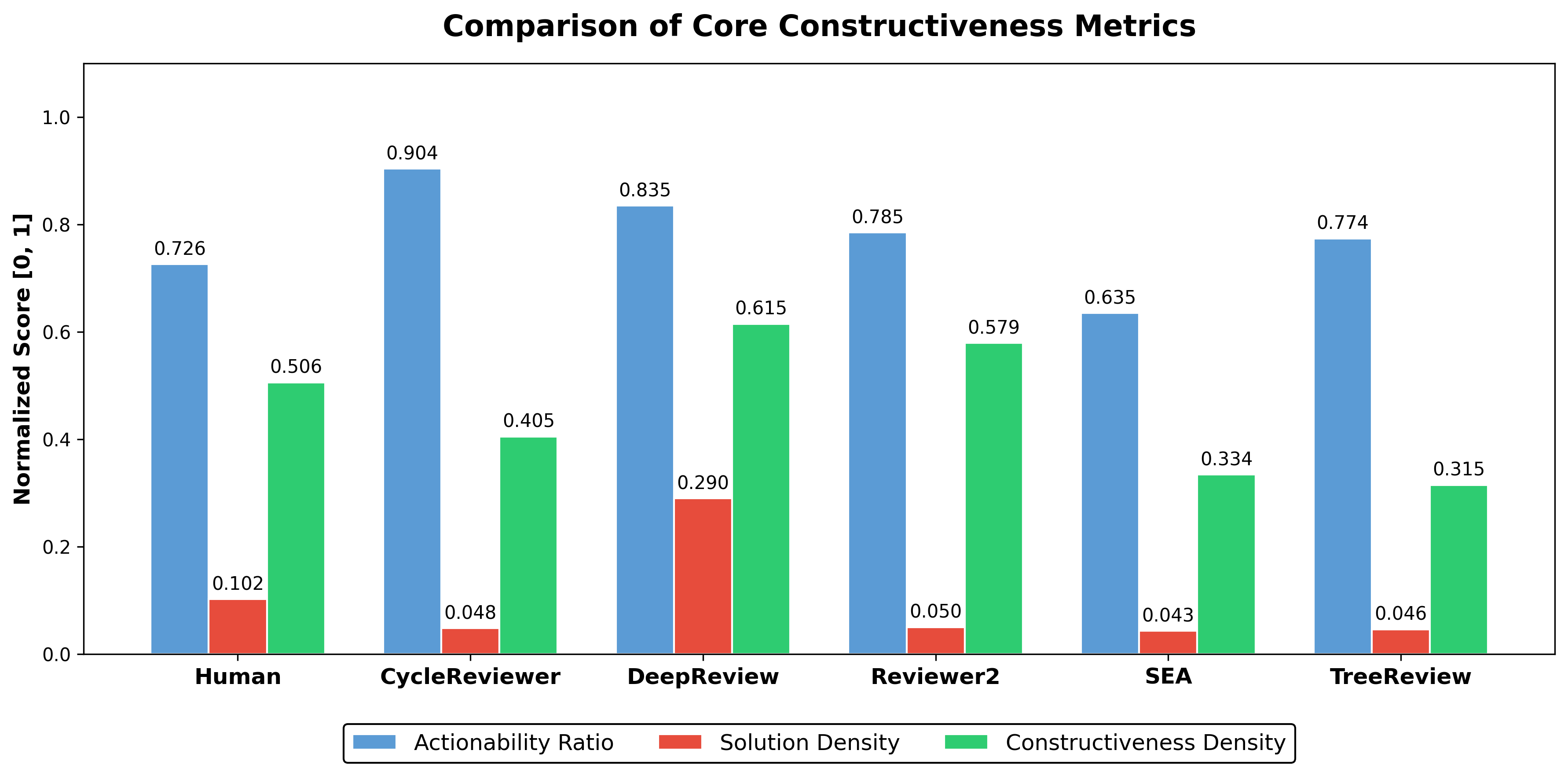}
    \caption{Comparison of core constructiveness metrics (Actionability Ratio, Solution Density, and Constructiveness Density).}
    \label{fig:core_constructiveness}
\end{figure}

To provide a more nuanced evaluation, we derive three auxiliary density metrics from the five core dimensions ($D_1-D_5$). Let $\mathbb{I}(\cdot)$ be an indicator function. We define: (i) the \textbf{Actionability Ratio} ($AR$) as the proportion of comments providing at least a general direction ($D_1 \ge 1$); (ii) the \textbf{Solution Density} ($SD$) as the percentage of comments offering explicit, implementable fixes ($D_4 = 2$); and (iii) the \textbf{Constructiveness Density} ($CD$) as the proportion of comments achieving high-quality structural alignment ($CLC \ge 0.5$). Formally:

\begin{align}
    AR(R) &= \frac{1}{n} \sum_{j=1}^n \mathbb{I}(D_1(c_j) \ge 1), \\
    SD(R) &= \frac{1}{n} \sum_{j=1}^n \mathbb{I}(D_4(c_j) = 2), \\
    CD(R) &= \frac{1}{n} \sum_{j=1}^n \mathbb{I}(CLC(c_j) \ge 0.5)
\end{align}

As presented in the macro-averaged results (Figure~\ref{fig:mcs_radar}) and the venue-specific breakdown (Figure~\ref{fig:core_constructiveness}), a granular examination of these metrics reveals distinct behavioral signatures distinguishing human reviewers from LLM baselines.

\paragraph{The Solution Bottleneck.} 
The most profound divergence between human experts and top LLMs lies in the capacity to propose explicit improvements. The data reveals a systemic limitation in traditional peer review: while humans are highly proficient at pinpointing concrete flaws ($D_2$ Specificity $\approx 1.72$), they frequently fail to provide actionable remedies. This is evidenced by human reviews' remarkably low Solution Density ($SD \approx 0.102$), indicating that only $10\%$ of human comments contain an explicit fix ($D_4 = 2$). In stark contrast, DeepReview fundamentally alters this paradigm. It consistently achieves an $SD$ approaching $0.29$ and $D_4$ scores near $0.80$ across all venues. DeepReview excels because it does not merely diagnose problems; it proactively prescribes explicit, implementable solutions, thereby maximizing its utility to the authors.

\paragraph{The Verbosity Paradox of Reviewer2.} 
The sub-dimensional breakdown elegantly explains Reviewer2's deceptively high overall scores. Reviewer2 achieves exceptional Justification scores ($D_3 > 0.90$) and a strong Constructiveness Density ($CD \approx 0.58$), largely as an artifact of its highly verbose, two-stage rubric-driven generation style. It excels at providing extensive reasoning for its claims. However, its abysmal Solution Density ($SD \approx 0.05$) and low Solution score ($D_4 \approx 0.26$) reveal a critical flaw: it exhaustively explains \textit{why} something is wrong but almost never tells the author \textit{how} to fix it, rendering its voluminous critiques practically inert.

\paragraph{Actionability without Depth in CycleReviewer.}
Conversely, CycleReviewer exhibits a contrasting failure mode. It achieves the highest Specificity ($D_2 \approx 1.89$) and Actionability Ratio ($AR > 0.90$), meaning almost all of its comments reference concrete paper elements and offer at least a general direction ($D_1 \ge 1$). Yet, it suffers a catastrophic drop in Justification ($D_3 \approx 0.32$) and Solution Density ($SD \approx 0.05$). This signature points to a shallow, "checklist-style" reviewing behavior: it successfully targets specific sections with general demands (e.g., "needs more baselines"), but fails to provide the evidentiary backing or the explicit fixes required of a profound scientific critique.

\paragraph{Tone and Professionalism.}
Finally, the Tone dimension ($D_5$) confirms that well-calibrated LLMs can systematically elevate the discourse of peer review. \textbf{DeepReview} ($D_5 \approx 1.73$) consistently outputs more professional, neutral, and encouraging feedback than human reviews under the primary judge ($D_5 \approx 1.61$), effectively mitigating the dismissive or hostile language occasionally encountered in human peer reviews.

\paragraph{Summary of Constructiveness.} 
This multi-dimensional analysis highlights a fundamental difference in reviewing paradigms. Human experts primarily act as \textit{diagnosticians}, highly effective at pinpointing errors but lacking in actionable guidance. LLM baselines often mask this deficiency with extreme verbosity (Reviewer2) or superficial demands (CycleReviewer). In contrast, \textbf{DeepReview} transcends these limitations, acting more as a \textit{collaborator} by bridging the critical gap between identifying flaws ($D_2$) and formulating explicit, professionally toned solutions ($D_4, D_5$), it represents a significant step toward genuinely constructive automated peer review.

\subsection{Review Sensitivity to Paper Quality: Accept vs.\ Reject Analysis}

\begin{figure}[H]
    \centering
    \includegraphics[width=\linewidth]{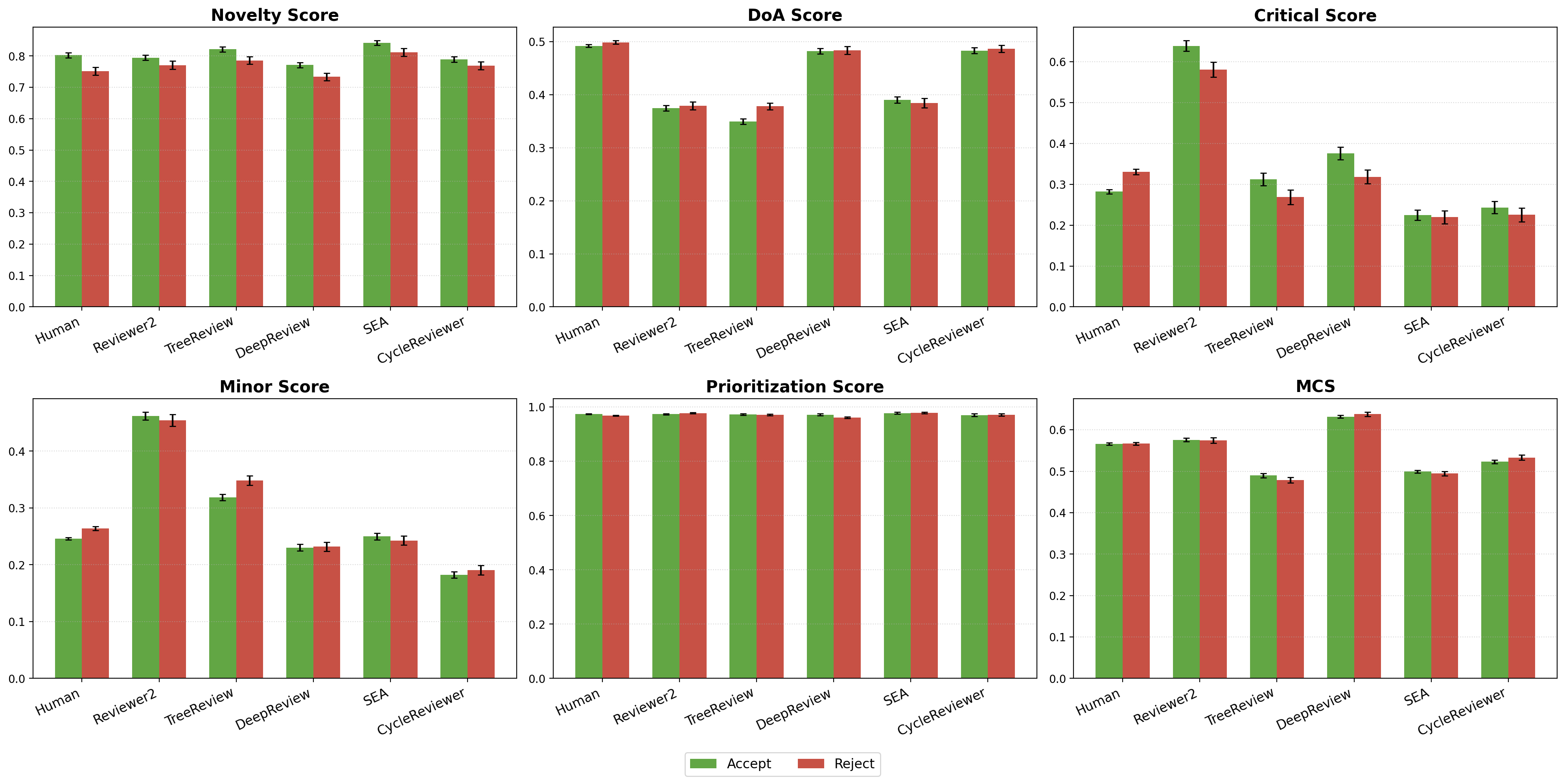}
    \caption{Mean scores of Human and LLM reviewers on six evaluation dimensions, stratified by paper decision (Accept vs.\ Reject)}
    \label{fig:core_accept_reject}
\end{figure}

\paragraph{Statistical Methodology.}
To evaluate whether the generated reviews effectively distinguish between high-quality and low-quality submissions, we compare the metric scores assigned to accepted versus rejected papers (Figure~\ref{fig:core_accept_reject}). Specifically, we apply the Mann-Whitney U test (two-sided, unpaired) to compare per-paper metric scores between accepted and rejected papers for each reviewer baseline. To control the family-wise error rate within each reviewer, $p$-values are adjusted using the Holm-Bonferroni procedure across all metrics tested for that reviewer. The mean score differences ($\Delta = \text{Accept} - \text{Reject}$) and their significance levels are detailed in Table \ref{tab:metric_diffs}.

\begin{table}[H]
\centering
\caption{Metric differences across reviewers. Bold values indicate statistically significant results; (ns) denotes non-significant.}
\label{tab:metric_diffs}
\vspace{0.15cm}
\resizebox{\linewidth}{!}{%
\begin{tblr}{
  column{1} = {l},
  column{2-7} = {c},
  hline{1,8} = {-}{0.08em},
  hline{2} = {-}{0.05em},
  colsep = 6pt,
}
\textbf{Metric} & \textbf{Human} & \textbf{Reviewer2} & \textbf{TreeReview} & \textbf{DeepReview} & \textbf{SEA} & \textbf{CycleReviewer} \\
Novelty Score        & $\mathbf{+0.051^{*}}$        & $+0.024\textsuperscript{ns}$ & $+0.035\textsuperscript{ns}$ & $+0.037\textsuperscript{ns}$ & $+0.029\textsuperscript{ns}$ & $+0.021\textsuperscript{ns}$ \\
DoA Score            & $-0.006\textsuperscript{ns}$ & $-0.004\textsuperscript{ns}$ & $-0.028\textsuperscript{ns}$ & $-0.001\textsuperscript{ns}$ & $-0.006\textsuperscript{ns}$ & $-0.003\textsuperscript{ns}$ \\
Critical Score       & $\mathbf{-0.049^{***}}$      & $-0.058\textsuperscript{ns}$ & $+0.043\textsuperscript{ns}$ & $+0.005\textsuperscript{ns}$ & $-0.005\textsuperscript{ns}$ & $+0.018\textsuperscript{ns}$ \\
Minor Score          & $\mathbf{-0.018^{**}}$       & $-0.008\textsuperscript{ns}$ & $-0.030\textsuperscript{ns}$ & $-0.002\textsuperscript{ns}$ & $-0.007\textsuperscript{ns}$ & $-0.004\textsuperscript{ns}$ \\
Prioritization Score & $\mathbf{+0.006^{***}}$      & $-0.004\textsuperscript{ns}$ & $+0.001\textsuperscript{ns}$ & $\mathbf{+0.011^{***}}$       & $-0.001\textsuperscript{ns}$ & $-0.001\textsuperscript{ns}$ \\
MCS                  & $-0.001\textsuperscript{ns}$ & $+0.001\textsuperscript{ns}$ & $+0.011\textsuperscript{ns}$ & $-0.006\textsuperscript{ns}$ & $+0.004\textsuperscript{ns}$ & $-0.010\textsuperscript{ns}$ \\
\end{tblr}%
}
\end{table}

\paragraph{Human Reviews Exhibit Strong Predictive Validity.}
Although human reviewers evaluate manuscripts blindly without any knowledge of the final editorial outcome, their assessments exhibit a robust, statistically significant correlation with the eventual decisions. Specifically, manuscripts that are ultimately accepted garner blind reviews with significantly higher Novelty Scores ($\Delta = +0.051^{*}$) and tighter structural prioritization ($\Delta_{\text{Prior.}} = +0.006^{***}$). Conversely, papers that are ultimately rejected accumulate substantially more critical diagnostic feedback, reflected in significantly worse scores for both Critical ($\Delta = -0.049^{***}$) and Minor ($\Delta = -0.018^{**}$) flaws. The magnitude of the Critical score difference is notably $2.7\times$ larger than that of the Minor score. This confirms that human reviewers naturally calibrate the severity and focus of their critiques according to the inherent scientific merit of the manuscript, providing a valid signal that organically drives the final accept/reject decision.

\paragraph{LLMs Exhibit Evaluative Invariance Across Quality Tiers.}
In contrast to human reviewers, whose critiques strongly correlate with the final editorial outcome, LLM reviewers exhibit a highly invariant and consistent evaluative pattern regardless of whether a paper is ultimately accepted or rejected. Across all five automated systems, the metric differences are predominantly uniform, with only DeepReview's Prioritization Score reaching statistical significance ($\Delta = +0.011^{***}$). While the LLMs demonstrate a slight directional alignment with humans on interpretive dimensions---yielding marginally positive $\Delta$ for Novelty and negative $\Delta$ for Minor flaws---these shifts lack statistical power. Notably, this stability is most pronounced in the \textit{Critical Score} dimension. Rather than being swayed by the overall quality of the submission, LLMs apply their internal diagnostic heuristics independently: Reviewer2 maintains a strict standard ($-0.058\textsuperscript{ns}$), TreeReview swings inversely ($+0.043\textsuperscript{ns}$), and the others remain stable near zero.

\paragraph{Summary of Evaluative Invariance.}
The results in Table~\ref{tab:metric_diffs} highlight two fundamentally different review paradigms. Human reviewers display high sensitivity, naturally adjusting the severity of their critiques based on the overall scientific merit of the submission. In contrast, LLMs function as invariant diagnostic scanners. They generate reviews with remarkably stable metric distributions, applying a uniform evaluative standard across all papers. While this invariant behavior suggests that LLMs are not susceptible to the ``halo effect'' of high-quality submissions, it also implies a trade-off: their strict consistency limits their capacity to organically discriminate and highlight fatal methodological flaws with the same adaptive precision as human experts.

\subsection{Robustness Across Judges}
\label{app:judges}
To assess whether PRISM's conclusions depend on the judge, we re-ran the full pipeline (extraction, verification and scoring for all four dimensions) with two additional judges from different model families, \texttt{Claude Haiku 5.5} and \texttt{GPT-5.6-Sol}, alongside \texttt{Gemini~2.5 Flash Lite}, on a 150-paper subset (50 papers per venue: ICLR 2026, ICML 2025, NeurIPS 2025), covering human reviews and the reviews of all five automated systems. Table~\ref{tab:judges_scores} reports per-system scores and Table~\ref{tab:judges_rank} the rankings of the five automated systems with Kendall's $W$.

\begin{table}[H]
\centering\footnotesize
\caption{PRISM scores under three judges (150 papers). Bold: best per metric and judge.}
\label{tab:judges_scores}
\begin{tblr}{
  colspec={l l c c c c c c},
  colsep=4pt, rowsep=0.3pt,
  row{1}={font=\bfseries},
  hline{1,Z}={0.08em}, hline{2}={0.05em},
  hline{5,8,11,14,17}={dashed},
}
Metric & Judge & Human & SEA & Reviewer2 & TreeReview & DeepReview & CycleReviewer \\
\SetCell[r=3]{l} Constructiveness & Gemini & .571 & .504 & .570 & .477 & \textbf{.629} & .530 \\
 & Haiku & .517 & .432 & .620 & .512 & \textbf{.628} & .418 \\
 & GPT & .613 & .531 & .670 & .620 & \textbf{.716} & .567 \\
\SetCell[r=3]{l} Depth of Analysis & Gemini & \textbf{.494} & .380 & .381 & .357 & .481 & .485 \\
 & Haiku & \textbf{.449} & .265 & .389 & .278 & .426 & .330 \\
 & GPT & .441 & .309 & \textbf{.449} & .311 & .407 & .376 \\
\SetCell[r=3]{l} Critical Recall & Gemini & .308 & .237 & \textbf{.613} & .304 & .332 & .228 \\
 & Haiku & .374 & .185 & \textbf{.425} & .144 & .260 & .159 \\
 & GPT & .316 & .197 & \textbf{.448} & .127 & .281 & .177 \\
\SetCell[r=3]{l} Minor Recall & Gemini & .252 & .232 & \textbf{.442} & .312 & .228 & .176 \\
 & Haiku & .284 & .145 & \textbf{.407} & .239 & .215 & .117 \\
 & GPT & .266 & .133 & \textbf{.375} & .311 & .252 & .116 \\
\SetCell[r=3]{l} Novelty & Gemini & .795 & \textbf{.841} & .798 & .829 & .771 & .787 \\
 & Haiku & .649 & \textbf{.708} & .646 & .618 & .629 & .607 \\
 & GPT & .630 & \textbf{.673} & .613 & .615 & .611 & .587 \\
\SetCell[r=3]{l} Prioritization & Gemini & .973 & .968 & \textbf{.977} & .974 & .974 & \textbf{.977} \\
 & Haiku & .965 & .948 & \textbf{.972} & .967 & .949 & .894 \\
 & GPT & \textbf{.902} & .824 & .896 & .840 & .885 & .783 \\
\end{tblr}
\end{table}

\begin{table}[H]
\centering\footnotesize
\caption{Rankings of the five automated systems under each judge, with Kendall's $W$ (R2: Reviewer2; Tree: TreeReview; Deep: DeepReview; Cycle: CycleReviewer).}
\label{tab:judges_rank}
\resizebox{\linewidth}{!}{%
\begin{tblr}{
  colspec={l l l l c},
  colsep=4pt, rowsep=0.3pt,
  row{1}={font=\bfseries},
  hline{1,Z}={0.08em}, hline{2}={0.05em},
}
Dimension & Gemini 2.5 Flash Lite & Claude Haiku 5.5 & GPT-5.6-Sol & $W$ \\
Minor Recall & R2 > Tree > SEA > Deep > Cycle & R2 > Tree > Deep > SEA > Cycle & R2 > Tree > Deep > SEA > Cycle & 0.96 \\
Critical Recall & R2 > Deep > Tree > SEA > Cycle & R2 > Deep > SEA > Cycle > Tree & R2 > Deep > SEA > Cycle > Tree & 0.87 \\
Constructiveness & Deep > R2 > Cycle > SEA > Tree & Deep > R2 > Tree > SEA > Cycle & Deep > R2 > Tree > Cycle > SEA & 0.82 \\
Novelty & SEA > Tree > R2 > Cycle > Deep & SEA > R2 > Deep > Tree > Cycle & SEA > Tree > R2 > Deep > Cycle & 0.80 \\
Depth of Analysis & Cycle > Deep > R2 > SEA > Tree & Deep > R2 > Cycle > Tree > SEA & R2 > Deep > Cycle > Tree > SEA & 0.78 \\
Prioritization & R2 $\approx$ Cycle > Tree $\approx$ Deep > SEA & R2 > Tree > Deep > SEA > Cycle & R2 > Deep > Tree > SEA > Cycle & 0.61 \\
\end{tblr}}
\end{table}

\paragraph{Findings consistent across judges.} Reviewer2 leads in Critical and Minor recall ($W=0.87$ and $0.96$), DeepReview in Constructiveness ($W=0.82$) and SEA in Novelty ($W=0.80$) under all three judges. Exact rankings differ in places, but the leading systems are unchanged, supporting the conclusion that automated reviewers have distinct strengths rather than one system dominating.

\paragraph{Depth of Analysis ($W=0.78$).} Gemini ranks CycleReviewer first, whereas Haiku and GPT rank DeepReview and Reviewer2 first, respectively. The intermediate outputs show that the judges identify premises and assess grounding differently: Gemini's scores are driven more by premise share, whereas the other judges separate systems more by grounding quality. DeepReview remains among the top two automated systems under all three judges, and human reviews rank first under Gemini and Haiku and second under GPT.

\paragraph{Prioritization ($W=0.61$).} This dimension has the lowest ranking agreement. However, nCPS is close to ceiling (0.97--0.99) for reviews containing valid flaws; reviews without valid flaws receive zero, and their frequency varies across judges because of differences in flaw-validity judgments. The ranking variation therefore reflects validity decisions and aggregation rather than substantial differences in how reviewers order the flaws they identify.

\section{Qualitative Analysis \& Case Studies}
\label{app:app_exp3}
\subsection{Depth of Analysis}

\paragraph{Case 1: The Evidentiary Collapse --- Why Claim-Heavy Reviewers Fail.} 

\noindent\textit{Paper: NV-Embed: Generalist Text Embeddings from Decoder-Only LLMs}

\smallskip
\noindent\textbf{Context.}
NV-Embed proposes a generalist embedding model built on decoder-only LLMs, introducing
(1)~a latent attention layer replacing mean-pooling, and
(2)~a two-stage contrastive instruction-tuning pipeline.
The model achieves top-1 performance on the MTEB benchmark.

\begin{table}[H]
\centering
\caption{Per-reviewer DoA statistics for NV-Embed. Avg~GS is the mean grounding score normalized to $[0,1]$ (raw scores 0/1/2 divided by 2)}
\label{tab:case_1rb_stats}
\setlength{\tabcolsep}{6pt}
\begin{tabular}{lcccccc}
\toprule
\textbf{Reviewer} & \textbf{DoA\textsubscript{HM}} & \textbf{R\textsubscript{premise}} & \textbf{Avg GS} & \textbf{Total Args} & \textbf{Premises} & \textbf{Claims} \\
\midrule
\textbf{Human}   & 0.574 & 0.673 & 0.500 & 55 & 37 & 18 \\
DeepReview       & \textbf{0.626} & 0.733 & 0.546 & 15 & 11 &  4 \\
Reviewer2        & 0.178 & 0.152 & 0.215 & 46 &  7 & 39 \\
\bottomrule
\end{tabular}
\end{table}

\paragraph{Human: Dense Technical Grounding Across All Aspects.}
Human reviewers produce 37 premises from 55 total arguments ($R_{\text{premise}} = 0.673$),
with strong grounding quality ($\overline{\text{GS}} = 0.500$).
Premises are component-specific, naming the paper's actual technical building blocks
rather than describing them in generalities:

\begin{quote}\small
\textit{[GS\,=\,1.0, Methodology]} ``The techniques used are: 1.\ latent attention layer that
achieves better pooling/combination of the last layer embeddings.''
\end{quote}
\begin{quote}\small
\textit{[GS\,=\,1.0, Methodology]} ``2.\ a two-stage contrastive instruction tuning method.
First step tuning with in-batch negative and hard negative on retrieval datasets,
and the second step tuning on non-retrieval datasets.''
\end{quote}
\begin{quote}\small
\textit{[GS\,=\,2.0, Experiment]} ``The model achieves top performance on the MTEB benchmark.''
\end{quote}

\noindent Human reviewers also raise critical, component-targeted observations,
such as questioning whether a single innovative algorithmic piece exists
beyond the two engineering contributions.
This combination of evidential support and critical evaluation defines the human gold standard
on this paper.

\paragraph{DeepReview: Exceeds Human DoA via Precision-to-Volume Economy.}
DeepReview produces only 15 arguments, yet 11 are premises ($R_{\text{premise}} = 0.733$,
exceeding Human's 0.673), with the highest average grounding score of any reviewer
($\overline{\text{GS}} = 0.546$).
Its premises are architecturally precise:

\begin{quote}\small
\textit{[GS\,=\,1.0, Methodology]} ``The authors introduce a novel latent attention layer for
pooling embeddings, which outperforms traditional methods like average pooling and
\texttt{<EOS>} token embedding.''
\end{quote}
\begin{quote}\small
\textit{[GS\,=\,2.0, Experiment]} ``Ablation experiments in Table~2 confirm the contribution
of the latent attention layer over alternative pooling strategies.''
\end{quote}

\paragraph{Reviewer2: Self-Referential Grounding and Volume Inflation.}
Reviewer2 generates 46 arguments but only 7 qualify as premises, and \textbf{4 of those 7
carry grounding score~0} ($\overline{\text{GS}} = 0.215$, barely above the minimum).
The 39 claims are structured section-by-section summaries offering no independent
analytical judgment:

\begin{quote}\small
\textit{[Claim, Methodology]} ``The paper introduces NV-Embed, a generalist embedding model
based on decoder-only large language models (LLMs), aimed at enhancing performance in
downstream tasks such as retrieval\ldots''
\end{quote}

\noindent Even the few substantiated premises restate the paper's own justifications
rather than providing independent evaluations:

\begin{quote}\small
\textit{[GS\,=\,0.0, Methodology]} ``The introduction of a latent attention layer for sequence
pooling is theoretically grounded in dictionary learning concepts and is argued to mitigate
information dilution compared to mean pooling or last token pooling.''
\end{quote}
\begin{quote}\small
\textit{[GS\,=\,1.0, Experiment]} ``These results are backed by detailed comparisons and
tabulated scores (see Table~14).''
\end{quote}

\noindent The low grounding quality compounds the low premise ratio:
$R_{\text{premise}} = 0.152$ and $\overline{\text{GS}} = 0.215$ together produce
$\text{DoA}_{\text{HM}} = 0.178$~--- a \textbf{69\% drop} from Human's $0.574$.

\smallskip
\noindent\textbf{Key insight.}
Reviewer2's failure on this paper illustrates a form of analytical failure beyond
pure volume-inflation: even its few ``premises'' ground claims in the paper's own assertions
rather than in independent analytical observations.
The DoA metric correctly penalizes this through the joint harmonic mean of
$R_{\text{premise}}$ and $\overline{\text{GS}}$ --- both of which must be high to achieve
human-level analytical depth.

\paragraph{Case 2: The Surface-Level Trap in Practice.}

\noindent\textit{Paper: VLAP --- Visual-Language Alignment via Pre-trained Word Embeddings}

\noindent\textbf{Context.}
VLAP proposes a lightweight vision-language alignment method that maps visual representations
directly into the pre-trained word embedding space of a frozen LLM using a single trainable
linear layer, supervised by an optimal transport-based assignment objective and an image
captioning loss.
The design is deliberately minimal: the LLM and visual encoder remain frozen, with only the
linear projection trained.

\begin{table}[H]
\centering
\caption{Per-reviewer DoA statistics for VLAP.
         Human row is the mean across 5 individual reviewers.
         Avg~GS normalized to $[0,1]$; DoA\textsubscript{HM} $= \text{HM}(R_{\text{premise}},\,\overline{\text{GS}})$.}
\label{tab:case_vlap}
\setlength{\tabcolsep}{5pt}
\begin{tabular}{lccccccc}
\toprule
\textbf{Reviewer} & \textbf{DoA\textsubscript{HM}} & \textbf{R\textsubscript{premise}}
  & \textbf{Avg GS} & \textbf{Total} & \textbf{Premises} & \textbf{Claims}
  & \textbf{\% Clarity} \\
\midrule
\textbf{Human (mean, $n$=5)} & 0.567 & 0.614 & 0.527 & 14 & 8 & 6 & 2\% \\
Reviewer2        & \textbf{0.551} & 0.569 & 0.534 & 51 & 29 & 22 & \textbf{0\%} \\
CycleReviewer    & 0.483 & 0.529 & 0.444 & 17 &  9 &  8 & \textbf{0\%} \\
SEA              & 0.417 & 0.500 & 0.357 & 14 &  7 &  7 & 29\% \\
DeepReview       & 0.263 & 0.759 & 0.159 & 29 & 22 &  7 & \textbf{0\%} \\
\textbf{TreeReview} & 0.252 & 0.194 & 0.357 & 36 &  7 & 29 & \textbf{29\%} \\
\bottomrule
\end{tabular}
\end{table}

\noindent\textbf{Reviewer consensus: Clarity is irrelevant on a paper with a simple design.}
This is a particularly instructive paper for the surface-level trap because the simplicity of
VLAP's design --- a single linear layer, two losses, frozen backbones --- leaves almost no
room for legitimate reproducibility criticism.
Accordingly, \textbf{DeepReview, CycleReviewer, and Reviewer2 allocate 0\%}
of their premise budget to Clarity, and \textbf{Human} allocates a negligible \textbf{2\%}
(Table~\ref{tab:case_vlap}).
Instead, they focus exclusively on the paper's technical formulation and experimental
comparisons:

\begin{quote}\small
\textit{[GS\,=\,2.0, Novelty]} ``Contrastive alignment in ALBEF, BLIP, and the first-stage
alignment by BLIP2 includes image-text matching and image-grounded text generation.''
\hfill\textit{(Human\_2})
\end{quote}
\begin{quote}\small
\textit{[GS\,=\,2.0, Methodology]} ``An optimal transport-based training objective is proposed
to enforce the consistency of word assignments for paired multimodal data.
This allows frozen LLMs to ground their word embedding space in visual data.''
\hfill\textit{(Human\_3)}
\end{quote}
\begin{quote}\small
\textit{[GS\,=\,2.0, Methodology]} ``Using this method, experiments are done on 3 tasks
--- image captioning, VQA, image-text retrieval --- showing the method outperforms existing
methods.'' \hfill\textit{(Human\_5)}
\end{quote}

\noindent Reviewer2, despite its typical tendency toward low $R_{\text{premise}}$, here
achieves $\text{DoA}_{\text{HM}} = 0.551$ with 29 premises from 51 arguments ($R_{\text{premise}} = 0.569$,
$\overline{\text{GS}} = 0.534$) --- all directed at Methodology and Experimental Design.
CycleReviewer produces 9 premises from 17 arguments with 0\% Clarity and an average
$\overline{\text{GS}} = 0.444$.

\noindent\textbf{SEA: Clarity premises praising, not criticizing.}
SEA allocates 29\% to Clarity, matching TreeReview's proportion.
However, its two Clarity premises are
\emph{positive quality affirmations} about the paper, not complaints:

\begin{quote}\small
\textit{[GS\,=\,0.0, Clarity]} ``The methodology is clearly explained, making it accessible
and understandable, which is crucial for reproducibility and further research.''
\end{quote}
\begin{quote}\small
\textit{[GS\,=\,0.0, Clarity]} ``The paper is well-structured, with comprehensive
experiments and detailed analyses.''
\end{quote}

\noindent Although these also carry GS\,=\,0 (generic, no specific section cited), they
function as strength notes rather than reproducibility criticisms.
SEA then pivots to three substantive Experimental premises, maintaining
$\text{DoA}_{\text{HM}} = 0.417$.

\noindent\textbf{TreeReview: reproducibility boilerplate displaces technical analysis.}
TreeReview produces only 7 premises from 36 total arguments ($R_{\text{premise}} = 0.194$).
Of these 7, \textbf{2 (29\%) are Clarity premises}, both carrying GS\,=\,0 and targeting
the same generic reproducibility axis:

\begin{quote}\small
\textit{[GS\,=\,0.0, Clarity]} ``This omission hinders reproducibility and limits the
ability of other researchers to build upon the work.''
\end{quote}
\begin{quote}\small
\textit{[GS\,=\,0.0, Clarity]} ``This would greatly enhance the reproducibility of the
method.''
\end{quote}

\noindent Neither statement names a specific omission.
The first is a consequence claim (``hinders reproducibility'') with no antecedent in the
review --- it is evidently the continuation of a claim made in a preceding \emph{claim}
argument, not a self-contained premise.
The second is a one-sentence recommendation that names no missing artifact.
On a paper whose entire contribution is a single linear layer with two objectives,
these statements convey essentially no analytical information.

\noindent By contrast, TreeReview's five non-Clarity premises do engage with the paper's
mechanism (optimal transport alignment, the single linear layer, efficiency), but all at
GS\,=\,1 with no external literature anchoring.
This yields $\overline{\text{GS}} = 0.357$ and, combined with
$R_{\text{premise}} = 0.194$, $\text{DoA}_{\text{HM}} = \mathbf{0.252}$ ---
a \textbf{56\% drop} from Human's 0.567.

\noindent\textbf{Key insight.}
The surface-level trap on this paper is not caused by a genuinely unclear manuscript:
four of the six reviewers independently assess that VLAP requires no Clarity criticism
at all.
The trap is therefore \emph{triggered internally} by TreeReview's reviewing heuristic ---
a tendency to produce generic reproducibility premises (``hinders reproducibility'',
``greatly enhance reproducibility'') regardless of whether the paper's design warrants them.
These boilerplate premises consume 2 of TreeReview's 7 available premise slots and
contribute nothing to the analytical depth of the review.

\subsection{Novelty Assessment}

\paragraph{Case 1: The Speculative Critique Trap}

We illustrate the Novelty Score (NS) metric through a representative paper from ICLR, which proposes \textbf{CONPAIR}, a contrastive compositional dataset and \textbf{EVOGEN}, a curriculum contrastive learning framework for improving compositional text-to-image (T2I) generation in diffusion models. Table~\ref{tab:novelty_case1} summarises key statistics for the Human and SEA reviewers on this paper.

\begin{table}[H]
\centering
\small
\caption{Per-reviewer novelty statistics for paper \textbf{CONPAIR}.
  Stances: \textbf{N}=\emph{novel}, \textbf{NN}=\emph{not\_novel},
  \textbf{SW}=\emph{somewhat\_novel}, \textbf{U}=\emph{unclear}.
  $\bar{s}$ denotes the raw mean per-claim score on the $[-2, +2]$ scale; the normalized review-level metric is $NS=(\bar{s}+2)/4$.}
\label{tab:novelty_case1}
\setlength{\tabcolsep}{6pt}
\begin{tabular}{lcccccc}
\toprule
\textbf{Reviewer} & \textbf{$\bar{s}$} & \textbf{\#Claims} & \textbf{\%N} & \textbf{\%NN} & \textbf{\%SW} & \textbf{\%U} \\
\midrule
Human             & $1.20$      & 10                & 50\%         & 30\%          & 0\%           & 20\%         \\
SEA               & $1.73$      &  5                & 80\%         &  0\%          & 20\%          &  0\%         \\
\bottomrule
\end{tabular}
\end{table}

\noindent
Both reviewers correctly identify the paper's core contributions, specifically, the novel use of contrastive learning in the denoising encoder and the CONPAIR dataset's value as a hard-negative compositional benchmark.
The divergence in $\bar{s}$ arises not from a disagreement on \emph{what} is novel, but from how reviewers handle \emph{uncertain} assessments.

\noindent\textbf{Human reviewer} produces 10 claims with a nuanced mixture of stances.
Claim C8 are labelled \emph{unclear} and carry speculative criticism that the metric system cannot verify against any paper in the related-work pool:

\begin{quote}
\textit{[C8, unclear]} ``The ContraFusion model is compared against other methods using the T2I-CompBench dataset. However, it is trained on the Com-Diff dataset, which \textbf{likely overlaps} noticeably with the T2I-CompBench test set.''
\end{quote}

\noindent
This concern about training/test overlap is plausible but entirely speculative: no related paper in the retrieved pool provides evidence for or against dataset overlap between Com-Diff and T2I-CompBench.
Consequently, 7 of 11 related-paper comparisons for C8 return \textsc{Unsupported} or \textsc{Insufficient}, and the claim's per-pair score averages to $-0.18$, substantially dragging down the overall Human $\bar{s}$.
By contrast, the Human's other well-evidenced claims score positively:

\begin{quote}
\textit{[C4, novel]} ``Introducing a contrastive loss in the denoising encoder representation is an interesting idea.'' \hfill $(s_k=+2.0)$
\end{quote}
\begin{quote}
\textit{[C5, not\_novel]} ``A few other datasets with hard-negative compositional images exist, albeit small. Two examples are COLA and Winoground.'' \hfill $(s_k=+2.0)$
\end{quote}

\noindent
C5 exemplifies \emph{calibrated} \textit{not\_novel} assessment: the reviewer correctly names prior datasets (COLA, Winoground) that are confirmed by related-work retrieval, earning a full $+2.0$ score even though the stance is non-endorsing.

\noindent\textbf{SEA reviewer} produces only 5 claims, all either \emph{novel} or \emph{somewhat\_novel}, with no speculative or \emph{unclear} claims:

\begin{quote}
\textit{[C3, novel]} ``The proposed dataset, CONPAIR, is well-designed with a clear progression from simple to complex compositional scenarios.'' \hfill $(s_k=+2.0)$
\end{quote}
\begin{quote}
\textit{[C4, novel]} ``The multi-stage fine-tuning strategy is innovative and addresses the issue of models being overwhelmed by mixed-difficulty data during training.'' \hfill $(s_k=+2.0)$
\end{quote}
\begin{quote}
\textit{[C5, somewhat\_novel]} ``The use of a VQA model to revise generated captions and ensure text-image alignment is an effective approach.'' \hfill $(s_k=+2.0)$
\end{quote}

\noindent
Each SEA claim targets a specific, verifiable contribution (curriculum structure, fine-tuning strategy, VQA-based alignment), and none raises unverifiable speculation. As a result, SEA achieves a higher raw mean claim score ($\bar{s}=1.73$ vs.\ $1.20$) despite generating only half the number of claims.

\vspace{4pt}
\noindent\textbf{Insight — Speculative Assessment Penalty.}
This case illustrates a systematic pattern: Human reviewers raise \emph{unclear}-stance concerns (e.g., suspected data contamination, alignment-quality trade-offs) that are reasonable in context but impossible to corroborate through paper-pool evidence.
The NS metric penalises such claims because they lack verifiable grounding—exactly the behaviour the metric is designed to detect. SEA tends to avoid this failure mode: it focuses on claims that are directly grounded in the methods described in the paper, thus achieving higher evidence-calibrated novelty scores. This finding does \emph{not} imply SEA is a ``better reviewer''; rather, it shows that SEA's positively-biased and evidence-anchored claim style is systematically rewarded by the evidence-grounded NS metric, while the richer, more skeptical human review style can be penalized when speculative concerns appear.

\paragraph{Case 2: Coverage Without} Paper \texttt{3TGUvHmZ2v} from ICML, examines a theoretical paper that characterises the expressivity of fixed-precision Transformer decoders using formal language theory.
The paper establishes three main results:
\textbf{(R1)}~without positional encoding (NoPE), fixed-precision Transformers can recognise only finite and co-finite languages;
\textbf{(R2)}~adding absolute positional encoding (APE) extends expressibility to cyclic languages;
\textbf{(R3)}~relaxing parameter bounds further allows recognition of letter-set languages.
Table~\ref{tab:novelty_case2} summarises reviewer statistics.

\begin{table}[h]
\centering
\small
\caption{Per-reviewer novelty statistics for paper \texttt{3TGUvHmZ2v}.
  Stances: \textbf{N}=\emph{novel}, \textbf{NN}=\emph{not\_novel},
  \textbf{SW}=\emph{somewhat\_novel}.
  $\bar{s}$ denotes the raw mean per-claim score on the $[-2, +2]$ scale; the normalized review-level metric is $NS=(\bar{s}+2)/4$.}
\label{tab:novelty_case2}
\setlength{\tabcolsep}{6pt}
\begin{tabular}{lcccccc}
\toprule
\textbf{Reviewer}  & \textbf{$\bar{s}$} & \textbf{\#Claims} & \textbf{\%N} & \textbf{\%NN} & \textbf{\%SW} & \textbf{Overlap w/ Human} \\
\midrule
Human              & $1.467$     & 5                 & 80\%         & 0\%           & 20\%          & ---                       \\
DeepReview         & $1.500$     & 12                & 58\%         & 42\%          & 0\%           & 5/5 + 7 new               \\
\bottomrule
\end{tabular}
\end{table}

\noindent
\textbf{Human reviewer} makes 5 well-targeted claims, all centred on the paper's three core theoretical results.
Three claims earn the maximum per-claim score $s_k=+2.0$ by directly naming the language classes established:

\begin{quote}
\textit{[H-C1, novel]} ``This paper demonstrates that fixed-precision Transformer decoders without positional encoding are limited to recognizing only finite or co-finite languages$\ldots$'' \hfill $(s_k=+2.0)$
\end{quote}
\begin{quote}
\textit{[H-C5, novel]} ``The paper explores the expressive capabilities of Transformer decoders constrained by a fixed-precision setting, such as specific floating-point arithmetic$\ldots$utilising formal language theory.'' \hfill $(s_k=+2.0)$
\end{quote}

\noindent
Two claims are penalised to $s_k=+0.667$: H-C2 makes a vague generality (``limited to \emph{finite memorization}'') rather than naming the class precisely, and H-C4 (\emph{somewhat\_novel}: ``adding positional encoding improves expressivity but does not alleviate the main limitations'') lacks the concreteness of specifying \emph{cyclic} languages.
The result is a compact but calibrated review: every claim is evidenced, but some are imprecise enough that the related-work pool only partially supports them.

\noindent
\textbf{DeepReview} produces 12 claims, $2.4\times$ more, at a nearly identical raw mean claim score ($\bar{s}=1.500$ vs.\ $1.467$).
Its first 7 claims cover \emph{all} of Human's 5 novelty dimensions, often with greater precision, while claims DR8–DR12 open an entirely new dimension absent from the human review: well-evidenced critical analysis of the paper's \emph{scope limitations}.

\noindent
\emph{Covering Human's novelty dimensions with increased precision:}
\begin{quote}
\textit{[DR2, novel]} ``Introducing absolute positional encoding extends their capabilities to recognizing \textbf{cyclic languages}, while allowing non-finite floating-point values further expands their expressivity to \textbf{letter-set languages}.'' \hfill $(s_k=+2.0)$
\end{quote}
\begin{quote}
\textit{[DR5, novel]} ``The paper's focus on \textbf{constant precision}, as opposed to \emph{logarithmic} precision, is a significant strength, as it reflects the practical constraints of real-world implementations.'' \hfill $(s_k=+2.0)$
\end{quote}

\noindent
DR2 subsumes H-C4 (\emph{somewhat\_novel}) by naming both upgraded classes precisely, earning $s_k=+2.0$ vs.\ H-C4's $+0.667$.
DR5 makes an explicit constant-vs-log-precision distinction that Human's C5 only gestures towards.

\noindent
\emph{Identifying additional gaps through well-evidenced limitation claims:}
\begin{quote}
\textit{[DR10, not\_novel]} ``The paper's analysis of positional encoding is limited to absolute positional encoding (APE) and no positional encoding (NoPE).
It does not explore \textbf{relative positional encodings}, which are commonly used in modern Transformer architectures.'' \hfill $(s_k=+2.0)$
\end{quote}
\begin{quote}
\textit{[DR12, not\_novel]} ``Finally, the paper's analysis is primarily \textbf{theoretical}, and it lacks empirical validation of its findings.'' \hfill $(s_k=+2.0)$
\end{quote}

\noindent
Crucially, these limitation claims are \emph{not\_novel} in stance but still earn $s_k=+2.0$.
This is because the novelty metric rewards \emph{calibration}: the claim that ``relative PE is not studied here'' is verifiable against the related-work pool (papers using RoPE, ALiBi, etc.\ do exist, confirming the gap), so the claim is evidence-grounded.
By contrast, three of DeepReview's limitation claims (DR6: APE/cyclic redundancy; DR8: multi-layer not explored; DR11: sinusoidal vs.\ learned PE) earn $s_k=0.0$, because either they are internally redundant with other DeepReview claims or the related-work pool cannot confirm they are genuine gaps.

\begin{table}[h]
\centering
\small
\caption{Semantic overlap between Human and DeepReview novelty claims for \texttt{3TGUvHmZ2v}.
  DeepReview DR1--DR7 cover all Human claims; DR8--DR12 add new gap analysis.}
\label{tab:novelty_case2_overlap}
\begin{tabular}{lll}
\toprule
\textbf{Human claim} & \textbf{DeepReview equivalent(s)} & \textbf{$s_k$ (DR)} \\
\midrule
H-C1: finite/co-finite (NoPE)              & DR1, DR4 & $+2.0, +2.0$ \\
H-C2: ``finite memorization'' (vague)      & DR3 (rigorous framework) & $+2.0$ \\
H-C3: regular language subclass (NoPE)     & DR1, DR7 & $+2.0, +2.0$ \\
H-C4: PE improves but limits remain (vague)& DR2 (APE $\to$ cyclic/letter-set) & $+2.0$ \\
H-C5: fixed-precision setting (practical)  & DR5 (constant vs.\ log precision) & $+2.0$ \\
\midrule
--- (not in Human) & DR8: multi-layer/multi-head gap & $0.0$ \\
--- (not in Human) & DR9: softmax$\to$hardmax gap & $+2.0$ \\
--- (not in Human) & DR10: relative PE gap (RoPE) & $+2.0$ \\
--- (not in Human) & DR11: sinusoidal vs.\ learned PE & $0.0$ \\
--- (not in Human) & DR12: no empirical validation & $+2.0$ \\
\bottomrule
\end{tabular}
\end{table}

\vspace{4pt}
\noindent\textbf{Insight --- Coverage Expansion with Preserved Calibration.}
This case illustrates a second systematic pattern distinct from Case 1: DeepReview does not achieve a higher raw mean claim score $\bar{s}$ by avoiding criticism, but by \emph{expanding coverage} while maintaining the same calibration quality as human reviewers.
DeepReview's extra volume comes in two flavours: (i) \emph{precision elaboration}---naming specific language classes and precision regimes more exactly than Human; and (ii) \emph{gap enumeration}---identifying well-evidenced scope limitations (no relative PE, no empirical validation) that human reviewers do not consider.
The NS metric rewards both, because both types of claim can be verified against the related-work pool.
Thus DeepReview achieves $2.4\times$ the claim count with only $\Delta\bar{s}=+0.033$, demonstrating that LLM reviewers can generate \emph{denser} novelty assessments that are \emph{equally} well-calibrated to the literature, not merely more verbose.

\subsection{Flaw Identification \& Major Issues Prioritization}

\paragraph{Case 1: Complementary Blind Spots --- The Equation-Level Scanner vs.\ the Practical Assessor.}

\noindent\textit{Paper \texttt{Tv2JDGw920}} (ICML)

\smallskip
\noindent\textbf{Context.}
This paper proposes a preconditioning-based optimizer for Domain Generalization that leverages the One-Step Generalization Ratio (OSGR) to dynamically balance parameter-wise gradient updates.
The paper provides three theoretical analyses (OSGR equalization, PAC-Bayes bound, convergence proof) and extensive experiments across five DG benchmarks.
The canonical flaw bank contains 52 entries (after de-duplication: ${\sim}24$ unique flaws), of which 20 are valid upon independent verification (9 identified only by the LLM reviewer, 9 only by human reviewers, and 2 by both) --- all Minor in severity.

\begin{table}[H]
\centering
\caption{Flaw coverage comparison for \texttt{Tv2JDGw920}.
  After de-duplication, the LLM and Human reviewers each independently identify ${\sim}9$ unique valid flaws, with only \textbf{2 flaws shared} between them, demonstrating largely complementary diagnostic profiles.}
\label{tab:flaw_case1_stats}
\setlength{\tabcolsep}{5pt}
\begin{tabular}{lcccc}
\toprule
\textbf{Reviewer} & \textbf{Valid Flaws} & \textbf{LLM-only} & \textbf{Human-only} & \textbf{Shared} \\
\midrule
\textbf{Human (3 reviewers)} & 11 & --- & 9 & 2 \\
\textbf{Reviewer2 (LLM)}     & 11 & 9  & --- & 2 \\
\midrule
\textbf{Union}               & 20 & 9 & 9 & 2 \\
\bottomrule
\end{tabular}
\end{table}

\noindent\textbf{What the LLM catches that all Humans miss: systematic equation-level scrutiny.}
Reviewer2 identifies 9 unique valid flaws that none of the three human reviewers raise. These flaws share a distinctive pattern: they arise from systematically walking through the paper's mathematical derivations and cross-referencing theoretical claims against their practical implementation.

\begin{quote}\small
\textit{[LLM-only, Methodology]}
``In the PAC-Bayes analysis, the prior $\pi$ is approximated using all data except the current batch. Why is this approximation valid, and what is the impact of this choice on the tightness of the generalization bound?''
\end{quote}

\noindent This flaw targets the data-dependent prior in Theorem~3.6 (Appendix~C.3.1), where the paper states the prior is ``approximated with stochastic gradient descent using all data excluding the current mini-batch.'' Classical PAC-Bayes theory requires the prior to be chosen \emph{independently} of training data. While modern PAC-Bayes extensions accommodate data-dependent priors, the paper does not formally invoke such extensions, leaving a gap in the theoretical justification.

\begin{quote}\small
\textit{[LLM-only, Methodology]}
``In Corollary 3.2, the preconditioning factor $p_j$ is derived under the assumption that gradients are independent across parameters. However, in practice, gradients are often highly correlated.''
\end{quote}

\noindent Verifying this requires tracing from Equation~(48) through~(50) in Appendix~C.2, where the factorization $\mathbb{E}(\mathbf{p} \odot \boldsymbol{\varepsilon} \cdot \boldsymbol{\varepsilon}') = \sum_j p_j \cdot \sigma^2(\theta_j)$ implicitly assumes zero cross-parameter covariance --- a standard but unverified assumption in deep networks with shared activations across layers.

\begin{quote}\small
\textit{[LLM-only, Clarity]}
``The paper frames the use of OSGR in the optimizer as novel, [but] it is not entirely clear how this differs from existing adaptive learning rate methods. For instance, Adam already adapts learning rates based on gradient magnitudes.''
\end{quote}

\noindent Table~1 in the paper decomposes the proposed optimizer and Adam into convergence and alignment terms, but the structural similarity between the two requires parsing dense notation to appreciate the distinction --- precisely the kind of close reading that time-constrained reviewers may skip.

\smallskip
\noindent\textbf{What Humans catch that the LLM misses: claim--evidence calibration and field norms.}
Human reviewers identify 9 unique valid flaws that Reviewer2 entirely overlooks. These flaws target gaps between what the paper \emph{claims} and what the \emph{evidence} supports --- a form of practical wisdom rooted in familiarity with community expectations.

\begin{quote}\small
\textit{[Human-only, Methodology]}
``The paper claims [the optimizer] promotes domain-invariant features, but it doesn't directly evaluate this claim by examining feature representations. For example, some DG papers use metrics like center divergence between domain features.''
\hfill\textit{(Human\_2)}
\end{quote}

\noindent The paper provides qualitative visualizations showing class separation across domains, but no quantitative domain-invariance metric (e.g., MMD, center divergence) --- a standard evaluation in the DG literature that the LLM does not demand.

\begin{quote}\small
\textit{[Human-only, Methodology]}
``The claim that uniformly distributed OSGR across parameters indicates better generalization\ldots is stated as a conjecture rather than a theorem, and while intuitively supported, it's not rigorously demonstrated.''
\hfill\textit{(Human\_2)}
\end{quote}

\noindent The paper transparently labels this as a ``Conjecture'' (\S3.2) and provides partial support via Jensen's inequality (Appendix~C.3.2). However, only a human reviewer flags the intellectual-honesty gap between a conjecture and a theorem --- the LLM accepts the conjecture's supporting evidence without questioning its formal status.

\smallskip
\noindent\textbf{Where the LLM hallucinates: asserting absence of content that exists.}
Reviewer2 also generates several flaws that are \emph{directly contradicted} by the paper.
The most striking: the LLM claims ``the paper does not adequately explain the computational cost relative to simpler optimizers,'' despite the paper explicitly reporting training times across multiple iteration budgets (e.g., proposed method: 4,292\,s vs.\ Adam: 5,443\,s at 5K iterations).
Similarly, the LLM asserts the paper ``lacks qualitative insights (e.g., feature maps, attention weights),'' overlooking multiple visualizations already present in the manuscript.
These fabrications follow a consistent pattern: the LLM evaluates flaw claims in isolation without cross-referencing the manuscript's actual content.

\smallskip
\noindent\textbf{Key insight.}
This case demonstrates that LLM and human reviewers operate as \emph{complementary diagnostic instruments} with near-zero overlap.
The LLM excels at systematic, equation-level verification --- surfacing implicit assumptions (gradient independence, data-dependent priors) and theory-practice gaps that require cross-referencing proofs against algorithms.
Human reviewers excel at claim--evidence calibration --- demanding quantitative backing for qualitative claims and recognizing when a conjecture substitutes for a theorem.
Neither perspective subsumes the other: the union of their flaw sets produces substantially broader diagnostic coverage than either alone, while both are susceptible to distinct failure modes (LLM: content hallucination; Human: limited equation-level scrutiny under time pressure).

\paragraph{Case 2: The Diagnostic Volume Advantage --- Broader Coverage Through Exhaustive Scanning.}

\noindent\textit{Paper \texttt{aapUBU9U0D} (ICLR)}

\smallskip
\noindent\textbf{Context.}
This paper proposes an iterative data augmentation pipeline for fine-tuning LLMs on Operations Research (OR) tasks.
The method generates synthetic optimization problems via an evolutionary strategy, validates them through four LLM-based checkers, and fine-tunes LLaMA-3-8B on the resulting corpus.
Performance is evaluated on three OR benchmarks.
The canonical flaw bank contains 28 entries, all valid upon verification, contributed equally by human reviewers (14~flaws) and Reviewer2 (14~flaws).

\begin{table}[H]
\centering
\caption{Flaw coverage and aspect distribution for \texttt{aapUBU9U0D}.
  Despite equal flaw counts, the two reviewer groups cover substantially different \emph{aspects} of the paper, with Reviewer2 providing broader topical coverage across all four macro-categories.}
\label{tab:flaw_case2_stats}
\setlength{\tabcolsep}{5pt}
\begin{tabular}{lccccc}
\toprule
\textbf{Reviewer} & \textbf{Valid} & \textbf{Methodology} & \textbf{Experimental} & \textbf{Clarity} & \textbf{Applicability} \\
\midrule
\textbf{Human (4 rev.)} & 14 & 3 & 7 & 2 & 2 \\
\textbf{Reviewer2 (LLM)} & 14 & 4 & 4 & 1 & 5 \\
\midrule
\textbf{Union}           & 28 & 7 & 11 & 3 & 7 \\
\bottomrule
\end{tabular}
\end{table}

\noindent\textbf{Human reviewers: deep but concentrated coverage.}
The four human reviewers collectively produce 14 valid flaws, with a strong concentration on Experimental Design (7 of 14, 50\%).
Their critiques are precise and field-specific, reflecting deep familiarity with OR benchmarking conventions:

\begin{quote}\small
\textit{[Human, Experimental Design]}
``No comparison with traditional OR solvers is provided.
The paper claims to advance `automation of decision-making' but never shows whether LLM-based modeling is competitive with established OR methods.''
\end{quote}

\begin{quote}\small
\textit{[Human, Experimental Design]}
``The results should clarify whether COPT or GUROBI is used for evaluation, and whether baselines were given equal solver access.''
\end{quote}

\noindent These flaws target the paper's core experimental validity --- precisely the type of domain-expert criticism that requires knowledge of solver ecosystems and OR benchmarking norms.
However, human coverage leaves notable gaps: only 2 of 14 flaws address Applicability and Limitations, and no human reviewer questions the pipeline's scalability to large-scale industrial problems or its dependence on English-only data.

\smallskip
\noindent\textbf{Reviewer2: broader aspect coverage with systematic gap enumeration.}
Reviewer2 produces an equal number of valid flaws (14), but distributes them more evenly across aspect categories.
Notably, 5 of its 14 flaws target Applicability and Limitations --- a category that human reviewers largely neglect:

\begin{quote}\small
\textit{[LLM-only, Applicability]}
``The scalability of the pipeline to large-scale optimization problems with thousands of variables and constraints is not evaluated.''
\end{quote}

\begin{quote}\small
\textit{[LLM-only, Applicability]}
``The pipeline's applicability beyond linear and mixed-integer programming --- for example, to nonlinear or stochastic optimization --- remains unverified.''
\end{quote}

\begin{quote}\small
\textit{[LLM-only, Methodology]}
``The four validation checkers are entirely LLM-prompt-based.
No formal algorithmic specification, error bounds, or coverage guarantees are provided for the checking pipeline.''
\end{quote}

\noindent These flaws arise from the LLM's systematic scanning pattern: rather than evaluating only the paper's explicit claims, Reviewer2 probes the \emph{boundaries} of the contribution --- what is not tested, not formalized, and not generalized.
This ``boundary probing'' behavior is precisely what drives Reviewer2's elevated recall at the macro level: by exhaustively questioning scope and applicability, the LLM surfaces valid concerns that time-constrained human reviewers deprioritize in favor of core experimental scrutiny.

\smallskip
\noindent\textbf{The aspect complementarity pattern.}
When the two flaw sets are merged, the union covers all four macro-categories substantially: Methodology (7), Experimental Design (11), Clarity (3), and Applicability (7).
Neither reviewer group alone achieves this breadth.
Human reviewers anchor the evaluation in domain-specific experimental rigor, while Reviewer2 extends coverage into methodological formalization and scope limitations.

\smallskip
\noindent\textbf{Key insight.}
This case concretizes the volume-to-coverage advantage observed at the aggregate level.
Reviewer2's exhaustive scanning style does not merely produce \emph{more} flaws --- it produces flaws in \emph{different diagnostic categories} than human reviewers, systematically covering scope limitations and methodological formalization gaps that humans deprioritize.
The resulting union of human and LLM flaw sets achieves broader aspect coverage than either alone, reinforcing the practical value of LLM-augmented peer review as a diagnostic broadening mechanism rather than a replacement for human expertise.

\subsection{Multi-dimensional Constructiveness}

\paragraph{Case: The Actionability Gap --- From Diagnosis to Prescription.}

\noindent\textit{Paper: GenColor: A Diffusion-Based Framework for Color
Enhancement in Digital Photography} (NeurIPS)

\smallskip
\noindent\textbf{Context.}
GenColor proposes a no-reference color enhancement pipeline consisting of three
learned components: 
(1)~a diffusion-based Color Generation Module, 
(2)~a Texture Preservation Module, and 
(3)~a post-processing Global Adjustment step.
The system is trained on the authors' proprietary ARTISAN dataset and evaluated
via user studies and no-reference image quality metrics.

\begin{table}[H]
\centering
\caption{Per-system Constructiveness scores for GenColor.
All dimensions are on a $[0,2]$ raw scale.}
\label{tab:case_gencolor_constructiveness}
\setlength{\tabcolsep}{5pt}
\begin{tabular}{lccccccr}
\toprule
\textbf{System} & \textbf{MCS} & \textbf{D1:Act} & \textbf{D2:Spec}
  & \textbf{D3:Just} & \textbf{D4:Sol} & \textbf{D5:Tone} & \textbf{ARCs} \\
\midrule
\textbf{Human}        & 0.488 & 0.721 & 1.767 & 0.372 & 0.326 & 1.698 & 43 \\
\textbf{DeepReview}   & \textbf{0.724} & \textbf{1.588} & \textbf{2.000}
  & 0.882 & \textbf{1.059} & 1.706 & 17 \\
Reviewer2        & 0.800$^*$ & 1.000 & 2.000 & \textbf{2.000} & 1.000 & \textbf{2.000} & 24 \\
CycleReviewer    & 0.400 & 0.833 & 1.833 & 0.167 & 0.167 & 1.000 &  6 \\
TreeReview       & 0.434 & 0.793 & 1.517 & 0.241 & 0.207 & 1.586 & 29 \\
\bottomrule
\multicolumn{8}{l}{\small $^*$Reviewer2's MCS is inflated by near-perfect D3 scores
(verbose observation-only summaries; see text).}
\end{tabular}
\end{table}

\paragraph{Human: Precise Diagnosis, Absent Prescription.}
Human reviewers produce 43 ARCs with respectable specificity
($\overline{D2} = 1.767$), demonstrating solid familiarity with the paper's
technical details.
However, D4 (Solution) averages only $0.326$:
the majority of human ARCs identify a problem but stop short of
prescribing a remedy.
Representative human weakness ARCs illustrate this gatekeeping pattern:

\begin{quote}\small
\textit{[D1\,=\,1, D4\,=\,0]}
``The method's near-deterministic nature raises concerns about user control
and the ability to capture personalized styles.''
\end{quote}
\begin{quote}\small
\textit{[D1\,=\,1, D4\,=\,0]}
``The proposed method has a relatively long runtime compared to
lightweight comparison models.''
\end{quote}
\begin{quote}\small
\textit{[D1\,=\,1, D4\,=\,0]}
``Previous methods trained on the proposed dataset perform worse,
questioning the dataset's general usefulness.''
\end{quote}

\noindent Each statement correctly identifies a real limitation of GenColor:determinism, efficiency, dataset scope but offers no concrete path forward
for the authors to address it.
This is the hallmark of \textit{diagnostic gatekeeping}: the review functions as a verdict, not a guide.

\paragraph{DeepReview: Prescriptive Constructiveness at Scale.}
DeepReview produces only 17 ARCs, yet achieves
$\overline{D1} = 1.588$ and $\overline{D4} = 1.059$,
both substantially above Human.
Critically, \textbf{six of its ARCs reach the maximum solution score
D4\,=\,2}~--- meaning the feedback specifies not only \emph{what} is missing
but \emph{how} to address it:

\begin{quote}\small
\textit{[D1\,=\,2, D4\,=\,2]}
``Include comprehensive ablation studies by systematically removing or
modifying components to evaluate their individual contribution.''
\end{quote}
\begin{quote}\small
\textit{[D1\,=\,2, D4\,=\,2]}
``Provide a detailed analysis of the computational cost of each component
and the overall pipeline, including training time, inference time,
and memory requirements.''
\end{quote}
\begin{quote}\small
\textit{[D1\,=\,2, D4\,=\,2]}
``Clearly articulate the novelty of the specific combination and modifications
of existing components, detailing unique aspects such as the training regime,
degradation scheme, and weight blending strategy.''
\end{quote}

\noindent Compared to the human weakness ARCs above, these comments address
the \emph{same underlying concerns} (ablation coverage, efficiency, novelty
justification) but close the loop by telling the authors exactly what artifact
to produce or section to revise.

\paragraph{Reviewer2: High MCS via Justification Inflation, Not Solutions.}
Reviewer2 achieves the highest raw MCS on this paper (0.800), driven entirely
by near-perfect D3 scores ($\overline{D3} = 2.000$).
Its 24 ARCs are detailed, well-grounded observations~--- but they are
\emph{observations}, not directives.
D1 averages only 1.000 and D4 averages 1.000, because every ARC points to a
general direction at best (``no systematic analysis is provided'',
``no theoretical basis is provided'') without specifying an implementable fix.
This reveals a limitation of MCS as a holistic score: high D3 can mask absent D4.
The D4 sub-score surfaces this asymmetry directly.

\paragraph{CycleReviewer and TreeReview: Vagueness and Volume.}
CycleReviewer contributes 6 ARCs with $\overline{D4} = 0.167$ and
$\overline{D3} = 0.167$~--- vague, unsubstantiated comments such as:
\begin{quote}\small
\textit{[D1\,=\,1, D4\,=\,0]}
``The method might struggle with images exhibiting complex textures
or patterns.''
\end{quote}
TreeReview produces 29 ARCs but maintains $\overline{D4} = 0.207$;
its highest-scoring feedback requests reproducibility details
(training parameters, hyperparameter selection) rather than engaging
with the paper's core design choices.

\smallskip
\noindent\textbf{Key insight.}
This case illustrates the central behavioral gap in constructiveness:
even when human reviewers are technically perceptive (high D2, moderate D3),
they default to \emph{problem identification without resolution}.
DeepReview's architectural orientation toward explicit remediation~---
reflected in D4 exceeding human baseline by $+0.733$ and D1 by $+0.867$~---
demonstrates that high constructiveness is not a matter of writing more,
but of \emph{closing the feedback loop} from critique to actionable prescription.

\section{Limitations and Future Works}
\label{sec:app_limitation}
While PRISM provides a rigorous, multi-dimensional benchmarking framework for automated peer review, we acknowledge several limitations that highlight avenues for future research.

\paragraph{Domain Generalization.}
Our dataset comprises $600$ manuscripts exclusively from premier machine learning and representation learning venues (ICLR, ICML, NeurIPS). The structural norms, citation densities, and evaluation criteria in these venues differ from those in other scientific disciplines (e.g., clinical medicine, humanities, or pure mathematics). Consequently, the current instantiation of PRISM may require recalibration before deployment in non-ML domains.

\paragraph{Temporal Scope.}
\label{sec:app_temporal_scope}
PRISM is anchored to the three most recently concluded venue-years available at the time of dataset construction (ICLR 2026, ICML 2025, and NeurIPS 2025). This temporal choice was made to reduce the risk of benchmark contamination: where cutoff information is available, the selected manuscripts and their human reviews were released after the reported training cutoffs of the evaluated reviewer systems and the LLM judge. A benchmark constructed from artifacts that a model could plausibly have encountered during training would provide a weaker separation between genuine reviewing capability and memorization. 
Nevertheless, temporal filtering cannot guarantee the complete absence of contamination. Training corpora and model update schedules, particularly for proprietary models, are not fully observable, and post-cutoff release dates therefore provide only a practical mitigation rather than a guarantee that benchmark instances were never encountered during training.
We additionally verified that the selected venue-years do not materially alter our conclusions: across an earlier and broader pool of candidate venue-years considered during dataset construction, the relative ranking of all five automated reviewer systems remains unchanged across PRISM dimensions. However, as scientific manuscripts and peer reviews continue to accumulate publicly, currently uncontaminated benchmark instances may eventually enter the training corpora of future reviewer or judge models.
We therefore view PRISM primarily as a reusable evaluation framework rather than a fixed dataset: future benchmark releases can periodically refresh the evaluation corpus with newly released conference cycles while retaining the underlying evaluation methodology.

\paragraph{LLM Dependency: Hallucination, Prompt Sensitivity, and Judge Bias.}
A foundational premise of PRISM is delegating complex tasks---such as text atomization, fact-finding, and scoring---to frontier LLMs. It is well-documented that LLMs are heavily susceptible to hallucinations (fabricating non-existent critiques or citations) and prompt sensitivity (where minor structural variations in instructions yield divergent outputs). To actively mitigate these vulnerabilities, PRISM strictly departs from monolithic, single-prompt evaluation. By decomposing the framework into constrained, multi-phase pipelines and enforcing deterministic decoding, we significantly restrict the generation space and effectively filter out hallucinated noise. Nevertheless, this dependency introduces residual bottlenecks: atomizing isolated sentences inherently risks context loss, fact-finding remains bounded by the coverage of external retrieval APIs, and models acting as judges may still retain subtle internal priors for specific rhetorical styles. 
In this work, our primary evaluation pipeline is instantiated using \texttt{Gemini~2.5 Flash Lite}. We measured judge sensitivity with \texttt{Claude Haiku 5.5} and \texttt{GPT-5.6-Sol} (App.~\ref{app:judges}), and prompt sensitivity with paraphrased and definition-only prompts (App.~\ref{app:prompt}). The leading systems for flaw recall, constructiveness and novelty are stable, but DoA and prioritization remain the most judge- and prompt-sensitive dimensions, and model-specific biases cannot be fully ruled out. Future work should develop uncertainty quantification to prevent edge-case extraction errors from cascading into downstream metrics.

\paragraph{Retrieval Coverage.}
The stage-wise audit (App.~\ref{app:audit}) identifies retrieval as the weakest stage: the candidate pool recovers 34\% of the works cited by reviewers and often contains the reviewed paper itself. NS should therefore be read as grounding within the retrieved literature. We plan three fixes: (i)~excluding the reviewed paper from its own pool (which lowers NS by 0.064--0.074 for every system with the ranking essentially unchanged), (ii)~filtering statements that are not novelty judgments, and (iii)~retrieving reviewer-cited works directly by title.

\paragraph{Human Baseline.}
Our human baseline consists of the human reviews submitted to ICLR 2026, ICML 2025 and NeurIPS 2025 under each venue's LLM policy. It represents real-world peer-review practice rather than strictly unassisted human judgment: some reviews may involve AI assistance, and hybrid human--AI writing cannot be reliably identified retrospectively~\citep{sadeghian2026peerprism}. Each review was nevertheless submitted under a human reviewer's responsibility, unlike fully automated reviewing.

\paragraph{Verification Scope.}
The flaw verifier sees only the manuscript, which is the information reviewers have when they write their reviews; concerns established only later, in the author--reviewer discussion, are outside its scope. Discussion-based filtering, as used by CoCoReviewBench~\citep{deng2026cocoreviewbench}, is complementary.

\paragraph{Future Works} 
\label{app:future_work}
We identify four priority directions:
(1)~\textit{Cross-domain generalization}---recalibrating PRISM for clinical medicine, social sciences, and pure mathematics. 
(2)~\textit{Judge robustness}---extending the three-judge study (App.~\ref{app:judges}) to more judge families and to agreement with human raters on DoA and prioritization. 
(3)~\textit{Deeper outcome validation}---while Section~\ref{sec:accept_reject} shows that only human review scores are predictive of the accept/reject decision, extending this analysis to additional downstream outcomes, such as post-review author satisfaction, would further test whether PRISM metrics capture meaningful review quality rather than superficial fluency.
(4)~\textit{Benchmark sustainability}---periodically refreshing the evaluation corpus with newly released conference cycles to limit contamination risk while retaining PRISM's underlying evaluation methodology.
\input{sections/appendix_H}

%% file: sections/appendix_H.tex
\section{Additional Validation and Analyses}
\label{app:additional}

\subsection{Extended Comparison with Related Work}
\label{app:ext_related}
PRISM answers a different question from the works below. Some generate reviews, in which case PRISM can benchmark them as systems; others evaluate reviews using human reference reviews, rubric-based assessments, or authorship detection. While rubric-based methods may also assess multiple dimensions, PRISM extracts individual units (argumentative units, novelty claims, flaw arguments, atomic comments), verifies flaws against the manuscript and novelty claims against prior literature, and scores human and LLM reviews under the same protocol, without treating human reviews as the reference.

\begin{table}[H]
\centering\footnotesize
\caption{Extended comparison with recent reviewer systems and review evaluators.}
\label{tab:ext_related}
\begin{tblr}{
  width=\linewidth,
  colspec={X[1.1,l] X[1.1,l] X[1.7,l] X[2.1,l]},
  colsep=3pt, rowsep=1.5pt,
  row{1}={font=\bfseries},
  hline{1,Z}={0.08em}, hline{2}={0.05em},
}
Work & Type & How it evaluates reviews & How PRISM differs \\
REMOR~\citep{taechoyotin2025remor} & Reviewer system (reasoning LLM, SFT + GRPO) & Multi-aspect \emph{training reward}: criticism, examples, importance-based prioritization, relevance via text overlap (METEOR) with the manuscript & The reward measures what a review contains, not whether it is correct. PRISM verifies each flaw against the manuscript and each novelty claim against prior literature, and can benchmark REMOR as a system. \\
ScholarPeer~\citep{goyal2026scholarpeer} & Reviewer system (search-enabled multi-agent) & Pairwise LLM judge on 5 dimensions; H-Max scores an AI review 1--10 against the set of human reviews & Pairwise and review-level, so scores are not traceable to statements, and human reviews serve as the reference rather than being scored. \\
ReviewGrounder / ReviewBench~\citep{li-etal-2026-reviewgrounder} & Reviewer system + benchmark & Paper-specific rubrics derived from guidelines, the paper and the human reviews; an LLM evaluator scores 8 dimensions & Rubrics are anchored on human reviews and scored per review. PRISM verifies individual units and pools flaws from all reviewers. \\
CoCoReviewBench~\citep{deng2026cocoreviewbench} & Evaluation benchmark (3,900 papers) & Category-level comparison with human references, filtered using reviewer--author--meta-review discussions; categories without human reviews are skipped & The closest to PRISM: both target correctness rather than overlap. It cannot credit issues no human raised, and it scores a human review only against the other human reviews of the paper (leave-one-out); PRISM credits such issues and scores human and LLM reviews against the same verified reference. Its use of discussions is complementary. \\
PeerPrism~\citep{sadeghian2026peerprism} & AI-authorship detection benchmark (20,690 reviews) & Whether detectors identify the origin of a review's ideas vs.\ its text & A different question: who wrote a review, not how good it is. \\
PeeriScope~\citep{ebrahimi2026peeriscope} & Review-quality platform & Text features, rubric-guided LLM assessments and supervised prediction & A review-level quality profile, without verification against the manuscript or the literature. \\
RottenReviews~\citep{ebrahimi2025rottenreviews} & Review-quality benchmark & Expert annotations, feature-based models and LLM judges & Finds that LLM judges align poorly with experts on holistic quality, which motivates PRISM's unit-level verification. \\
\end{tblr}
\end{table}

\paragraph{Would the results differ on the same reviews?} By construction, in two ways. First, suites that score against human references (ReviewBench, CoCoReviewBench, H-Max, overlap metrics) give no credit to a valid flaw that no human raised; PRISM credits such flaws once they are verified against the manuscript. Second, because these suites treat human reviews as the reference, they cannot score humans and LLMs on equal terms, and therefore cannot reveal two of our central findings: some LLMs exceed individual human reviewers in critical-flaw recall (\S\ref{sec:exp_flaw}), yet only human review scores separate accepted from rejected papers (\S\ref{sec:accept_reject}).

\subsection{Prompt Sensitivity}
\label{app:prompt}
\paragraph{Design.} To isolate the effect of prompt wording, we re-ran all four pipelines on 30 papers $\times$ 6 reviewer types under the original prompts (P0) and two variants: (P1) paraphrased prompts with identical content, and (P2) definition-only prompts with all examples and anchor phrases removed. Re-running P0 on the same 30 papers gives the run-to-run noise floor. We measure per-review Spearman $\rho$ (consistency of individual review scores) and Kendall's $\tau$ over the six reviewer types (stability of system rankings), both relative to P0.

\begin{table}[H]
\centering\footnotesize
\caption{Prompt sensitivity (30 papers $\times$ 6 reviewer types). Noise floor: original prompts re-run on the same 30 papers.}
\label{tab:prompt_full}
\begin{tblr}{
  colspec={l c c c c},
  colsep=6pt, rowsep=0.3pt,
  row{1}={font=\bfseries},
  hline{1,Z}={0.08em}, hline{2}={0.05em},
}
Metric & Per-review $\rho$ (P1 / P2) & $\rho$ noise floor & $\tau$ (P1 / P2) & $\tau$ noise floor \\
Critical recall & 0.700 / 0.763 & 0.680 & 0.867 / 0.867 & 0.600 \\
Minor recall & 0.668 / 0.707 & 0.691 & 0.733 / 1.000 & 0.867 \\
NS & 0.380 / 0.225 & 0.235 & 0.200 / 0.733 & $-0.200$ \\
MCS & 0.670 / 0.643 & 0.769 & 0.867 / 0.600 & 0.867 \\
nCPS & 0.452 / 0.581 & 0.567 & 0.200 / 0.733 & 0.067 \\
DoA & 0.389 / 0.253 & 0.487 & 0.867 / 0.067 & 0.067 \\
\end{tblr}
\end{table}

\paragraph{Extraction consistency.} The units extracted under P1/P2 overlap with the original extractions about as much as two runs of the original prompts overlap with each other: micro-flaws 0.751 / 0.680 (floor 0.683), constructiveness ARCs 0.629 / 0.618 (floor 0.626), and novelty claims 0.770 / 0.733, with the stance changed in only 8\% / 7\% of aligned claims. Unit counts per review stay within about $\pm 12\%$. Prompt variation therefore does not cause a broad collapse in extraction, although it does not rule out omissions of particular claims.

\paragraph{Where sensitivity remains.} For flaws, agreement is at the noise floor and the top system (Reviewer2) is kept. For novelty, the residual variation comes from retrieval rather than scoring: with claims and evidence fixed, changing only the scoring prompt gives $\rho$ 0.410 / 0.357. For constructiveness, changing only the rubric stays at the re-scoring baseline ($\rho$ 0.716 / 0.736 vs.\ 0.755); under P2 the top pair (DeepReview, Reviewer2) swaps, a pair not separated under the original prompts. DoA is the most prompt-dependent axis by design: P2 removes the anchor examples that calibrate the grounding scale, which raises DoA for every system (+0.054 to +0.170); paraphrasing alone keeps the ranking ($\tau = 0.867$), and no swapped DoA pair is separated under the original prompts. Across all metrics on 30 papers, only one rank swap involves a pair that the original prompts separate: Human/Reviewer2 on nCPS under P1. We draw no ranking conclusion from nCPS, since all systems score $\approx 0.97$.

\subsection{Stage-wise Audit and Disagreement Analysis}
\label{app:audit}
\paragraph{Audit protocol.} The expert study (App.~\ref{sec:app_human_validation}) scores pre-extracted units, so it does not test extraction or retrieval. Researchers who took part in the human validation therefore audited these stages on a sample of reviews, using the same definitions as the pipeline. For each extracted unit they judged whether it was a valid unit of its type (precision), and for each review whether an important unit had been missed (omission). For novelty they also judged whether the retrieved evidence was sufficient to assess each claim.

\paragraph{Extraction and retrieval.} Precision and omission rates are reported in Table~\ref{tab:audit}. Novelty-claim extraction (42\% precision) is the weak extraction stage; its errors are mostly statements that are not novelty judgments rather than missed claims. Retrieval is the weakest stage: the candidate pool recovers only 34\% of the works cited by reviewers (matched by title or abstract), the evidence is judged sufficient for 17\% of claims, and the pool contains the reviewed paper itself for 56--70\% of reviews. Removing the reviewed paper lowers NS by 0.064--0.074 for every system and leaves the ranking unchanged, except for one near-tied pair (gap $< 0.004$).

\paragraph{Effect on final scores.} We recomputed every score after (A1) removing the units the auditors marked invalid, and (A2) additionally applying their corrections that require no new judge calls. Flaw and Constructiveness scores are unaffected: per-review shifts are $\le 0.017$, $\tau = 1.00$, and all LLM--human gap signs are preserved. DoA shifts by 0.029 under A1 ($\tau = 0.73$); under A2 the shift is 0.111, which overstates the effect because missing claims are added without their premises. NS levels shift by 0.06 and its ranking is not stable ($\tau = 0.33$); removing non-novelty statements raises the human NS and roughly halves the LLM--human gaps. We therefore consider the Flaw and Constructiveness conclusions robust to extraction errors and read NS comparisons as grounding within the retrieved literature.

\paragraph{Disagreement analysis.} Expert labels are themselves one reading per item, so we treat them as an alternative expert view rather than as ground truth, and test whether this view changes PRISM's system comparisons. We drew a stratified sample of 370 expert--judge disagreements (30--35 per component, balanced across systems), coded the cause of each manually (all coding was done by human annotators), and recomputed the metrics and rankings with expert labels in place of judge labels, one component at a time (Table~\ref{tab:disagreement}).

\begin{table}[H]
\centering\footnotesize
\caption{Causes of expert--judge disagreement and their effect on system comparisons when expert labels replace judge labels.}
\label{tab:disagreement}
\begin{tblr}{
  width=\linewidth,
  colspec={X[1.1,l] X[2.2,l] X[2.2,l]},
  colsep=3pt, rowsep=1.5pt,
  row{1}={font=\bfseries},
  hline{1,Z}={0.08em}, hline{2}={0.05em},
}
Component & Why the labels differ & Does the system comparison change? \\
Flaw aspect, DoA aspect & Neighbouring topics (e.g., Methodology vs.\ Experimental); limitation or applicability critiques with no matching class for annotators & \textbf{No.} Aspect labels feed no score, only descriptive plots. \\
Flaw severity, D1, D3 & Borderline calls (Critical vs.\ Minor; how specific or justified is ``enough''), in both directions for all systems & \textbf{No.} Only systems that were already statistically tied swap places. \\
Novelty & Mainly sentences that are not novelty claims, and loosely related retrieved papers & \textbf{No.} The shift is similar for every system, so the comparison is preserved. \\
Flaw validity & The judge is conservative in rejecting flaws, especially generic ones & \textbf{No.} Expert labels shift recall toward human reviews; Reviewer2 remains the strongest LLM. \\
D2, D4, D5 & Boundary of ``specific'' and of what counts as a suggested solution & \textbf{No} for the overall MCS ranking (unchanged up to ties); some per-dimension differences become smaller. \\
DoA role and grounding & Reviewer2's numbered ``Header: evidence'' format is segmented differently by the judge and by annotators & \textbf{Partly.} Reviewer2 ranks first under expert labels; the other systems keep their relative order. \\
\end{tblr}
\end{table}

\noindent The judge therefore agrees with experts wherever the main conclusions are drawn: DeepReview leads on MCS, D1 and D4; SEA matches or exceeds human reviews on NS; Reviewer2 is the strongest LLM on flaw recall; and claims with a ``Novel'' stance are better supported. These conclusions hold under both label sources. The per-dimension constructiveness statements of \S\ref{sec:exp_mcs} (e.g., on Solution and Tone) are reported under the primary judge, since some of these differences shrink under expert labels.

\paragraph{Single-reviewer flaws.} The pooled reference requires no agreement among reviewers: each pooled flaw is verified individually against the manuscript, and the verifier is not told how many reviewers raised it. Single-reviewer flaws are the common case: 83.6\% of human-raised flaws in the pool come from a single reviewer. The verifier keeps 94.2\% [93.6, 94.8] of single-reviewer flaws vs.\ 96.4\% [95.7, 96.9] of flaws raised by several reviewers. On the expert-annotated subset, it retains 50 of the 53 single-reviewer flaws that experts marked as valid Critical concerns (94\% [85, 98]), matching its rate for expert-validated Critical flaws raised by several reviewers (93\%).

\subsection{Single-prompt LLM Baselines}
\label{app:naive}
PRISM is model-agnostic, so naive single-prompt reviewers can be evaluated directly. We evaluate two general-purpose models, \textbf{Qwen3.6-27B} and \textbf{DeepSeek-V4.1-Flash}. Each receives the full manuscript and a single prompt asking for a standard conference review (summary, strengths, weaknesses, questions), with no tools and no iterative refinement. All rows of Table~\ref{tab:naive} are computed on the same 60 papers (from ICLR 2026, ICML 2025 and NeurIPS 2025) with the primary judge.

\begin{table}[H]
\centering\footnotesize
\caption{Single-prompt baselines vs.\ benchmarked systems and human reviews on the same 60 papers.}
\label{tab:naive}
\begin{tblr}{
  colspec={l c c c c c c},
  colsep=5pt, rowsep=0.3pt,
  row{1}={font=\bfseries},
  hline{1,Z}={0.08em}, hline{2}={0.05em}, hline{8}={dashed},
}
Reviewer & Depth & Novelty & Critical Recall & Minor Recall & nCPS & Constructiveness \\
Human & 0.481 & 0.797 & 0.370 & 0.270 & 0.981 & 0.572 \\
SEA & 0.360 & 0.817 & 0.163 & 0.211 & 0.976 & 0.511 \\
CycleReviewer & 0.477 & 0.821 & 0.187 & 0.178 & 0.986 & 0.555 \\
DeepReview & 0.471 & 0.782 & 0.263 & 0.255 & 0.965 & 0.647 \\
Reviewer2 & 0.377 & 0.816 & 0.731 & 0.446 & 0.974 & 0.579 \\
TreeReview & 0.348 & 0.803 & 0.224 & 0.331 & 0.978 & 0.476 \\
Naive Qwen3.6-27B & 0.377 & 0.775 & 0.563 & 0.389 & 0.970 & 0.556 \\
Naive DeepSeek-V4.1-Flash & 0.410 & 0.819 & 0.685 & 0.529 & 0.949 & 0.645 \\
\end{tblr}
\end{table}

\noindent Strong general-purpose models are competitive on flaw recall (DeepSeek 0.685 / 0.529) and constructiveness (0.645, on par with DeepReview), but not across the board: DeepSeek has the lowest prioritization score (nCPS 0.949) and remains below human reviews on depth, and Qwen has the lowest novelty score (0.775). No reviewer, naive or specialized, dominates all dimensions, and a single overall quality score would hide exactly these capability-specific differences.